\documentclass[10pt,twocolumn,letterpaper]{article}

\usepackage{iccv}
\usepackage{times}
\usepackage{epsfig}
\usepackage{graphicx}
\usepackage{amsmath}
\usepackage{amssymb}
\usepackage[accsupp]{axessibility} 

\usepackage{algorithm}
\usepackage[noend]{algpseudocode}
\usepackage{url}
\usepackage{pgfplots}
\usepackage{tikz}
\usepackage{subcaption}
\usepackage{booktabs}
\usepackage{multirow}
\usepackage{pgfplotstable}
\usepgfplotslibrary{fillbetween}
\usepackage{bbm}
\usepackage{anyfontsize}
\usepackage{tabularray}
\usepackage{color, colortbl}
\usepackage{float}

\definecolor{cBP}{HTML}{108AE3}
\definecolor{cMF}{HTML}{E3571E}
\definecolor{cdeepBP}{HTML}{1852CC}
\definecolor{cdeepBP8}{HTML}{10AEE3}
\definecolor{cdeepBP32}{HTML}{1CC4DA}
\definecolor{cdeepMF}{HTML}{D62728}
\definecolor{cdeepMF8}{HTML}{ED521F}
\definecolor{cdeepMF32}{HTML}{F69C40}
\definecolor{cMV}{RGB}{44,160,44}
\definecolor{cCL}{HTML}{ED1FD2}
\definecolor{cBayesDGC}{HTML}{A631F5}
\definecolor{Gray}{gray}{0.9}

\DeclareMathOperator*{\argmax}{arg\,max}

\newcommand{\ok}[1]{{\color{cMF}{#1}}}

\newcommand{\khy}[1]{{\color{cBP}{#1}}}
\newcommand{\omh}[1]{{\color{violet}{#1}}}

\newcommand{\set}[1]{\mathcal{#1}}

\usepackage[pagebackref=true,breaklinks=true,colorlinks,bookmarks=false]{hyperref}

\iccvfinalcopy % *** Uncomment this line for the final submission

\def\iccvPaperID{7270} % *** Enter the ICCV Paper ID here

\ificcvfinal\fi

\pgfplotsset{compat=1.18}

\begin{document}

%%%%%%%%% TITLE
\title{Adaptive Superpixel for Active Learning in Semantic Segmentation}

% \author{Anonymous submission}

\author{
$\text{Hoyoung Kim}^1$ \quad\quad 
$\text{Minhyeon Oh}^2$ \quad\quad
$\text{Sehyun Hwang}^2$ \quad\quad
$\text{Suha Kwak}^{1,2}$ \quad\quad
$\text{Jungseul Ok}^{1,2} \thanks{Corresponding author.}$\\
\\
$\text{Graduate School of AI, POSTECH}^1$, \quad\quad
$\text{Dept. of CSE, POSTECH}^2$\\
{\tt\small \{cskhy16, minhyeonoh, sehyun03, suha.kwak, jungseul\}@postech.ac.kr}
% Institution1 address\\
% {\tt\small firstauthor@i1.org}
%  minhyeonoh@postech.ac.kr
% For a paper whose authors are all at the same institution,
% omit the following lines up until the closing ``}''.
% Additional authors and addresses can be added with ``\and'',
% just like the second author.
% To save space, use either the email address or home page, not both
}

\maketitle
% Remove page # from the first page of camera-ready.
\ificcvfinal\thispagestyle{empty}\fi

%%%%%%%%% ABSTRACT
\begin{abstract}
Learning semantic segmentation requires pixel-wise annotations, which can be time-consuming and expensive. To reduce the annotation cost, we propose a superpixel-based active learning (AL) framework, which collects a dominant label per superpixel instead. To be specific, it consists of adaptive superpixel and sieving mechanisms, fully dedicated to AL. At each round of AL, we adaptively merge neighboring pixels of similar learned features into superpixels. We then query a selected subset of these superpixels using an acquisition function assuming no uniform superpixel size. This approach is more efficient than existing methods, which rely only on innate features such as RGB color and assume uniform superpixel sizes. Obtaining a dominant label per superpixel drastically reduces annotators' burden as it requires fewer clicks. However, it inevitably introduces noisy annotations due to mismatches between superpixel and ground truth segmentation. To address this issue, we further devise a sieving mechanism that identifies and excludes potentially noisy annotations from learning. Our experiments on both Cityscapes and PASCAL VOC datasets demonstrate the efficacy of adaptive superpixel and sieving mechanisms.

\end{abstract}

%%%%%%%%% BODY TEXT

\let\hl=\undefined
\newcommand\hl[1]{\textcolor{blue}{#1}}

\begin{figure}[!t]
    % \captionsetup[subfigure]{font=footnotesize}
    \centering
    \begin{subfigure}[h!]{.49\linewidth}
        \centering
        \includegraphics[scale=0.231]{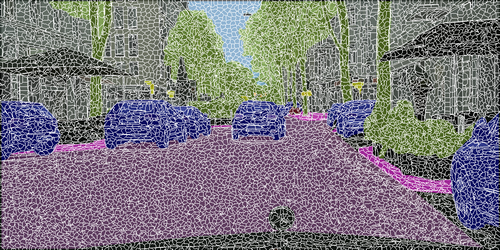}
        \caption{Over-segmented $(t=0)$}
        \label{(a)-adaptive}
        \vspace{2mm}
    \end{subfigure}
    \begin{subfigure}[h!]{.49\linewidth}
        \centering
        \includegraphics[scale=0.231]{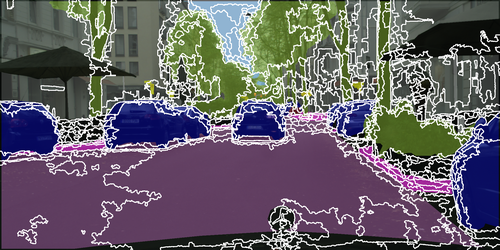}
        \caption{Adaptive merged $(t=2)$}
        \label{(b)-adaptive}
        \vspace{2mm}
    \end{subfigure}
    \begin{subfigure}[h!]{.49\linewidth}
        \centering
        \includegraphics[scale=0.231]{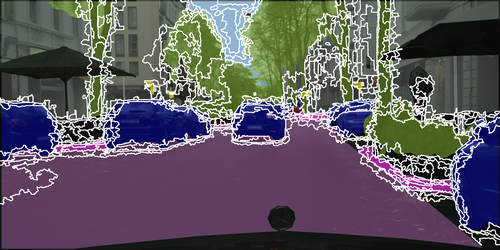}
        \caption{Adaptive merged $(t=4)$}
        \label{(c)-adaptive}
    \end{subfigure}
    \begin{subfigure}[h!]{.49\linewidth}
        \centering
        \includegraphics[scale=0.231]{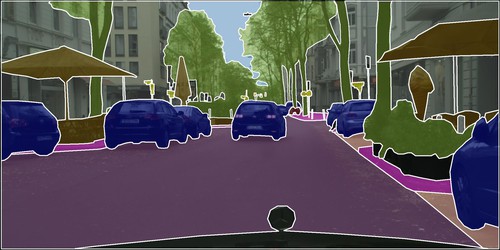}
        \caption{Oracle}
        \label{fig:(d)_adaptive_merged_superpixels}
    \end{subfigure}
    \caption{{\em Examples of adaptive superpixels.} (a) We begin active learning with over-segmented superpixels. (b, c) In each round $t$, we merge superpixels in an adaptive manner using the model from the previous round. % $t-1$. 
    (d) As the round progresses, adaptive superpixels look similar to oracle ones.}
    \label{fig:adaptive_merged_superpixels}
    % \vspace{-3mm}
\end{figure}

\begin{figure*}[t!]
    \centering
    \includegraphics[scale=0.52]{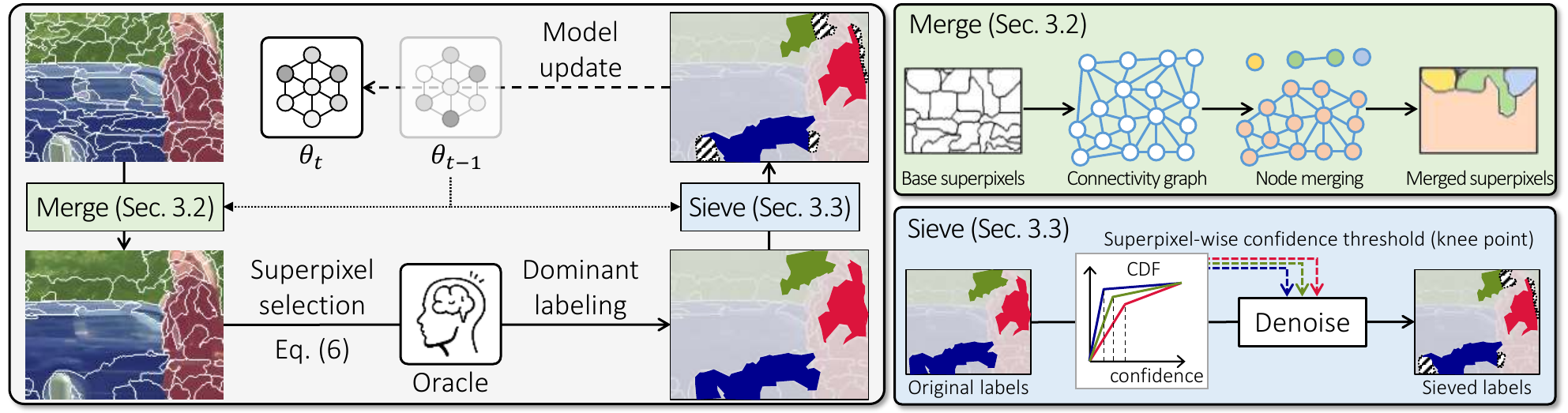}
    \caption{{\em An overview of the proposed framework.} In each round $t$, we merge superpixels with a graph using the latest model, and obtain dominant labels for selected superpixels. The dominant labels are selectively propagated to pixels with confidence above the detected knee point, resulting in the creation of a sieved dataset. Finally, we train a model with the sieved one.}
    \label{fig:method}
    % \vspace{-2mm}
\end{figure*}

\section{Introduction}
With the advent of deep learning, many computer vision tasks including semantic segmentation have dramatically evolved in recent years.
Such advances are thanks to complex deep network models that can learn huge datasets. However, labeling such large datasets is prohibitively time-consuming and labor-intensive, in particular, for semantic segmentation tasks that demand a dense annotation on each pixel \cite{Cordts2016Cityscapes, everingham2015pascal}. Active learning (AL) offers an approach to alleviate the annotation cost  by selectively querying only the most informative samples to annotators.

%selectively obtaining labels for the most informative pixels determined by a well-designed acquisition function.
%In general, the acquisition function prioritizes the selection of unlabeled samples that a model is uncertain about, in order to improve the performance.
% =====
% what is efficient query?
% reviisiting
% =====

Designing an effective form of annotation query is critical in practice 
as it determines the actual annotation cost such as the number of clicks required
and the informativeness per annotation query.
For semantic segmentation, an image-wise query can be asked for a complete annotation on the semantic of every pixel in an image~\cite{chengliang2020suggestive,desai2022active,sinha2019variational,xie2020deal,yang2017suggestive}.
This is a daunting task requiring an enormous amount of clicks to indicate boundaries (using polygons or contours) for each semantic segment or to annotate semantic pixel-wisely, while 
the diversity of contexts which we can observe in a single image is restricted. 
Alternatively, one can design a {\it region-based}
query enquiring only about the dominant label of a small region such as rectangle patch \cite{casanova2019reinforced,qiao2022cpral,xie2022towards} or superpixel 
%(a cluster of neighboring superpixels of similar color )
\cite{cai2021revisiting,siddiqui2020viewal}. 
This is known to be simple yet effective 
as it requires only a single click per query while enabling AL to put more focus on significant regions
and to avoid annotation wastes.

AL with the region-based query needs a delicate generation of candidate regions to be queried. A small region size dilutes the budget efficiency, whereas the dominant labeling even by a perfect annotator is prone to give noisy labels when regions are too large to be consisting of pixels with a single class. However, the previous works~\cite{casanova2019reinforced,qiao2022cpral,siddiqui2020viewal}
rely on a fixed candidate set of regions of uniform size, while we could adjust the size and shape of candidate regions as we train the semantic segmentation model over rounds of AL. This limitation remains even in recent work \cite{cai2021revisiting}
with superpixel candidates providing less 
risk of noisy labels than rectangle ones
since the superpixels are produced, only at the beginning, by a conventional superpixel algorithm,
where conventional superpixel algorithms ~\cite{achanta2012slic,jianbo2000normalized,van2012seeds} cluster adjacent pixels of similar innate features (\eg., color) with implicit or explicit regularization to make similar sizes or shapes of superpixels, \ie., limited freedom of query region.

%\vspace{-.4mm}

\let\oldparskip=\parskip
\parskip=0mm

In this paper, to fully enjoy the benefit in terms of annotation cost while suppressing the risk of noisy labels, 
we devise an AL framework, illustrated in Figure~\ref{fig:method}, consisting of adaptive merging and sieving methods.
The adaptive merging method % (Figure~\ref{fig:method})
repeatedly evolves the candidate superpixels for dominant labeling at every round with the latest model and no explicit regularization on the size and shape of superpixels.
% we propose an adaptive merging algorithm~(Figure~\ref{fig:method}) that aims to increase labeling efficiency by merging adjacent and similar superpixels using a trained model. 
This indeed enables the continual improvement 
of
the superpixels' ability to accurately capture the boundaries of semantic objects~(Figure~\ref{(b)-adaptive} and \ref{(c)-adaptive}),
and a proper variation in the sizes and shapes of superpixels, \ie., larger superpixels being attached to larger semantic objects (\eg., road and building) and smaller ones to smaller objects (\eg., human and vehicle) as shown in the ideal ones~(Figure~\ref{fig:(d)_adaptive_merged_superpixels}).

% where the former
% evolves the candidate superpixels for dominant labeling at every round,
% and the latter identifies and leaves out annotations with the high risk of noisy labels given the latest model.

% In the merging process, superpixels with different dominant labels may be merged, and even if superpixels are merged accurately, noisy labels can exist due to the inherent imperfectness of superpixels.
% As the dominant label of a superpixel propagates to its constituent pixels, the ground-truth of a pixel and the dominant label can differ.

Given the adaptive superpixels, we establish
a corresponding acquisition function being aware of irregular superpixel sizes.
It prioritizes uncertain superpixels of rare classes
in order to query the most informative superpixels while balancing class distributions in the entire annotations. In addition, 
to alleviate the inevitable noise in the dominant labeling, we propose a sieving technique 
that excludes labeled pixels of high potential risks of being different classes than the dominant one. 
To be specific, we identify such pixels of potentially noisy labels by 
per-superpixel sieving with distinct thresholds over superpixels. This provides stabler denoising 
than uniform sieving with a constant threshold, which might aggravate class imbalance in the sieved annotations. 
%the  provides class imbalance.

%relatively unsure pixels within each queried superpixel from training. 
%The remaining labeled areas are then utilized for training the model.
%Here, the per-superpixel sieving provides class imbalance.

% prevents class imbalance 
% since the noise is recognized using a distinct threshold for each superpixel.

\parskip=\oldparskip

Through the integration of adaptive merging and sieving into an AL framework, we achieve improved 
accuracy and budget-efficiency over a baseline method. 
% that relies solely on the dominant labels of superpixels.
% Notably, the merging is effective under small superpixels and the sieving is effective under large superpixels, which is related to the qunantity and quality of labels, respectively.
Notably, the merging demonstrates effectiveness under small-sized superpixels, while the sieving plays a critical role given large-sized superpixels.
%which are associated with the quantity and quality of labels, respectively.
Moreover, we show a consistent improvement 
over existing methods in various settings. %independent of the base size.
% Notably, the benefits are apparent when the size of base superpixels is small. 
% as the merging of adjacent and similar superpixels becomes more effective.
% In addition, we show a consistent improvement independent of the base size.
% minimal performance guarantee that is invariant to the choice of the base size.
We provide a thorough justification of the proposed method using various quantitative measures, 
where we introduce 
a new evaluation metric for superpixel algorithms 
that assesses both (achievable) accuracy and recall,
where the recall is overlooked in 
the existing one, the achievable segmentation accuracy~(ASA)~\cite{liu2011entropy}
but important in the context of AL.
This may give new insights into developing superpixel algorithms.

Our main contributions are summarized as follows:
\begin{itemize}
    \item We propose an adaptive merging algorithm where superpixels are updated at each round (Section~\ref{sec:adaptive-merging}), and show the effectiveness of adaptive merging rather than only merging once (Section~\ref{sec:effect-of-adaptive}).
    \item We alleviate the side effect of noisy labels via a sieving technique (Section~\ref{sec:sieving-technique}), and demonstrate especially efficient under large superpixels (Section~\ref{sec:effect-of-adaptive}).
    \item
    In various realistic experiments, we demonstrate the consistent improvement of the proposed AL framework, consisting of the adaptive merging and sieving methods with the dedicated acquisition function, over existing ones (Section~\ref{sec:effect-of-adaptive}).
    \item
    We provide an insightful analysis on proper superpixels for AL with 
    the new evaluation metric of superpixel algorithms being aware of usage in AL (Section~\ref{sec:confusion-matrix}).
    % \vspace{-1mm}
    % \item
    % and quantitatively analyze adaptive superpixels for reasoning of the improvement (Section~\ref{sec:confusion-matrix}).
    %\item We improve performance with adaptive merging and sieving (Section~\ref{sec:effect-of-adaptive}), and propose new metrics as analysis tools for their effectiveness (Section~\ref{sec:confusion-matrix}).
    % \item We present an oracle superpixel baseline as an upper bound (Section~\ref{para:oracle-superpixels}), and propose novel achievable metrics to evaluate the suitability of superpixels in active learning (Section~\ref{sec:confusion-matrix}).
\end{itemize}

\section{Related work}

% =====================================================
% 
% + Labeling unit in Active learning for semantic segmentation
%   - “whole image sampling”
%   - “Region-based method”: data variability 높임
%   - "Patch-based method"
%   - "Suerpixel-based method"
% + 제안된 방법은 이전 웍들과 달리 사전에 계산된 superpixel 에 의존적이지 않고, 모델이 학습함에 따라 superpixel 을 적절히 merging 함으로써 annotation efficiency 를 증가시킨다.
% + Acquisition function
% + Clustering-based Al for classification 과의 유사점
% ===================================================== 
%\hsh{Conventional AL for segmentation approaches  }
% \ok{Need to start with high annotation cost for segmentation.}

\noindent\textbf{Active learning for segmentation.}
To reduce the labeling cost of semantic segmentation, active learning for segmentation selectively collects labels among unlabeled samples, and they utilize different predefined labeling units.
% To reduce the labeling cost of semantic segmentation, AL for segmentation selectively collects labels among unlabeled samples.
% They utilize different predefined labeling units.
Early approaches~\cite{sinha2019variational,yang2017suggestive} perform image-wise selection and mask labeling.
% Early approaches~\cite{yang2017suggestive,sinha2019variational} perform image-wise selection and obtain a ground-truth mask for a whole image.
% Point-based approach~\cite{shin2021labor} tries to select the most informative points among all pixels and assign a class label to the selected one.
% Patch-based methods~\cite{casanova2019reinforced,golestaneh2020importance,xie2022towards} divide images into rectangular patches and annotate pixels within the patch.
Patch-based methods~\cite{casanova2019reinforced,colling2020metabox+,golestaneh2020importance,mackowiak2018cereals,xie2022towards} divide images into rectangular patches and provide mask label~\cite{casanova2019reinforced,golestaneh2020importance,xie2022towards} or polygon overlay of an object~\cite{colling2020metabox+,mackowiak2018cereals} within the selected patch.
% Furthermore, polygon-based approaches~\cite{colling2020metabox+,mackowiak2018cereals} obtain a class corresponding to the patch and intersections with object boundaries, rather than pixel-wise annotations.
% Superpixel-based approaches~\cite{colling2020metabox+, cai2021revisiting, siddiqui2020viewal} divide the image through an over-segmentation algorithm and utilize each divided superpixel as a labeling unit.
Recently, superpixel-based approaches~\cite{cai2021revisiting, siddiqui2020viewal} split images to perceptually meaningful regions called superpixel by running an off-the-shelf over-segmentation algorithm~\cite{achanta2012slic, ren2003learning, van2012seeds}.
Each superpixel is labeled with a single dominant class, and thus it can be obtained efficiently~\cite{cai2021revisiting}, while a label noise may occur depending on the quality of the superpixel.
We present a new efficient labeling unit, that is initialized with the superpixel but its quality continuously improves by the proposed merging algorithm.
To the best of our knowledge, the proposed method is the first approach to improve the labeling units during active learning for segmentation.

\smallskip
\noindent\textbf{Learning from noisy labels for segmentation.}
Considering the difficulty in acquiring high-quality labels~\cite{Cordts2016Cityscapes}, semantic segmentation often suffers from noisy annotations.
Previous studies address the label noise by using
gradient similarity to the clean label~\cite{yaolearning},
structural constraints~\cite{acuna2019devil,li2021superpixel}, and
noise-aware loss~\cite{oh2021background, yang2020learning}.
A recent approach captures the moment when different classes memorize noisy labels~\cite{liu2022adaptive}.
Most of these methods~\cite{li2021superpixel, oh2021background, yang2020learning} utilize a single confidence threshold to detect label noise within data.
Unlike previous approaches, we propose to detect an adaptive confidence threshold 
per every superpixel queried,
using the Kneedle algorithm~\cite{satopaa2011finding} (Section~\ref{sec:sieving-technique}).
Filtering with the sample-adaptive threshold prevents superpixels with low overall confidence or superpixels containing minor classes from being ignored.
\smallskip\noindent\textbf{
Superpixel mechanisms and their evaluation metric.}
Numerous studies segment an image into superpixels to reduce the computation burden of pixels.
Cut-based approaches~\cite{liu2011entropy,jianbo2000normalized,veksler2010superpixels,zhang2011superpixels} create superpixels by adding multiple minimum cuts into a graph with pixel nodes.
Other methods evolve homogeneous clusters from the initial set of points~\cite{achanta2012slic,levinshtein2009turbopixels}.
For real-time applications, a simple hill-climbing optimization is utilized to enforce color similarity~\cite{van2012seeds}.
Most of methods aim at generating superpixels of predefined size or shape,
and 
the generated superpixels are evaluated by achievable segmentation accuracy and boundary recall compared with ground truth~\cite{liu2011entropy, van2012seeds} or by examining the regularity in superpixel shape~\cite{giraud2017robust,machairas2014waterpixels,schick2012measuring}.
To save labeling costs in active learning, 
it is more important to obtain superpixels as close to the ground-truth segments as possible
without such constraints on the shape or size of superpixels. 
% In addition, the regularity of superpixels accelerates a consistent size of superpixels.
To this end, we propose the merging method (Section~\ref{sec:adaptive-merging}), and a new evaluation metric of superpixel mechanism, that
also takes account of the size of ground-truth segments (Section~\ref{sec:confusion-matrix}).
The proposed metric not only highlights the difference of the ideal superpixel required in active learning
than that in the previous context, but also gives a guideline to develop superpixel algorithms for active learning. 

\omh{
% A superpixel refers to a set of pxiels, where the semantic categories are perceptually equivalent. 
% To generate superpixels, cut-based algorithms \cite{jianbo2000normalized,veksler2010superpixels,zhang2011superpixels,liu2011entropy} represent each pixel as a node and assign the similarity between adjacent pixels to the edge connecting them, followed by gradually adding multiple minimum cuts to achieve a boundary-preserving partition.
% Other approaches develop homogeneous clusters from the initial set of points \cite{levinshtein2009turbopixels, achanta2012slic}.
% To enforce compact and homogeneous superpixels, superpixels are created by maximizing the entropy rate of a random walk on a graph with a balancing term \cite{liu2011entropy}.
% For real-time applications, SEEDS \cite{van2012seeds} leverages a simple hill-climbing optimization to enforce color similarity.
% However, the tendency of superpixel algorithms to produce over-segmented and homogeneous superpixels makes them unsuitable for active learning, which requires a large amount of labels in a single click. 
% Recently, there exists attempts to use deep neural networks with ground truth for superpixel algorithms \cite{tu2018learning, jampani2018superpixel, yang2020superpixel}.

% Increasing precision is analogous to reducing noise.
% Increasing recall is to increase throughput.
% Achievable Precision~(AP) and Recall~(AR) - our contribution (is it right to mention it in related works?)
% Average Precision and Recall
% Achievable Precision and Recall
% Why AP, AR, and AF?

}

\section{Proposed framework}

Given an unlabeled image set $\set{I}$, we consider an active learning scenario with dominant labeling, 
where a query asks an oracle annotator for the dominant class label $\text{D}(s) \in \set{C} := \{1,2,..., C\}$ of an associated superpixel $s$,
and we issue a batch $\set{B}_t$ of $B$ queries for each round $t$. 
Once we enquire the batch $\set{B}_t$, we train a model $\theta_t$ based on the annotations obtained so far.
Recalling a superpixel $s$ is a cluster of neighboring pixels,
the dominant labeling demands much less annotation effort 
than the pixel-wise labeling on every individual pixel $x$ in the same superpixel $s$ or manual segmentation to indicate boundaries separating semantics. The benefit becomes greater with larger superpixels.
Meanwhile, it is prone to noisy labeling as superpixels can be blunt, \ie., including pixels of different semantics.
% In order to fully enjoy the benefit in terms of annotation cost 
% while suppressing the risk of noisy labels, 
% we adaptively merge pixels with similar {\it learned} features on semantics.
% This means that our superpixels evolve round by round as the model improves while superpixels from existing methods cannot as they are based on innate pixel features (e.g., color or position).
% We also devise a sieving mechanism to exclude pixels of high risk of noisy labeling from training.

In order to fully enjoy the benefit in terms of annotation cost while suppressing the risk of noisy labels, 
our framework begins with a warm-up round ($t = 0$; Section~\ref{sec:warm-up}; line \ref{alg1-line1}-\ref{alg1-line2} in Algorithm \ref{algorithm1}) to prepare an initial model from random querying  and iterates
subsequent rounds ($t = 1, 2, \dots$) with the adaptive merging (Section~\ref{sec:adaptive-merging}; line \ref{alg1-line4}-\ref{alg1-line5} in Algorithm \ref{algorithm1}) and sieving (Section~\ref{sec:sieving-technique}; line \ref{alg1-line6}-\ref{alg1-line7} in Algorithm \ref{algorithm1}) methods
to evolve superpixels for dominant labeling round by round and filter out annotations with the high risk of noisy labels given the latest model.
The overall procedure is summarized in Figure~\ref{fig:method} and Algorithm~\ref{algorithm1}. %, \ref{algorithm2}.

\subsection{Warm-up round}
\label{sec:warm-up}
The adaptive merging and sieving methods demand a trained model.
To obtain an initial model, 
we start with a canonical warm-up round, which is identical to the first round of previous work \cite{cai2021revisiting}.
We first use an off-the-shelf superpixel algorithm, namely SEEDS \cite{van2012seeds}, to partition the pixels in each image $i \in \set{I}$
into a set ${S}_0(i)$ of superpixels,
and to produce a base segmentation $\set{S}_0:= \bigcup_{i \in \set{I}} {S}_0(i)$.
Querying a batch~$\set{B}_0$ of $B$ superpixels randomly selected from $\set{S}_0$, we then train a model $\theta_0$ using the dominant labels for $\set{B}_0$. Specifically, % in each round $t=0,1,2,...$, 
to obtain $\theta_0$,
we first initialize $\theta$ at a model pretrained on ImageNet,
and then train it to minimize the following cross-entropy (CE) loss:
\begin{equation}
%\ell_{\text{CE}} =
\hat{\mathbb{E}}_{(x, y) \sim {\mathcal{D}}_0} [ \text{CE}(y, f_\theta (x))] \;,
\end{equation}
where $\set{D}_0 := \{(x, y) : 
\exists s \in \set{B}_0,
x \in s, y(c) = \mathbbm{1}{[c = \text{D}(s)]}$

\!\!\!\!\!\!\!$ \forall c \in \set{C}\}$
is the training data for round $t=0$ without sieving,
and $\! f_\theta( x ) \! \in \! \mathbb{R}^{\! |\set{C}| \!}\!$ is $\! \theta$'s estimate of class probability on pixel $ x $.
%where \text{CE} implies pixel-wise cross entropy between one-hot vector with only one for true dominant label $y$ and predicted label $\tilde y$, where $y, \tilde y \in \mathbb{R}^{|\mathcal{C}|}$.

%\ok{Why SEEDS?}
We remark that we use the initial model $\theta_0$ for round $t=1$.
%and also reuse $\set{S}_0$ as the base superpixels for the entire rounds $t\ge 1$.
In our framework, SEEDS to generate $\set{S}_0$ can be replaced with any other unless $\set{S}_0$ is a fair over-segmentation of semantics 
with a low risk of noisy labeling while partially enjoying the benefit of low annotation cost. 
We note that SEEDS clusters neighboring pixels of similar 
% (relative)
colors
while a semantic consists of multiple colors, typically.
SEEDS, ready-to-use in OpenCV \cite{opencv_library}, easily provides the desired over-segmentation \cite{cai2021revisiting} and a decent performance of $\theta_0$.
%For practitionor, it makes sense (ready-to-use, algorithmic complexity, budget efficiency...)
In addition, the warm-up round with SEEDS corresponds to that in existing work \cite{cai2021revisiting}. Hence, this also enables a fair comparison of our main contributions, \ie., adaptive merging and sieving methods, to existing works. 

\subsection{Adaptive merging}
\label{sec:adaptive-merging}
%\ok{we merge base superpixels for algorithmic complexity}
% Given a model $\theta_{t-1}$ at round $t$ 
% We here describe the procedure 
In advance of dominant labeling in round $t \ge 1$, we first merge the base superpixels in $\set{S}_0$ to obtain $\set{S}_t$ using the model $\theta_{t-1}$ from the previous round. We then select a batch $\set{B}_t$ of $B$ superpixels from $\set{S}_t$ to be annotated using an acquisition function that prioritizes uncertain superpixels of rare class labels.
For simplicity, we often omit the subscript $t-1$
and write $\theta$ for $\theta_{t-1}$.

\begin{algorithm}[t!]
\caption{Proposed Framework}
\begin{algorithmic}[1]
\Require 
  Image set $\mathcal{I}$,
  %superpixel-partitioning function $\mathcal{P}$,
  batch size $B$, and final round~$T$.  
\State Produce base superpixels $\set{S}_0 := \bigcup_{i \in \set{I}}\set{S}_0(i)$ \label{alg1-line1}
%\State Query dominant label $\text{D}(s)$, $\forall s\in\mathcal{B}_1$
%\State $\mathcal{D} \gets \{(s,y[s])\colon s\in\mathcal{B}_1\}$
\State Obtain model $\theta_0$ training with $\mathcal{D}_0$ \label{alg1-line2}
%:= \{(x, \text{D}(s))\colon x \in s, s\in\mathcal{B}_0\}$ 
% querying randomly select $B$ superpixels $\mathcal{B}_0 \subset \mathcal{S}_0$  
% \Comment{Warm-up round}
  \For {$t = 1, 2, \dots, T$}
    \State Adaptively merge the base superpixels and obtain \label{alg1-line4}
    
    \;\;\;\;$\mathcal{S}_t \gets \bigcup_{i \in \set{I}} \Call{AM}{{S}_0(i),\theta_{t-1}}$
    %\{s\in \Call{Adapt}{{S}_0(i),\theta_{t-1}} \colon i\in\set{I}\}$
    % \State Convert superpixels into a graph for all images
    % \State $\mathcal{S}_t \gets \{s\in\mathcal{A}_t(X) \colon X\in\mathcal{X}\}$
    
    \State Select and query $B$ superpixels $\mathcal{B}_t \subset \mathcal{S}_t$ with \eqref{acquisition_function} 
    %for dominant labeling
    \label{alg1-line5}
    %\State Query dominant label $\text{D}(s)$, $\forall s\in\mathcal{B}_t$ \label{alg1-line6}
    \State Sieve %the dominant labeling on every queried superpixel 
    $s \in \bigcup_{t'=0}^{t} \mathcal{B}_{t'}$ and 
    obtain $\set{D}_t$ in \eqref{sieved-superpixel}
    %$s, \forall s \in \bigcup_{ \in [t]} \mathcal{B}_{i}$ via \eqref{sieved-superpixel} 
    \label{alg1-line6}
    % \State Sieve, $\forall (s,y[s]) \in D_t$ into $(\tilde s, y[s]) \in \tilde{D}_t$ via \eqref{sieved-superpixel}
    \State Obtain model $\theta_t$ training with the sieved ${\mathcal{D}}_{t}$ \label{alg1-line7} 
    %and the loss in \eqref{cross-entropy}
    %using the loss in \eqref{eq:final-loss} \label{alg1-line8}
    % := \{(\tilde s,\text{D}(s)) : s \in \mathcal{B}_{[t]} \} $ 
    % via \eqref{cross-entropy} \\
    %\Comment{Main round}
  \EndFor
\State \Return $\theta_T$ \label{alg1-line8} 
\end{algorithmic}
\label{algorithm1}
\end{algorithm}

\smallskip\noindent\textbf{Adaptive merging.} 
To obtain $\set{S}_t := \bigcup_{i \in \set{I}} {S}_t(i)$,
the merging process converts base superpixels $S_0(i)$ into merged ones $S_t(i)$ for each image $i \in \set{I}$.
We hence focus on how we merge given base superpixels $S$ for an image.
To begin with, we convert the superpixels $S$ into a connected graph $\set{G}(S) \! = \! (S, \set{E}(S))$
where $S$ is the set of nodes, each of which corresponds to a base superpixel $s \in S$, and $\set{E}(S)$ is the edge set such that
$(s, n) \in \set{E}(S)$ if a pair of superpixels $s, n \in S$
are adjacent.
Starting from a root node $s \in S$, we then merge neighboring superpixels of similar class predictions with the root $s$ along the breadth-first search tree.
To be specific, a neighbor $n$ is amalgamated with root $s$
only if 
\begin{equation}
d_\text{JS} \big( f_\theta(s) \parallel f_\theta(n) \big) < \epsilon \;,
\label{eq:jsd}
\end{equation}
where  $f_\theta(s) := \frac{\sum_{x \in s} f_\theta(x)}{|\{x : x\in s\}|}$
is the averaged class prediction of superpixel $s \in S$,
and $d_\text{JS}$ is 
a symmetric %\footnote{Asymmetric discrepancy such as KL divergence often causes unstable merging.} 
measure of discrepancy between two distributions, namely
the square root of Jensen-Shannon (JS) divergence.
More formally, 
%between $f_\theta(s)$ and $f_\theta(n)$. 
\begin{equation}
d_\text{JS}(p \parallel q) := \sqrt{\frac{d_\text{KL}(p \parallel \frac{p+q}{2}) + d_\text{KL}(q \parallel \frac{p+q}{2})}{2}} \;,
\end{equation}
where $d_{\text{KL}}$ is the Kullback-Leibler divergence.
Once every node has been either merged to a root
or played as a root, we collect the merged superpixels into $S_t(i)$.
The merging process is formally described in Algorithm~\ref{algorithm2}.

Recalling \eqref{eq:jsd} and the fact that $d_{\text{JS}}$ is a distance metric,
we can guarantee that any pair of superpixels $s$ and $n$ 
has the prediction discrepancy at most $2 \epsilon$ and thus similar uncertainty and predicted label if they are merged.
Hence, the threshold $\epsilon$ governs the impurity of predictions in a merged superpixel.
We also remark that 
the merging process is fully dedicated to collecting pixels of similar predictions
as a part of saving the annotation budget for querying similar pixels repeatedly.
Hence, the merged superpixels can have various sizes differently from existing superpixel algorithms that regularize the superpixel size to be even \cite{giraud2017robust,machairas2014waterpixels,schick2012measuring}.

\begin{algorithm}[t!]
\caption{Adaptive Merging (\textsc{AM})} %(\Call{AM}{a})}
\begin{algorithmic}[1]
\Require Base superpixels $S$, model $\theta$, and threshold $\epsilon$. 
% \Ensure Pairwise disjoint family $\mathcal{S}_t(X)$ of superpixels, each representing merged one
%\Procedure{Adapt}{$S$, $\theta$}
  \State Set $S' \gets \emptyset$ and $\mathcal{G}(S) \gets (S, \set{E}(S))$
    \State Mark $s$ as unexplored for each $s\in S$
  % \EndFor
  \For {$s\in S$ in descending order of $u_\theta(s)$} \label{line:order} 
    \If {$s$ is unexplored}
      \State $S' \gets S' \cup \{\text{\Call{Merge}{$s$, $f_\theta(s)$; $\set{G}$, $\theta$}}\}$
    \EndIf
  \EndFor
  \State \Return $S'$
%\EndProcedure
\Procedure{Merge}{$s$, $f$; $\mathcal{G}$, $\theta$}
  \State Mark $s$ as explored and set $s' \gets s$
  \For {each neighbor $n$ of $s$ in $\set{G}$}
    \If {$n$ is unexplored and $d_\text{JS}(f \!\parallel\! f_\theta(n))<\epsilon$}
      \State $s' \gets s' \cup \text{\Call{Merge}{$n$, $f$; $\mathcal{G}$, $\theta$}}$
    \EndIf
  \EndFor
  \State \Return $s'$
\EndProcedure
\end{algorithmic}
\label{algorithm2}
\end{algorithm}

\smallskip\noindent\textbf{Acquisition function.} 
From the merged superpixels $\set{S}_t$, 
we then select a batch $\set{B}_t \subset \set{S}_t$
of size $B$ to be labeled, according to an acquisition function
that estimates the benefit from labeling a merged superpixel,
where the benefit would be huge for uncertain superpixels of rare class labels.
In what follows, we define an uncertainty measure of superpixel
in \eqref{eq:uncertainty}
and a popularity estimate of class in \eqref{eq:size-aware-class-balance}, and then introduce an acquisition function in \eqref{acquisition_function}.

Recalling $f_\theta(x) \! \in \! \mathbb{R}^{|\set{C}|}$
is the probability such that $f_\theta(c; x)$ is the estimated probability
that the class $c$ of pixel $x$,
we adapt best-versus-second-best~\cite{joshi2009multi}
for uncertainty measures of pixel $x$ and superpixel $s$ as follows:
% which disregards low probabilities of insignificant classes:
\begin{align}
u_\theta(x) &:= \frac{\max_{c \in \mathcal{C} \setminus\{y_\theta(x)\}} {f_\theta(c; x)}}{\max_{c \in \set{C} }f_\theta(c;x)}\;, \\
u_\theta(s) &:= \frac{\sum_{x \in s} u_\theta(x)}{|\{x: x \in s\}|} \;, 
\label{eq:uncertainty}
\end{align}
% \begin{equation}
% u_\theta(x)\!:=\!\frac{\underset{c \in \mathcal{C} \setminus\{y_\theta(x)\}}{\text{max}} {f_\theta(c; x)}}{\max_{c \in \set{C} }f_\theta(c;x)},
% u_\theta(s) \!:=\! \frac{\sum_{x \in s} u_\theta(x)}{|\{x: x \in s\}|}, 
% \label{eq:uncertainty}
% \end{equation}
where $y_{\theta}(x) := \argmax_{c \in \set{C}} f_\theta(c;x)$ is the estimated dominant label of pixel $x$ in a given model $\theta$.
% and $u_\theta(s)$ is the uncertainty of superpixel $s$.
% To reduce the time complexity, we merge superpixels in descending of the uncertainty in Algorithm.~\ref{algorithm2}.
% The merging order based on high uncertainty is related to the time complexity, which we describe later.
% Each denominator and numerator indicate the best and the second-best value of the feature $f_\theta(x)$, respectively.
% Similar to the class distributions of a superpixel $s$ in \ref{eq:feature-of-superpixel}, we define the uncertainty as: 
% \begin{equation}
% u_\theta(s) := \frac{\sum_{x \in s} u_\theta(x)}{|s|}\;. 
% \end{equation}
% From the merged nodes, we generate adaptive superpixels $\mathcal{S}_t$ for round $t$.
% We next use an acquisition function to select $b$ superpixels from $\mathcal{S}_t$ to obtain labels.

We then define a popularity estimate $p(c; \theta)$ of class $c \in \set{C}$
given $\theta$ as follows: 
%  we calculate the posterior of the class distribution based on the number of pixels that make up superpixels:
\begin{equation}
p(c;\theta) := \frac{|\{ x : \exists s \in \mathcal{S}_t, \text{D}_\theta(s) = c, x \in s\}|}{|\{x : \exists s \in \mathcal{S}_t, x \in s \}|} \;,
\label{eq:size-aware-class-balance}
\end{equation}
where $\text{D}_\theta(s) :=\argmax_{c \in \mathcal{C}} | \{x \in s : y_\theta(x) = c \} |$ is the majority of predicted labels in superpixel $s$. We note that low $p(c;\theta)$ implies that class $c$ is rare in the prediction of $\theta$. It is noteworthy that we compute the class popularity
in pixel-level due to the various sizes of our merged superpixels,
while the previous work \cite{cai2021revisiting}
proposes a superpixel-wise class popularity, 
$\frac{|\{ s : \text{D}_\theta(s) = c, s \in \mathcal{S}_t\}|}{|\{s : s \in \mathcal{S}_t \}|} \;$, assuming superpixels of uniform size.

Using the uncertainty $u_\theta(s)$ in \eqref{eq:uncertainty}
and the class popularity $p(c;\theta)$ in \eqref{eq:size-aware-class-balance}, 
we define the following acquisition function 
$a(s; \theta)$
prioritizing uncertain superpixels of rare classes:
\begin{equation}
a(s; \theta):= u_\theta(s) \exp \big({-p(\text{D}_\theta(s) ; \theta)} \big)\;.
\label{acquisition_function}
\end{equation}
We select $B$ superpixels of highest values of $a(s; \theta_{t-1})$
from the merged $\set{S}_t$ for query batch $\set{B}_t$.

\smallskip\noindent\textbf{Remarks.}
We note that it is possible to produce $\set{S}_t$ from scratch
rather than from base segmentation $\set{S}_0$.
To reduce the computational cost for the adaptive merging process,
we however compose $\set{S}_t$ by merging base superpixels in $\set{S}_0$
from SEEDS, which is known to generate an over-segmentation of semantics.
Moreover, it is computationally expensive to explore all the possible mergers
and obtain $\set{S}_t$ followed by the query selection.
We hence conduct the merging process only for a certain portion of base superpixels with the highest values of uncertainty (c.f., line~\ref{line:order} in Algorithm~\ref{algorithm2}) and then
select $\set{B}_t$ to be queried since the acquisition function would select 
merged superpixels of high uncertainty in the end.
% Details are in the appendix.
Further details are presented in Appendix~\ref{sec:rationale-merging}.

% To reduce the time complexity of converting superpixels into a graph, we explore and merge superpixels with high uncertainty. Details are in the appendix.

\subsection{Sieving}
\label{sec:sieving-technique}
% After obtaining the dominant label on batch $\mathcal{B}_t$,

% % The dataset for round $t$ is represented as follows:
% % \begin{equation}
% % \!\!\!\! \mathcal{D}_{t} := \{(x, y) : x \in s \in \bigcup_{i \in [t]} \mathcal{B}_{i}, y(c) = \mathbbm{1} [c = \text{D}(s)] \} \;, 
% % % ... 
% % % (s_{t \times b}, \text{D}(s_{t \times b})), 
% % % \text{  where  } s \in \bigcup_{i \in [t]} \mathcal{B}_{t} \}\;,
% % \end{equation}
% % which consists of all superpixels selected prior to round $t$.

% We sieve

% Then, training.
% analog to ... 

% \smallskip\noindent\textbf{Adaptive sieving}

% \smallskip\noindent\textbf{Training.} 

% %$\text{D}(s)$ indicates a dominant label from ground-truth.

Despite the sophisticated design of the adaptive merging,
a queried superpixel can inevitably include 
pixels of classes different from the dominant one,
in particular, as we select superpixels of which model predictions are unsure. Hence, the dominant labeling is liable to make noisy annotations.
% Even when we collect the dominant labels from an oracle, who always tells truth, it is possible that 
% the labeling is noisy for pixels 
% Furthermore, our merging algorithm occasionally combines superpixels with different dominant labels, which can exacerbate the label noise.
% While previous studies have addressed noisy labels in segmentation \cite{acuna2019devil, liu2022adaptive, oh2021background}, these methods deal with a different type of noise than that caused by the imperfection of superpixels.
To alleviate such side effects of the dominant labeling, we propose a simple sieving technique that 
filter out pixels that have high potential risks of being 
different classes than the dominant one.
We observe that
for a queried superpixel $s$ and given model $\theta$, 
the risk of %noisy labeling on pixel $x$
mismatch between the dominant label $D(s)$ and the true label of pixel $x \in s$
would be high when $f_\theta \big( \text{D}(s); x \big)$
is low.
From this observation, we define 
\begin{equation}
h(s;\theta) := \{ x \in s: f_\theta \big( \text{D}(s); x \big) \geq \phi(s; \theta) \} \;,
\label{eq:sieving}
\end{equation}
% \begin{equation}
% \tilde s := \{ x \in s: f_\theta \big( \text{D}(s); x \big) \geq k(s) \} \;,
% \end{equation}
where 
%for superpixel $s$ given model $\theta$,
$\phi(s; \theta)$ is a knee point of the cumulative distribution function of values of $f_\theta \big( \text{D}(s); x \big)$ in superpixel $s$, detected by
Kneedle algorithm~\cite{satopaa2011finding}.
 In addition, 
the knee point detection allows us to have a tailored sieving threshold to each superpixel. This is important to avoid the case that the remained pixels are heavily biased to relatively easy labels after sieving.
Further details are in Appendix \ref{fig:sup-sieving}.
% Further details and examples of knee points are in the appendix.

We revisit all the queried superpixel $s \in \bigcup_{t'=0}^{t} \mathcal{B}_{t'}$ and sieve them using \eqref{eq:sieving}
with the latest model $\theta_{t-1}$
since the model evolves round by round.
%This enables that the sieving can evolve over rounds as model $\theta$ improves.
We finally obtain the following sieved dataset ${\mathcal{D}}_{t}$ for round $t \ge 1$:
\begin{equation}
{\mathcal{D}}_{t} := \left\{(x, y) : 
\begin{aligned}
&\exists s \in \cup_{t'=0}^{t} \mathcal{B}_{t'}, \ x \in h(s;\theta_{t-1}), \\
&y(c) = \mathbbm{1}{[c = \text{D}(s)]} \ \forall c\in\set{C} 
\end{aligned}
\right\} \;.
\label{sieved-superpixel}
\end{equation}
Analogously to the warm-up round, 
initializing model $\theta$ at a model pretrained on ImageNet, we obtain $\theta_t$ trained to mainly minimize the following CE loss:
% \begin{equation}
% \mathcal{L}_{\text{CE}} = - \frac{1}{|\tilde{\mathcal{D}}_{[t]}|} \sum_{(\tilde{s}, y_s) \in \tilde{\mathcal{D}}_{[t]}} \sum_{x \in \tilde{s}} y_s \log p(y \mid x, \theta_t) \;,
% \label{cross-entropy}
% \end{equation}
\begin{equation}
\hat{\mathbb{E}}_{(x,y) \sim {\mathcal{D}}_t} [ \text{CE}(y, f_\theta (x))] \;.
\label{eq:final-loss}
\end{equation}

\section{Experiments}
% \begin{table}
%   \centering
%   \begin{tabular}{@{}lc@{}}
%     \toprule
%     Method & Budgets \\
%     \midrule
%     EntropyBox+ & 10.25$\%$ \\
%     MetaBox+    & 10.47$\%$ \\
%     SEEDS+AL & 7.85$\%$ \\
%     EAM+AL & \\
%     \midrule 
%     Oracle & \\
%     Ours & \\
%     \bottomrule
%   \end{tabular}
%   \caption{Budgets to obtain 95\% accuracy on Cityscapes}
%   \label{tab:example}
% \end{table}

\subsection{Experimental setup}
\label{para:oracle-superpixels}
\iffalse
\smallskip\noindent\textbf{Evaluation datasets.}
%To control various constraints including the size of superpixels and annotation budgets,
We use two widely known semantic segmentation datasets: Cityscapes \cite{Cordts2016Cityscapes} and PASCAL VOC 2012 (PASCAL) \cite{pascal-voc-2012}. 
Cityscapes consists of 2,975 training and 500 validation images with 19 classes, while PASCAL consists of 1,464 training and 1,449 validation images with 20 classes.
Regardless of the resolution of images, we set the average superpixel size, i.e., the number of pixels in superpixels, to 256 for all experiments except for one where we adjust the size.
After training a model with the superpixel selected by an acquisition function, we evaluate the mean Intersection-over-Union \cite{everingham2015pascal} of the model on the validation images.
\fi
\smallskip\noindent\textbf{Datasets.}
We use two semantic segmentation datasets: Cityscapes~\cite{Cordts2016Cityscapes} and PASCAL VOC 2012 (PASCAL)~\cite{pascal-voc-2012}.
Cityscapes comprises 2,975 training and 500 validation images with 19 classes, while PASCAL consists of 1,464 training and 1,449 validation images with 20 classes.

\begin{figure*}[t!]
    % \captionsetup[subfigure]{font=footnotesize,labelfont=footnotesize,aboveskip=0.05cm,belowskip=-0.15cm}
    \centering
    \hspace{-5mm}
    \begin{subfigure}{.23\linewidth}
        \centering
        \begin{tikzpicture}
            \begin{axis}[
                legend style={nodes={scale=0.6}, at={(2.15, 1.16)}},
                legend columns=-1,
                xlabel={The number of clicks},
                ylabel={mIoU (\%)},
                width=1.23\linewidth,
                height=1.23\linewidth,
                ymin=62.5,
                ymax=74.6,
                ytick={64, 66, 68, 70, 72, 74},
                xlabel style={yshift=0.15cm},
                ylabel style={yshift=-0.2cm},
                xmin=90,
                xmax=260,
                label style={font=\scriptsize},
                tick label style={font=\scriptsize},
                xticklabel={$\pgfmathprintnumber{\tick}$k}
            ]
            % Oracle
            \addplot[cCL, very thick, mark=pentagon*, mark size=2pt, mark options={solid}] table[col sep=comma, x=x, y=oracle-avg]{Data/limited_budget_cityscapes.csv};
            % AM-SP
            \addplot[cdeepBP, very thick, mark=diamond*, mark size=2pt, mark options={solid}] table[col sep=comma, x=x, y=am-sp-avg]{Data/limited_budget_cityscapes.csv};
            % M-SP
            \addplot[cdeepBP32, very thick, mark=square*, mark size=2pt, mark options={solid}] table[col sep=comma, x=x, y=m-sp-avg]{Data/limited_budget_cityscapes.csv};
            % SP
            \addplot[cdeepMF, very thick, mark=triangle*, mark size=2pt, mark options={solid}] table[col sep=comma, x=x, y=revisiting-avg]{Data/limited_budget_cityscapes.csv};
            % M-SP
            \addplot[cdeepBP32, very thick, mark=square*, mark size=2pt, mark options={solid}] table[col sep=comma, x=x, y=m-sp-avg]{Data/limited_budget_cityscapes.csv};
            % AM-SP
            \addplot[cdeepBP, very thick, mark=diamond*, mark size=2pt, mark options={solid}] table[col sep=comma, x=x, y=am-sp-avg]{Data/limited_budget_cityscapes.csv};
            % Oracle
            \addplot[cCL, very thick, mark=pentagon*, mark size=2pt, mark options={solid}] table[col sep=comma, x=x, y=oracle-avg]{Data/limited_budget_cityscapes.csv};
            
            % Oracle
            \addplot[name path=oracle-l, draw=none, fill=none] table[col sep=comma, x=x, y=oracle-l]{Data/limited_budget_cityscapes.csv};
            \addplot[name path=oracle-u, draw=none, fill=none] table[col sep=comma, x=x, y=oracle-u]{Data/limited_budget_cityscapes.csv};
            \addplot[cCL, fill opacity=0.15] fill between[of=oracle-l and oracle-u];
            % SP
            \addplot[name path=revisiting-l, draw=none, fill=none] table[col sep=comma, x=x, y=revisiting-l]{Data/limited_budget_cityscapes.csv};
            \addplot[name path=revisiting-u, draw=none, fill=none] table[col sep=comma, x=x, y=revisiting-u]{Data/limited_budget_cityscapes.csv};
            \addplot[cdeepMF, fill opacity=0.15] fill between[of=revisiting-l and revisiting-u]; 
            % M-SP
            \addplot[name path=m-sp-l, draw=none, fill=none] table[col sep=comma, x=x, y=m-sp-l]{Data/limited_budget_cityscapes.csv};
            \addplot[name path=m-sp-u, draw=none, fill=none] table[col sep=comma, x=x, y=m-sp-u]{Data/limited_budget_cityscapes.csv};
            \addplot[cdeepBP32, fill opacity=0.15] fill between[of=m-sp-l and m-sp-u]; 
            % AM-SP
            \addplot[name path=am-sp-l, draw=none, fill=none] table[col sep=comma, x=x, y=am-sp-l]{Data/limited_budget_cityscapes.csv};
            \addplot[name path=am-sp-u, draw=none, fill=none] table[col sep=comma, x=x, y=am-sp-u]{Data/limited_budget_cityscapes.csv};
            \addplot[cdeepBP, fill opacity=0.15] fill between[of=am-sp-l and am-sp-u]; 
            
            \legend{Oracle, AMSP+S (Ours), MSP+S, SP~\cite{cai2021revisiting}}
            \end{axis}
        % \node[above] at (3.7, 3.41) {\small Performance for varying budget};
        \end{tikzpicture}
        \caption{Cityscapes}
        \label{fig:(a)-effect}
    \end{subfigure}
    \hspace{1mm}    
    \begin{subfigure}{.23\linewidth}
        \centering
        \begin{tikzpicture}
            \begin{axis}[
                legend style={nodes={scale=0.35}, at={(0.03, 0.24)}, anchor=west}, 
                xlabel={The number of clicks},
                ylabel={mIoU (\%)},
                width=1.23\linewidth,
                height=1.23\linewidth,
                ymin=58,
                ymax=70.1,
                ytick={60, 62, 64, 66, 68, 70},
                xlabel style={yshift=0.15cm},
                ylabel style={yshift=-0.2cm},
                legend columns=2,
                xmin=9,
                xmax=26,
                label style={font=\scriptsize},
                tick label style={font=\scriptsize},
                xticklabel={$\pgfmathprintnumber{\tick}$k}
            ]
            % Oracle
            \addplot[cCL, very thick, mark=pentagon*, mark size=2pt, mark options={solid}] table[col sep=comma, x=x, y=oracle]{Data/limited_budget_pascal.csv};
            % M-SP
            \addplot[cBP, very thick, mark=square*, mark size=2pt, mark options={solid}] table[col sep=comma, x=x, y=m-sp]{Data/limited_budget_pascal.csv};
            % SP
            \addplot[cdeepMF, very thick, mark=triangle*, mark size=2pt, mark options={solid}] table[col sep=comma, x=x, y=revisiting]{Data/limited_budget_pascal.csv};
            % M-SP
            \addplot[cdeepBP32, very thick, mark=square*, mark size=2pt, mark options={solid}] table[col sep=comma, x=x, y=m-sp]{Data/limited_budget_pascal.csv};
            % AM-SP
            \addplot[cdeepBP, very thick, mark=diamond*, mark size=2pt, mark options={solid}] table[col sep=comma, x=x, y=am-sp]{Data/limited_budget_pascal.csv};
            
            % Oracle
            \addplot[name path=oracle-l, draw=none, fill=none] table[col sep=comma, x=x, y=oracle-l]{Data/limited_budget_pascal.csv};
            \addplot[name path=oracle-u, draw=none, fill=none] table[col sep=comma, x=x, y=oracle-u]{Data/limited_budget_pascal.csv};
            \addplot[cCL, fill opacity=0.15] fill between[of=oracle-l and oracle-u];
            % SP
            \addplot[name path=revisiting-l, draw=none, fill=none] table[col sep=comma, x=x, y=revisiting-l]{Data/limited_budget_pascal.csv};
            \addplot[name path=revisiting-u, draw=none, fill=none] table[col sep=comma, x=x, y=revisiting-u]{Data/limited_budget_pascal.csv};
            \addplot[cdeepMF, fill opacity=0.15] fill between[of=revisiting-l and revisiting-u];
            % M-SP
            \addplot[name path=m-sp-l, draw=none, fill=none] table[col sep=comma, x=x, y=m-sp-l]{Data/limited_budget_pascal.csv};
            \addplot[name path=m-sp-u, draw=none, fill=none] table[col sep=comma, x=x, y=m-sp-u]{Data/limited_budget_pascal.csv};
            \addplot[cdeepBP32, fill opacity=0.15] fill between[of=m-sp-l and m-sp-u];            
            % AM-SP
            \addplot[name path=am-sp-l, draw=none, fill=none] table[col sep=comma, x=x, y=am-sp-l]{Data/limited_budget_pascal.csv};
            \addplot[name path=am-sp-u, draw=none, fill=none] table[col sep=comma, x=x, y=am-sp-u]{Data/limited_budget_pascal.csv};
            \addplot[cdeepBP, fill opacity=0.15] fill between[of=am-sp-l and am-sp-u]; 
            
            \end{axis}
        % \node[above] at (1.7, 3.35) {\footnotesize PASCAL VOC 2012};s
        \end{tikzpicture}
        \caption{PASCAL}
        \label{fig:(b)-effect}
    \end{subfigure}
    \hspace{1mm}
    \begin{subfigure}{.23\linewidth}
        \centering
        \begin{tikzpicture}
            \begin{axis}[
                legend style={nodes={scale=0.6}, at={(2.1, 1.16)}},
                legend columns=-1,
                xlabel={Size of base superpixels},
                ylabel={mIoU (\%)},
                width=1.23\linewidth,
                height=1.23\linewidth,
                ymin=54.2,
                ymax=68.6,
                ytick={54, 56, 58, 60, 62, 64, 66, 68, 70},
                xlabel style={yshift=0.15cm},
                ylabel style={yshift=-0.2cm},
                xmin=0.7,
                xmax=4.3,
                label style={font=\scriptsize},
                tick label style={font=\scriptsize},
                xtick=data,
                xticklabels={64,256,1024,4096},
            ]
            % Oracle
            \addplot[cCL, very thick, mark=pentagon*, mark size=2pt, mark options={solid}] table[col sep=comma, x=x, y=oracle]{Data/region_size_cityscapes.csv};
            % AM-SP
            \addplot[cdeepBP, very thick, mark=diamond*, mark size=2pt, mark options={solid}] table[col sep=comma, x=x, y=am-sp]{Data/region_size_cityscapes.csv};
            % S-SP
            \addplot[cMV, very thick, mark=*, mark size=2pt, mark options={solid}] table[col sep=comma, x=x, y=s-sp]{Data/region_size_cityscapes.csv};
            % SP
            \addplot[cdeepMF, very thick, mark=triangle*, mark size=2pt, mark options={solid}] table[col sep=comma, x=x, y=sp] {Data/region_size_cityscapes.csv};
            % S-SP
            \addplot[cMV, very thick, mark=*, mark size=2pt, mark options={solid}] table[col sep=comma, x=x, y=s-sp]{Data/region_size_cityscapes.csv};
            % AM-SP
            \addplot[cdeepBP, very thick, mark=diamond*, mark size=2pt, mark options={solid}] table[col sep=comma, x=x, y=am-sp]{Data/region_size_cityscapes.csv};

            % Oracle
            \addplot[name path=oracle-l, draw=none, fill=none] table[col sep=comma, x=x, y=oracle-l]{Data/region_size_cityscapes.csv};
            \addplot[name path=oracle-u, draw=none, fill=none] table[col sep=comma, x=x, y=oracle-u]{Data/region_size_cityscapes.csv};
            \addplot[cCL, fill opacity=0.15] fill between[of=oracle-l and oracle-u];
            % SP
            \addplot[name path=sp-l, draw=none, fill=none] table[col sep=comma, x=x, y=sp-l]{Data/region_size_cityscapes.csv};
            \addplot[name path=sp-u, draw=none, fill=none] table[col sep=comma, x=x, y=sp-u]{Data/region_size_cityscapes.csv};
            \addplot[cdeepMF, fill opacity=0.15] fill between[of=sp-l and sp-u];
            % S-SP
            \addplot[name path=s-sp-l, draw=none, fill=none] table[col sep=comma, x=x, y=s-sp-l]{Data/region_size_cityscapes.csv};
            \addplot[name path=s-sp-u, draw=none, fill=none] table[col sep=comma, x=x, y=s-sp-u]{Data/region_size_cityscapes.csv};
            \addplot[cMV, fill opacity=0.15] fill between[of=s-sp-l and s-sp-u];
            % AM-SP
            \addplot[name path=am-sp-l, draw=none, fill=none] table[col sep=comma, x=x, y=am-sp-l]{Data/region_size_cityscapes.csv};
            \addplot[name path=am-sp-u, draw=none, fill=none] table[col sep=comma, x=x, y=am-sp-u]{Data/region_size_cityscapes.csv};
            \addplot[cdeepBP, fill opacity=0.15] fill between[of=am-sp-l and am-sp-u];
            
            \legend{Oracle,AMSP+S (Ours),SP+S,SP~\cite{cai2021revisiting}}
            \end{axis}
        % \node[above] at (3.6, 3.34) {\small Performance for varying superpixel size};
        \end{tikzpicture}
        \caption{Cityscapes}
        \label{fig:(c)-effect}
    \end{subfigure}
    \hspace{1mm}    
    \begin{subfigure}{.23\linewidth}
        \centering
        \begin{tikzpicture}
            \begin{axis}[
                legend style={nodes={scale=0.35}, at={(0.03, 0.24)}, anchor=west}, 
                xlabel={Size of base superpixels},
                ylabel={mIoU (\%)},
                width=1.23\linewidth,
                height=1.23\linewidth,
                ymin=53.8,
                ymax=66.6,
                ytick={54, 56, 58, 60, 62, 64, 66},
                xlabel style={yshift=0.15cm},
                ylabel style={yshift=-0.2cm},
                legend columns=2,
                xmin=0.7,
                xmax=4.3,
                label style={font=\scriptsize},
                tick label style={font=\scriptsize},
                xtick=data,
                xticklabels={4,16,64,256},
            ]
            \addplot[cdeepMF, very thick, mark=triangle*, mark size=2pt, mark options={solid}] table[col sep=comma, x=x, y=sp] {Data/region_size_pascal.csv};
            \addplot[cMV, very thick, mark=*, mark size=2pt, mark options={solid}] table[col sep=comma, x=x, y=s-sp]{Data/region_size_pascal.csv};
            \addplot[cdeepBP, very thick, mark=diamond*, mark size=2pt, mark options={solid}] table[col sep=comma, x=x, y=am-sp]{Data/region_size_pascal.csv};
            \addplot[cCL, very thick, mark=pentagon*, mark size=2pt, mark options={solid}] table[col sep=comma, x=x, y=oracle]{Data/region_size_pascal.csv};

            % SP
            \addplot[name path=sp-l, draw=none, fill=none] table[col sep=comma, x=x, y=sp-l]{Data/region_size_pascal.csv};
            \addplot[name path=sp-u, draw=none, fill=none] table[col sep=comma, x=x, y=sp-u]{Data/region_size_pascal.csv};
            \addplot[cdeepMF, fill opacity=0.15] fill between[of=sp-l and sp-u];
            % S-SP
            \addplot[name path=s-sp-l, draw=none, fill=none] table[col sep=comma, x=x, y=s-sp-l]{Data/region_size_pascal.csv};
            \addplot[name path=s-sp-u, draw=none, fill=none] table[col sep=comma, x=x, y=s-sp-u]{Data/region_size_pascal.csv};
            \addplot[cMV, fill opacity=0.15] fill between[of=s-sp-l and s-sp-u];
            % AM-SP
            \addplot[name path=am-sp-l, draw=none, fill=none] table[col sep=comma, x=x, y=am-sp-l]{Data/region_size_pascal.csv};
            \addplot[name path=am-sp-u, draw=none, fill=none] table[col sep=comma, x=x, y=am-sp-u]{Data/region_size_pascal.csv};
            \addplot[cdeepBP, fill opacity=0.15] fill between[of=am-sp-l and am-sp-u]; 
            % Oracle
            \addplot[name path=oracle-l, draw=none, fill=none] table[col sep=comma, x=x, y=oracle-l]{Data/region_size_pascal.csv};
            \addplot[name path=oracle-u, draw=none, fill=none] table[col sep=comma, x=x, y=oracle-u]{Data/region_size_pascal.csv};
            \addplot[cCL, fill opacity=0.15] fill between[of=oracle-l and oracle-u];
            % % AM-SP
            % \addplot[name path=am-sp-l, draw=none, fill=none] table[col sep=comma, x=x, y=am-sp-l]{Data/region_size_pascal.csv};
            % \addplot[name path=am-sp-u, draw=none, fill=none] table[col sep=comma, x=x, y=am-sp-u]{Data/region_size_pascal.csv};
            % \addplot[cdeepBP, fill opacity=0.15] fill between[of=am-sp-l and am-sp-u];
            % % S-SP
            % \addplot[name path=s-sp-l, draw=none, fill=none] table[col sep=comma, x=x, y=s-sp-l]{Data/region_size_pascal.csv};
            % \addplot[name path=s-sp-u, draw=none, fill=none] table[col sep=comma, x=x, y=s-sp-u]{Data/region_size_pascal.csv};
            % \addplot[cMV, fill opacity=0.15] fill between[of=s-sp-l and s-sp-u];
            % % S-SP
            % \addplot[name path=s-sp-l, draw=none, fill=none] table[col sep=comma, x=x, y=s-sp-l]{Data/region_size_pascal.csv};
            % \addplot[name path=s-sp-u, draw=none, fill=none] table[col sep=comma, x=x, y=s-sp-u]{Data/region_size_pascal.csv};
            % \addplot[cMV, fill opacity=0.15] fill between[of=s-sp-l and s-sp-u];
            \end{axis}
        \end{tikzpicture}
        \caption{PASCAL}
        \label{fig:(d)-effect}
    \end{subfigure}
    % \caption{{\em Effect of adaptive superpixels.} (a, b) As the round progresses, \textit{AM-SP} shows better performance than \textit{M-SP}. (c, d) \textit{AM-SP} is robust to the size of superpixels, while \textit{SP} is sensitive. Experiments are conducted three times.}
    \caption{{\em Effect of adaptive superpixels.} (a, b) mIoU versus the number of clicks as budget. (c, d) mIoU versus the size of base superpixels. Each experiment is conducted with three trials and the shaded region indicates ranges.}
    % \hsh{(a, b) mIoU versus varying budgets as the amount of clicks. (c, d) mIoU versus the size of superpixels. Each experiment is conducted with three trials and the shaded region indicates ranges.
    \label{fig:robustness}
    % \vspace{-2mm}
\end{figure*}
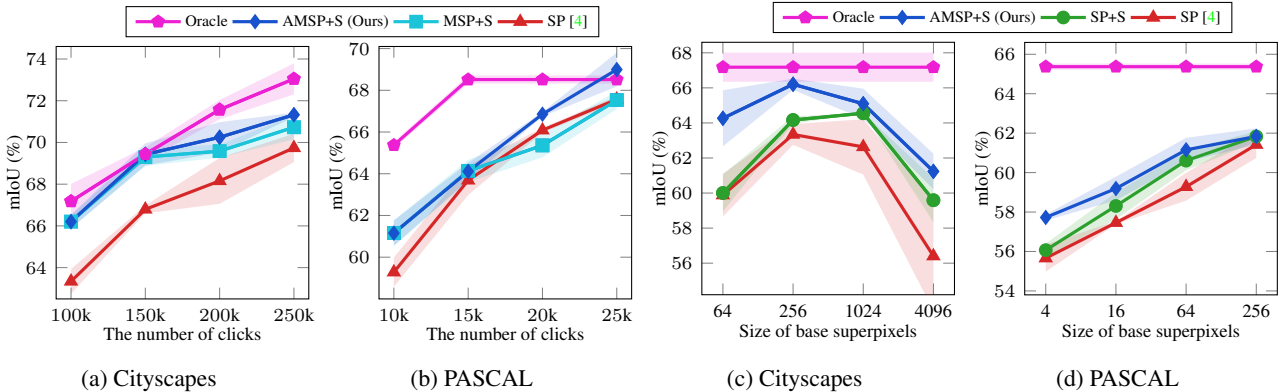

\smallskip\noindent\textbf{Implementation details.}
We adopt DeepLab-v3+ architecture with Xception-65~\cite{chen2018encoder} as our segmentation backbone.
During training, we use the SGD optimizer with a momentum of 0.9 and set a base learning rate to 7e-3.
We decay the learning rate by polynomial decay with a power of 0.9.
For Cityscapes, we resize training images to 769 $\times$ 769 and train a model for 60k iterations with a mini-batch size of 4.
Similarly, for PASCAL, we resize training images to 513 $\times$ 513 and train a model for 30k iterations with a mini-batch size of 12.
% We set $\epsilon$ to 0.1 for all quantitative experiments except for the one where we adjust it.
Unless specified, we set the value of $\epsilon$ to 0.1.

\iffalse
\smallskip\noindent\textbf{Baseline methods.} 
We compare our adaptive merging algorithm to the state-of-the-art in superpixel-based active learning \cite{cai2021revisiting} called \textit{SP}, the abbreviation of superpixel.
Our algorithm comprises two main components: a sieving technique and a merging algorithm.
We refer to the method that uses only the sieving technique as \textit{SP+S}, while the one that employs both sieving and adaptive merging is called \textit{AMSP+S}.
Furthermore, we compare our adaptive merging approach with \textit{MSP+S}, which performs merging only in the second round.
Note that \textit{MSP+S} and \textit{AMSP+S} are identical until round 2.
\fi

\smallskip\noindent\textbf{Baseline methods.} 
We compare our algorithm to \textit{SP} \cite{cai2021revisiting}, the state-of-the-art superpixel-based active segmentation method.
Our algorithm applies two proposed processes in each round: merging and sieving.
We call our complete method including adaptive merging as \textit{AMSP+S}, while the partial version that only uses the sieving without the merging is called \textit{SP+S}.
Additionally, we evaluate the modified version of our method that performs merging only once in the second round, called \textit{MSP+S}.
% Furthermore, we compare our adaptive merging approach with \textit{MSP+S}, which performs merging only in the second round.
% We refer to the method that uses only the sieving technique as \textit{SP+S}, while the one that employs both sieving and adaptive merging is called \textit{AMSP+S}.
% Furthermore, we compare our adaptive merging approach with \textit{MSP+S}, which performs merging only in the second round.
Note that \textit{AMSP+S} and \textit{MSP+S} are identical until the second round.

\iffalse
\smallskip\noindent\textbf{Oracle superpixels.}
To measure the maximum achievable performance of a model, we introduce an oracle superpixel baseline named \textit{Oracle}.
Usually, the performance of a model trained with full pixel-wise annotations of all images is considered the upper bound.
However, the fully supervised baseline is inconsistent with active learning since it disregards budget and round limitations.
Therefore, we suggest an oracle superpixel baseline that leverages ground truth as superpixels illustrated in Figure \ref{fig:(d)_adaptive_merged_superpixels}.
For example, if a car is obstructed by a pole and divided into two parts, it is treated as two superpixels.
The Cityscapes and PASCAL datasets contain 408K and 16k oracle superpixels, respectively.
It is worth noting that the PASCAL has a lower number of oracle superpixels due to the smaller number of classes per image.
For a fair comparison, we select the same number of superpixels per round from the oracle superpixels.
As a round progresses, all oracle superpixels are selected, and our oracle superpixel baseline is equal to the fully supervised baseline.
We report 95\% accuracy for the oracle baseline for comparable comparison.
\fi

\smallskip\noindent\textbf{Oracle baseline.}
The adaptive superpixel aims to merge every connected region with the same class labels.
% \omh{(It might be understood as referring to Ground-truth or Pseudo Label in Table \ref{tab:merging-methods})} 
Thus, the upper bound of it is to consider each region separated by the ground truth mask as a superpixel.
We refer to such ideal regions as oracle superpixels in Figure~\ref{(c)-qualitative}.
An active learning model trained using the oracle superpixels is called \textit{Oracle}.
Details are in Appendix \ref{fig:sup-oracle}.
% A detailed analysis of an oracle baseline is in the appendix.
As the number of oracle superpixels is limited, all of them are eventually labeled as the round progresses, and the performance of the trained model becomes equivalent to that of the pixel-wise fully supervised model.
We report 100\% and 90\% of the \textit{Oracle} performance for Cityscapes and PASCAL, respectively.
% Following the convention of the previous research \cite{cai2021revisiting}, we report 95\% of the \textit{Oracle} performance.

\smallskip\noindent\textbf{Evaluation protocol.}
% Regardless of the resolution of the images,
We set the average size of the superpixels to 256 and 64 pixels on Cityscapes and PASCAL, respectively, for all experiments except for one where we adjust the size.
% In the first round, we randomly select superpixels to train a model, ensuring that all methods, except for the \textit{Oracle}, start at the same point.
% In the following rounds, we adaptively merge superpixels and choose superpixels with the acquisition function in \eqref{acquisition_function}.
Following \textit{SP}~\cite{cai2021revisiting}, we use the number of clicks as the labeling budget.
We conduct 5 rounds of data sampling, where we allocate a budget of 50k and 5k for each round on Cityscapes and PASCAL, respectively.
% We allocate a budget of 50k and 5k for each round on Cityscapes and PASCAL, respectively.
% Once we obtain the dominant labels of the selected superpixels, we sieve the labels.
% We finally train a model with the sieved dataset and evaluate the mean Intersection-over-Union \cite{everingham2015pascal} of the model on the validation images.
In the first round, we randomly select superpixels to train a model, ensuring that all methods start at the same performance.
We evaluate the trained model with mean Intersection-over-Union~\cite{everingham2015pascal} on the validation images.
We emphasize that the average size of superpixels containing 64 pixels is more efficient on Cityscapes, as detailed in Appendix~\ref{sec:base-superpixel-sizes}.

\subsection{Effect of adaptive superpixels}
\label{sec:effect-of-adaptive}
\iffalse
\smallskip\noindent\textbf{Multi-round scenario.}
Active learning typically allocates its budgets across multiple rounds, with performance gradually enhancing as we identify useful superpixels in each round, guided by the current model.
Given a total budget of 250k and 50k for Cityscapes and PASCAL, respectively, we spend budgets equally over 5 rounds.
In the first round, we train a model by randomly selecting over-segmented superpixels common to all methods, and from the second round, the superpixels used by each method change.
A baseline \textit{SP} continues to use the over-segmented superpixels, our \textit{MSP+S} uses the merged superpixels generated with the model trained in round 1, and our \textit{AMSP+S} uses adaptively merged superpixels in every round.
Here, both our methods applied a sieving technique.
For an upper bound, we compare with \textit{Oracle} utilizing oracle superpixels.
In Figure \ref{fig:multi-rounds}, we observe that our 
\fi

\begin{figure*}[t!]
    % \captionsetup[subfigure]{font=footnotesize}
    \centering
    \begin{subfigure}{.33\linewidth}
        \centering
        \includegraphics[scale=0.322]{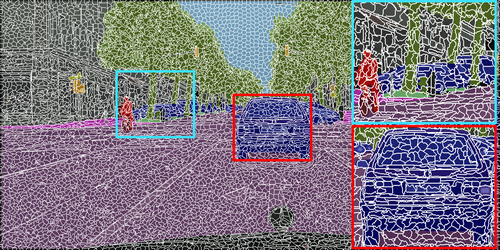}
        % \caption{$\text{ASA}(S;G)=0.021, \; \text{AF}(G;S)=0.355$}
        % \vspace{2mm}
        % \label{(a)-qualitative}
    \end{subfigure}
    \begin{subfigure}{.33\linewidth}
        \centering
        \includegraphics[scale=0.322]{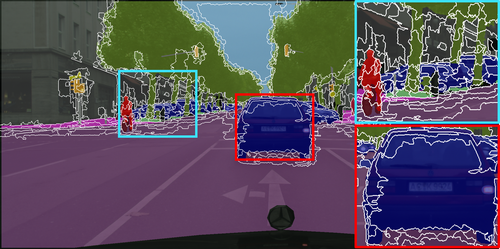}
        % \caption{$\text{ASA}(S;G)=0.89, \; \text{AF}(G;S)=0.283$}
        % \vspace{2mm}
        % \label{(b)-qualitative}
    \end{subfigure}
    \begin{subfigure}{.33\linewidth}
        \centering
        \includegraphics[scale=0.322]{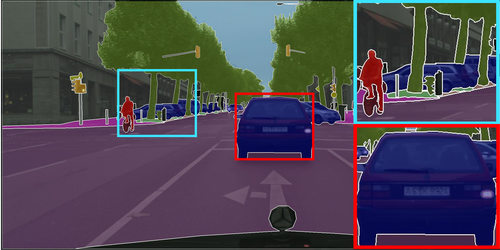}
        % \caption{$ASA(S;G)=1.00, \; AF(G;S)=1.00$}
        % \vspace{2mm}
        % \label{(c)-qualitative}
    \end{subfigure}
    \begin{subfigure}{.33\linewidth}
        \centering
        \includegraphics[scale=0.322]{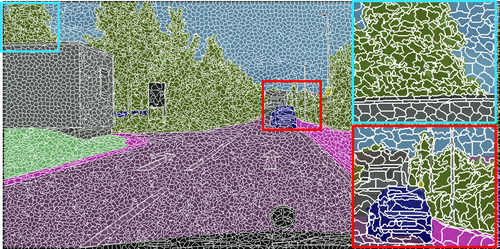}
        \caption{Base superpixels~\cite{van2012seeds}}
        \label{(a)-qualitative}
    \end{subfigure}
    \begin{subfigure}{.33\linewidth}
        \centering
        \includegraphics[scale=0.322]{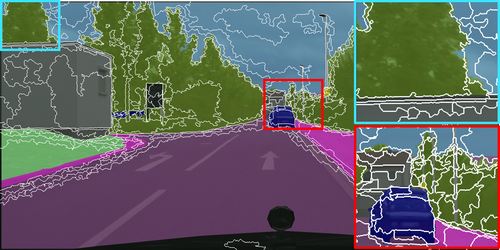}
        \caption{Merged superpixels (Ours)}
        \label{(b)-qualitative}
    \end{subfigure}
    \begin{subfigure}{.33\linewidth}
        \centering
        \includegraphics[scale=0.322]{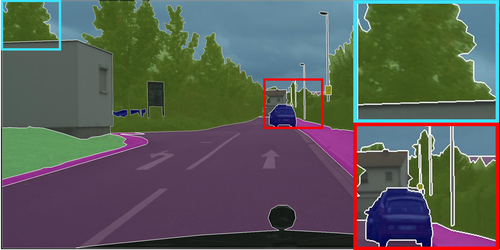}
        % \caption{$ASA(S;G)=1.00, \; AF(G;S)=1.00$}
        \caption{Oracle superpixels}
        \label{(c)-qualitative}
    \end{subfigure}
    % \caption{{\em Qualitative results of adaptive superpixels.} As the round progresses, (a) over-segmented superpixels becomes (b) adaptively merged ones, and they resemble (c) oracle superpixels, especially for the classes that the model is confident about.}
    \caption{{\em Qualitative results of adaptive superpixels.} (a) Base superpixel generated by SEEDS~\cite{van2012seeds} with size 256. (b) Superpixels generated with proposed adaptive merging at round 4. (c) Oracle superpixels generated from the ground truth.}
    \label{fig:qualitative}
    % \vspace{-3mm}
\end{figure*}

% \vspace{-3mm}
\iffalse
\smallskip\noindent\textbf{Multi-round scenario.}
Active learning typically allocates its budgets across multiple rounds, with performance gradually enhancing as we identify useful superpixels in each round.
Following the previous work \cite{cai2021revisiting}, we evaluate the performance during five rounds.
From the second round, the superpixels used by each method change.
A baseline \textit{SP} continues to use the over-segmented superpixels, \textit{MSP+S} uses the merged superpixels generated with the model trained in the first round, and \textit{AMSP+S} uses adaptively merged superpixels in every round.
For an upper bound, we compare with \textit{Oracle} utilizing oracle superpixels.
In Figures~\ref{fig:(a)-effect} and \ref{fig:(b)-effect}, we observe the performance improvement of adaptive merging.
\fi
\smallskip\noindent\textbf{Multi-round scenario.}
In Figures~\ref{fig:(a)-effect} and \ref{fig:(b)-effect}, we compare the performance of the proposed method to \textit{SP}~\cite{cai2021revisiting} varying budget for both of Cityscapes and PASCAL.
Note that the performance for round 0, \ie., 50K budget, is omitted as each method has the same performance at the warm-up round.
The results show that our adaptive superpixel (\textit{AMSP+S}) clearly outperforms the previous art in every budget setting on both of the datasets.
In particular, the \textit{AMSP+S} with only 150k clicks outperforms the previous art with 250k clicks in Cityscapes.
In the final round, the proposed method recovers 97\% and 92\% of the \textit{Oracle} performance for Cityscapes and PASCAL, respectively.
% Active learning typically allocates its budgets across multiple rounds, with performance gradually enhancing as we identify useful superpixels in each round.
% Following the previous work \cite{cai2021revisiting}, we evaluate the performance during five rounds.
% From the second round, the superpixels used by each method change.
% A baseline \textit{SP} continues to use the over-segmented superpixels, \textit{MSP+S} uses the merged superpixels generated with the model trained in the first round, and \textit{AMSP+S} uses adaptively merged superpixels in every round.
% For an upper bound, we compare with \textit{Oracle} utilizing oracle superpixels.
% In Figures~\ref{fig:(a)-effect} and \ref{fig:(b)-effect}, we observe the performance improvement of adaptive merging.
To show the effectiveness of our adaptive approach, we compare \textit{AMSP+S} to its one-shot merging version \textit{MSP+S} in Figures~\ref{fig:(a)-effect} and \ref{fig:(b)-effect}.
On both datasets, adaptive feature of \textit{AMSP+S} shows performance gain especially for the last two rounds.
The experiments conducted for additional rounds can be found in Appendix~\ref{sec:base-superpixel-sizes}.

\smallskip\noindent\textbf{Multi-size scenario.}
% The size of superpixels is an essential hyperparameter in superpixel-based AL, affecting both the quantity and quality of labels, \eg, a decrease in size results in fewer labels, and an increase in size leads to lower quality labels.
The size of superpixels is an essential hyperparameter in superpixel-based AL, affecting both the quantity and quality of labels.
% In superpixel-based AL, the size of superpixel affects both the quantity and quality of labels, where a decrease in size results in fewer labels, and an increase in size leads to lower quality labels.
In Figures~\ref{fig:(c)-effect} and \ref{fig:(d)-effect}, we compare the proposed method to \textit{SP}~\cite{cai2021revisiting} varying the base superpixel size for both of Cityscapes and PASCAL, in the second round.
Our adaptive superpixel (\textit{AMSP+S}) outperforms the previous art in various superpixel sizes on both of the datasets.
% We also evaluate , which shows that our adaptive merging is especially effective for small superpixels and our sieving is especially effective for large superpixels.
We also evaluate sieving only version (\textit{SP+S}) of our method, which quantifies contribution of each components in our method.
The performance improvement between \textit{SP} and \textit{SP+S} shows our sieving is especially helpful for large superpixels, and the performance gap between \textit{SP+S} and \textit{AMSP+S} shows our merging is especially effective for small superpixels.
Thanks to the proposed sieving and merging, \textit{AMSP+S} are comparably robust to the change of the superpixel size than \textit{SP}.
% To evaluate the sensitivity to the choice of the superpixel size, we train models with different superpixel sizes in the second round.
% Initially, we train a model using 50k and 5k superpixels randomly selected from Cityscapes and PASCAL, respectively, in round 1.
% We then experiment with different superpixel sizes in round 2 with the same budget used in round 1.
% Figures \ref{fig:(a)-robustness} and \ref{fig:(b)-robustness} demonstrate that our merged superpixels are robust, while the baseline is sensitive to size.

\smallskip\noindent\textbf{Qualitative results.}
The quality of the proposed adaptive superpixel is illustrated in Figure~\ref{fig:qualitative}.
As shown in Figure~\ref{(a)-qualitative}, superpixels used in the previous study \cite{cai2021revisiting} have uniform sizes for all areas regardless of their content.
In contrast, Figure~\ref{(b)-qualitative} demonstrates that adaptive superpixels accurately reflect the actual size of the content in images, carefully capturing small object classes while efficiently covering large background classes.
% As the round increases, the quality of the adaptive superpixel increases, and gradually approaches the oracle superpixel depicted in Fig. \ref{fig:(d)_adaptive_merged_superpixels}.
% As shown in Fig. \ref{(b)-adaptive} and \ref{(c)-adaptive}, as the round increases, 
% As shown in Figure 1b-c, as the round increases, the quality of the adaptive superpixel increases, and gradually approaches the oracle superpixel drawn on figure 1d.
More examples are in Appendix \ref{fig:sup-qual}.
% as shown in Fig. \ref{(b)-adaptive} and \ref{(c)-adaptive}, adaptive superpixel reflects the actual size of the content in images, delicately capturing small object classes while efficiently covering large background classes.
% Figure \ref{fig:adaptive_merged_superpixels} shows our adaptive superpixels along with their corresponding oracle superpixels that server as the ground-truth.
% In contrast to previous work \cite{cai2021revisiting}, which employs a constant number of superpixels for all images, our merged superpixels vary in number, depending on the objects within each image.
% As round progresses, the distribution of the number of superpixels becomes increasingly similar to that of the oracle superpixel.
% 
% 
% 
% size vs noise ratio

% A typical metric F1-score is calculated as the harmonic mean of Precision and Recall as:
% \begin{equation}
% \text{F1-score}(S, G) = \frac{2}{\frac{1}{\text{Precision}(S, G)} + \frac{1}{\text{Recall}(S, G)}} \;.
% \end{equation}

% \subsection{Robustness to Region Size}
% Round 1, color, proposed algorithm, transform feature embedding space into color, xy embedding space, 
% Round 2, feature, round 1 fixed

%% Don;t normalize?
%% Scale font size.
% \begin{figure*}[!]
%     \centering
%     \includegraphics[width=\textwidth]{Figures/matching-region-distribution.pdf}
%     \caption{Changing distribution of regions at each cycle of active learning. The region distribution converges to that of ground-truth annotations during training.}
%     \label{fig:matching-region-distribution}
% \end{figure*}

\section{Analyses of adaptive superpixels}
\iffalse
We propose novel achievable metrics to explore the suitable superpixels for active learning (Section~\ref{sec:confusion-matrix}).
% These metrics aid in the understanding of suitable superpixels in active learning.
In addition, we conduct quantitative analyses of adaptive superpixels under varying $\epsilon$ (Section~\ref{sec:merging-criteria}).
All analyses are conducted on Cityscapes with an average superpixel size of 256 pixels.
% , and propose novel evaluation metrics that leverages the concept of confusion matrix between superpixels (Sec. \ref{sec:confusion-matrix}). 
% These metrics aid in the understanding of suitable superpixels in active learning.
% All analyses are conducted on Cityscapes images with an average superpixel size of 256.
\fi
We propose new evaluation metrics to measure the quality of superpixel as a labeling unit for active segmentation, and utilize it to analyze our adaptive superpixels (Section~\ref{sec:confusion-matrix}).
We also conduct analyses about the effect $\epsilon$ to our adaptive superpixels (Section~\ref{sec:merging-criteria}).
All analyses are conducted on Cityscapes with an average superpixel size of 256 pixels.

\begin{table*}[!t]
  \centering
  \setlength\tabcolsep{4pt}
  \begin{tabular}{l|cc|ccc|ccc|c}
    \toprule
    Methods & ASA$(S;G)$ & ASA$(G;S)$ & AP$(S;G)$ & AR$(S;G)$ & AF$(S;G)$ & AP$(G;S)$ & AR$(G;S)$ & AF$(G;S)$ & mIoU \\ 
    \midrule
    $\text{SLIC}_{4096}$ & 0.887 & 0.082 & 0.897 & 0.046 & 0.066 & 0.695 & 0.259 & 0.185 & 53.18 \\
    $\text{SEEDS}_{4096}$ & 0.909 & 0.082 & 0.900 & 0.050 & 0.070 & 0.665 & 0.309 & 0.221 & 57.61 \\
    $\text{SLIC}_{256}$ & 0.956 & 0.013 & 0.958 & 0.007 & 0.012 & 0.400 & 0.622 & 0.278 & 58.04 \\
    $\text{SEEDS}_{256}$ & 0.961 & 0.014 & 0.960 & 0.007 & 0.012 & 0.395 & 0.647 & 0.297 & 58.97 \\
    \rowcolor{Gray}
    $\text{Merged}_2$ & 0.898 & 0.515 & 0.883 & 0.042 & 0.063 & 0.553 & 0.472 & 0.333 & \underline{60.00} \\
    \rowcolor{Gray}
    $\text{Merged}_4$ & 0.898 & 0.496 & 0.883 & 0.042 & 0.062 & 0.548 & 0.484 & 0.340 & \textbf{61.36} \\
    \midrule
    $\text{Merged}^*$ & 0.899 & 0.597 & 0.880 & 0.045 & 0.066 & 0.547 & 0.510 & 0.359 & 61.85 \\
    Oracle & 1.000 & 1.000 & 1.000 & 1.000 & 1.000 & 1.000 & 1.000 & 1.000 & 70.81 \\
    \bottomrule
  \end{tabular}
  % \caption{{\em Evaluation metrics of superpixels.} Superpixels are generated using SLIC \cite{achanta2012slic} and SEEDS \cite{van2012seeds}, with the subscript indicating the average size of superpixels. Our merged superpixels are evaluated, with the subscript value implying the round that used the superpixels and * representing full supervision. To compute the mIoU, we train a model with 100k randomly selected superpixels.}
  \caption{{\em Evaluation metrics of superpixels.}
  % Superpixel quality is measured by various evaluation metrics, where each superpixel is generated from  SLIC~\cite{achanta2012slic}, SEEDS~\cite{van2012seeds}, our adaptive merging (Merged), and the ground-truth (Oracle).
  The subscript indicates the average size of the superpixel for SLIC~\cite{achanta2012slic} and SEEDS~\cite{van2012seeds}, while it indicates the round for Merged.
  $\text{Merged}^*$ indicates superpixel merged by a model trained with full supervision.
  To compute the mIoU, we train a model with 100k randomly selected superpixels.}
  \label{tab:quantitative}
  % \vspace{-3mm}
\end{table*}

\subsection{Achievable metrics}
\label{sec:confusion-matrix}
\iffalse
Superpixels are evaluated by various metrics, but to our best knowledge, all metrics are related to the degree of over-segmentation.
For instance, achievable segmentation accuracy (ASA) \cite{liu2011entropy} measures the segmentation accuracy when each superpixel $s \in S$ is assigned to the corresponding dominant label, which is the ground-truth of the oracle superpixel $g \in G$ with the highest overlap. The ASA is calculated as follows:
\begin{equation}
\text{ASA}(S; G) := \frac{\sum_{s \in S} \max_g |s \cap g|}{\sum_{s \in S} |s|} \;,
\end{equation}
where $S$ and $G$ represent the generated and oracle superpixels from the same image, respectively. 
As an image becomes more over-segmented, \ie., the superpixel size becomes smaller, the ASA value increases. 
However, active learning aims to achieve the maximum benefit with the least amount of labeling effort, and therefore, the number of labels should be taken into account. 
Additionally, the ASA is heavily biased towards classes with a relatively large number of pixels.
\fi
While various evaluation metrics for superpixel are presented~\cite{giraud2017robust,liu2011entropy,machairas2014waterpixels,schick2012measuring,van2012seeds}, most of them aims to measure the quality of over-segmentation.
For instance, achievable segmentation accuracy (ASA) \cite{liu2011entropy} measures
the segmentation accuracy when each superpixel $s \in S$ is associated with the oracle superpixel with the largest overlap.
% the segmentation precision when each superpixel $s \in S$ is associated with the oracle superpixel with the largest overlap.
The ASA is calculated as follows:
\begin{equation}
\text{ASA}(S; G) := \frac{\sum_{s \in S} \max_{g \in G} |s \cap g|}{\sum_{s \in S} |s|} \;,
\end{equation}
where $S$ and $G$ represent the generated and oracle superpixels from the same image, respectively. 
As an image becomes more over-segmented, \ie., the superpixel size becomes smaller, the ASA value increases. 
However, active learning (AL) aims to achieve the maximum benefit with the least amount of labeling effort, and therefore, the number of labels should be taken into account. 
In addition, the ASA is heavily biased towards classes with a large number of pixels.

In order to measure the suitability of superpixels for AL, we introduce precision and recall between generated and oracle superpixels. 
A generated superpixel can be viewed as positive on the inside and negative on the outside, and its precision and recall with respect to the corresponding oracle one can be calculated.
For all generated superpixels, we define the achievable precision (AP) as follows:
\begin{equation}
\text{AP}(S;G) := \frac{1}{|S|} \sum_{s \in S} \frac{\max_{g \in G} |s \cap g|}{|s|} \;,
\end{equation}
where the summation is performed in superpixels, unlike in ASA, which implies pixel-wise precision.
As we put the same weight on each superpixel, AP is less influenced by large objects than ASA.
We note that AP is different to average precision \cite{everingham2009pascal, salton1983introduction}, used in object detection, which utilize the precision and recall curve.
We also define the achievable recall (AR) and F1-score (AF) as:
\begin{equation}
\text{AR}(S;G) := \frac{1}{|S|} \sum_{s \in S} \frac{\max_{g \in G} |s \cap g|}{|g'(s; G)|} \;, 
\end{equation}
\begin{equation}
\text{AF}(S;G) := \frac{2}{|S|} \sum_{s \in S} \frac{\max_{g \in G} |s \cap g|}{|s| + |g'(s; G)|} \;,
\end{equation}
where $g'(s; G) := \argmax_{g \in G} \left|s \cap g \right|$ 
%(\omh{$g'_j := \argmax_{g_j\in G} \left|s_i \cap g_j\right|$})
refers to the corresponding oracle superpixel.
Details are in Appendix \ref{fig:sup-metrics}.

\begin{figure}[!t]
    % \captionsetup[subfigure]{font=scriptsize,labelfont=scriptsize,aboveskip=0.05cm,belowskip=-0.15cm}
    \centering
    \hspace{-5mm}
    \begin{subfigure}{.47\linewidth}
        \centering
        \begin{tikzpicture}
            \begin{axis}[
                legend style={nodes={scale=0.35}, at={(0.03, 0.24)}, anchor=west}, 
                xlabel={ASA$(S;G)$},
                ylabel={mIoU (\%)},
                width=1.23\linewidth,
                height=1.23\linewidth,
                ymin=50.8,
                ymax=63.2,
                ytick={51, 53, 55, 57, 59, 61, 63},
                xlabel style={yshift=0.15cm},
                ylabel style={yshift=-0.2cm},
                legend columns=2,
                xmin=0.882,
                xmax=0.967,
                label style={font=\scriptsize},
                tick label style={font=\scriptsize},
                x tick label style={
                    /pgf/number format/.cd,
                        fixed,
                }
            ]
            \addplot[cdeepBP, only marks] table[col sep=comma, x=ASASG, y=mIoU]{Data/correlation_asasg.csv};
            \addplot[very thick, orange] table[col sep=comma, x=ASASG, y={create col/linear regression = {y=mIoU}}
            ]{Data/correlation_asasg.csv};
            \draw (0.5\linewidth, 0.35\linewidth) node {\scriptsize$\text{Corr} = 0.05$};
            \end{axis}
        \end{tikzpicture}
        % \caption{ASA(S;G) vs mIoU}
    \end{subfigure}
    \hspace{1mm}
    \begin{subfigure}{.47\linewidth}
        \centering
        \begin{tikzpicture}
            \begin{axis}[
                legend style={nodes={scale=0.35}, at={(0.03, 0.24)}, anchor=west}, 
                xlabel={AF$(G;S)$},
                ylabel={mIoU (\%)},
                width=1.23\linewidth,
                height=1.23\linewidth,
                ymin=50.8,
                ymax=63.2,
                xlabel style={yshift=0.15cm},
                ylabel style={yshift=-0.2cm},
                ytick={51, 53, 55, 57, 59, 61, 63},
                legend columns=2,
                xmin=0.174,
                xmax=0.371,
                label style={font=\scriptsize},
                tick label style={font=\scriptsize},
                x tick label style={
                    /pgf/number format/.cd,
                        fixed,
                }
            ]
            \addplot[cdeepBP, only marks] table[col sep=comma, x=AFGS, y=mIoU]{Data/correlation_afgs.csv};
            \addplot[very thick, orange] table[col sep=comma, x=AFGS, y={create col/linear regression = {y=mIoU}}
            ]{Data/correlation_afgs.csv};
            \draw (0.5\linewidth, 0.35\linewidth) node {\scriptsize$\text{Corr}=0.95$};
            \end{axis}
        \end{tikzpicture}
    \end{subfigure}
    \caption{{\em Relationship between metrics and mIoU.} The correlation between ASA$(S;G)$ and mIoU is low, while the correlation between AF$(G;S)$ and mIoU is high. For the correlation calculation, \textit{Oracle} in Table \ref{tab:quantitative} is excluded.}
    % \caption{{\em Relationship between metrics and mIoU.}
    % \hsh{mIoU of actively learned model using superpixels versus the quality of the corresponding superpixels measured with ASA~\cite{liu2011entropy} and proposed achievable F1-score (AF).}}
    \label{fig:co-relation}
    % \vspace{-4mm}
\end{figure}
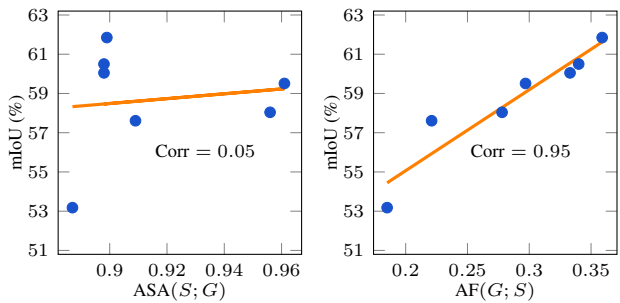

All the metrics evaluate generated superpixels in comparison to oracle ones. 
However, the size of superpixels is also important besides their quality in AL.
Therefore, it is necessary to evaluate the oracle superpixels against the generated superpixels, \ie., ASA$(G;S$), AP$(G;S)$, AR$(G;S)$ and AF$(G;S)$. We hence propose AF$(G;S)$
%which is the harmonic mean of   AP$(G;S)$, AR$(G;S)$, 
defined as:
\begin{equation}
\text{AF}(G;S) := \frac{2}{|G|} \sum_{g \in G} \frac{\max_{s \in S} |g \cap s|}{|g| + |s'(g; S)|} \;,
\end{equation}
where $s'(g; S) := \argmax_{s \in S} \left|g \cap s \right|$ refers to the generated superpixel with the highest overlap, which is linked to the maximum amount of labeling we receive.
Table~\ref{tab:quantitative} evaluates various superpixels through eight metrics.
Although our merged superpixels have a relatively low ASA$(S;G)$, they exhibit high ASA$(G;S)$ and AF$(G;S)$.

\smallskip\noindent\textbf{Correlation of metrics and mIoU.}
\iffalse
The selection of a suitable superpixel algorithm or superpixels is crucial for superpixel-based active learning. 
Given several available superpixels as shown in Table~\ref{tab:quantitative}, we randomly choose 100k superpixels to train a model and measure the resulting mIoU.
As observed by the strong correlation between AF$(G;S)$ and mIoU in Figure~\ref{fig:co-relation}, one can utilize superpixels with high AF$(G;S)$ values for the initial pool in active learning.
Further details are provided in the appendix.
\fi
To show the proposed metric can accurately evaluate the superpixel quality for active segmentation, we measure the correlation between various evaluation metric and the performance of actively learned model in Table~\ref{tab:quantitative} and Figure~\ref{fig:co-relation}.
We observe that the proposed AF$(G;S)$ shows the highest correlation to the performance of the actively learned model.
We except AF$(G;S)$ can select suitable superpixel algorithm for active learning, where the details are provided in Appendix \ref{fig:sup-metrics}.
% and determine a $\epsilon$ for merging in our algorithm with some validation images. 

% Average Precision (AP)
% \begin{equation}
% \text{AP}(G;S) : \frac{1}{|G|} \sum_{g_i \in G} \frac{\max_j |g_i \cap s_j|}{|g_i|} \;,
% \end{equation}

% ASA$(S;G)$ is an important metric in the literature of superpixel generation, however, it is 
% we note that the objective of active learning is to achieve the greatest gain with a single click.
% Thus, we suggest a measure of how many regions are labeled with a single click for each oracle superpixel $g$ as:
% \begin{equation}
% \text{ASA}(G ; S) = \frac{\sum_{g_j \in G} \max_i |s_i \cap g_j|}{\sum_{g_j \in G} |g_j|} \;.
% \end{equation}
% Note that both metrics have the same denominator.
% For a simplicity, we suggest the average of ASA (AASA) as:
% \begin{equation}
% \text{AASA}(S, G) = \frac{\text{ASA}(S;G) + \text{ASA}(G;S)}{2} \;.
% \end{equation}
% In Table 

\subsection{Ablation studies on epsilon} 
% Ablation studies on epsilon
\label{sec:merging-criteria}

\begin{figure}[!t]
    % \captionsetup[subfigure]{font=footnotesize,labelfont=footnotesize,aboveskip=0.05cm,belowskip=-0.15cm}
    \centering
    \hspace{-3mm}
    \begin{subfigure}{.47\linewidth}
        \centering
        \begin{tikzpicture}
            \begin{axis}[
                legend style={nodes={scale=0.5}, at={(0.48, 0.3)}, anchor=west}, 
                xlabel={$\epsilon$},
                ylabel={mIoU (\%)},
                width=1.23\linewidth,
                height=1.23\linewidth,
                ymin=62.4,
                ymax=67.2,
                ytick={62, 63, 64, 65, 66, 67},
                xlabel style={yshift=0.15cm},
                ylabel style={yshift=-0.2cm},
                legend columns=1,
                xmin=0.03,
                xmax=0.22,
                label style={font=\scriptsize},
                tick label style={font=\scriptsize},
                x tick label style={
                    /pgf/number format/.cd,
                        fixed,
                }
            ]
            \addplot[cdeepBP, very thick, mark=diamond*, mark size=2pt, mark options={solid}] table[col sep=comma, x=x, y=ours]{Data/wrong_merging_cityscapes.csv};
            \addplot[cdeepMF, very thick, mark=triangle*, mark size=2pt, mark options={solid}] table[col sep=comma, x=x, y=revisiting] {Data/wrong_merging_cityscapes.csv};
            \legend{AMSP+S,SP}
            \end{axis}
        \end{tikzpicture}
        \caption{Cityscapes}
        \label{fig:(c)-robustness}
    \end{subfigure}
    \hspace{1mm}
    \begin{subfigure}{.47\linewidth}
        \centering
        \begin{tikzpicture}
            \begin{axis}[
                legend style={nodes={scale=0.35}, at={(0.03, 64)}, anchor=west}, 
                xlabel={$\epsilon$},
                ylabel={mIoU (\%)},
                width=1.23\linewidth,
                height=1.23\linewidth,
                ymin=60.8,
                ymax=63.7,
                ytick={61, 62, 63, 64},
                xlabel style={yshift=0.15cm},
                ylabel style={yshift=-0.2cm},
                legend columns=2,
                xmin=0.02,
                xmax=0.22,
                label style={font=\scriptsize},
                tick label style={font=\scriptsize},
                x tick label style={
                    /pgf/number format/.cd,
                        fixed,
                }
            ]
            \addplot[cdeepMF, very thick, mark=triangle*, mark size=2pt, mark options={solid}] table[col sep=comma, x=x, y=revisiting] {Data/wrong_merging_pascal.csv};
            \addplot[cdeepBP, very thick, mark=diamond*, mark size=2pt, mark options={solid}] table[col sep=comma, x=x, y=ours]{Data/wrong_merging_pascal.csv};
            \end{axis}
        \end{tikzpicture}
        \caption{PASCAL}
        \label{fig:(d)-robustness}
    \end{subfigure}
    \caption{{\em Epsilon sensitivity.} We experiment on superpixels with varying $\epsilon$ and demonstrate the robustness of our \textit{AMSP+S}, while \textit{SP} is independent of $\epsilon$. Each experiment is conducted on the second round.}
    % \caption{{\em Epsilon sensitivity.} \hsh{mIoU of the proposed method and the baseline varying hyper-parameter $\epsilon$.
    % Each experiment is conducted on the second round.}}
    \label{fig:multi-rounds}
    % \vspace{-2mm}
\end{figure}
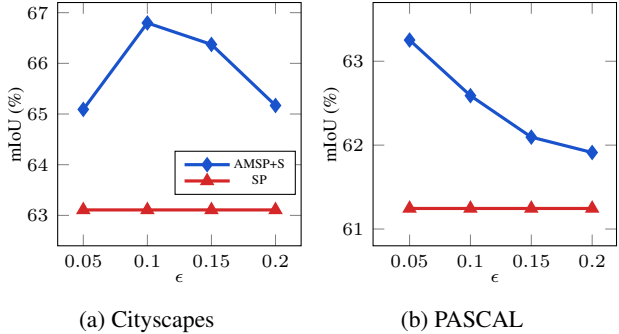

\smallskip\noindent\textbf{Epsilon sensitivity.}
\iffalse
We evaluate the sensitivity of our method to the choice of $\epsilon$, which determines the amount of the merging process.
% Furthermore, our algorithm generates distinct superpixels depending on the value of $\epsilon$. 
% Similar to the superpixel size, a large epsilon increases the number of labels by boosting the average size of the superpixels, while a small epsilon improves the quality of the labels by reducing the number of incorrect merges. 
In Figures~\ref{fig:(c)-robustness} and \ref{fig:(d)-robustness}, we observe that our merging approach not only exhibits robustness to $\epsilon$ but also outperforms the baseline over a wide range of $\epsilon$.
\fi
In Figures~\ref{fig:(c)-robustness} and \ref{fig:(d)-robustness}, we evaluate the sensitivity of our method to $\epsilon$, which determines the amount of the merging.
Proposed method show robustness to the change of $\epsilon$, where the change of mIoU is less than 2\% for both Citycapes and PASCAL when $\epsilon$ is between 0.05 and 0.2.
We observe that for every investigated $\epsilon$ values, \textit{AMSP+S} surpasses the performance of the previous art~\cite{cai2021revisiting}.

\smallskip\noindent\textbf{Adaptive epsilon.}
We fix $\epsilon$ to 0.10 in all quantitative experiments, but there may exist an optimal $\epsilon$ for each round. 
% To ensure a fair comparison, we exclude validation images from the experiments. 
% Nonetheless, 
% We analyze $\epsilon$ that maximizes the proposed AF$(G;S)$ metric for each round by assuming the existence of 10 validation images with fully pixel-wise annotations.
% As the round progresses, the improvement of the model enables us to merge aggressively, which is related to the increase in $\epsilon$ that maximizes AF$(G;S)$ over the rounds in Table \ref{tab:adaptive-epsilon}.
In Table \ref{tab:adaptive-epsilon}, we analyze $\epsilon$ that maximizes AF$(G;S)$ metric for each round by assuming the existence of 10 validation images with ground truth. 
As the round increases, the optimal $\epsilon$ increases as well, which implies that the improvement of the model enables us to merge aggressively.

\smallskip\noindent\textbf{Effect of epsilon.}
Table \ref{tab:merging-methods} presents the quality of the merging algorithm under various criteria, defining correctness based on the agreement of dominant labels in paired superpixels.
We merge a pair of superpixels when their ground-truth label is identical (Ground Truth), when their dominant top-1 model prediction is identical (Pseudo Label), and when the Euclidean Distance (ED) or Jensen-Shannon Divergence (JSD) of their averaged predictive probability is smaller than~$\epsilon$.
% 
% It is apparent that when using the ground truth, all superpixels are merged correctly.
% However, using pseudo labels leads to lower-quality merging as it ignores other minor classes.
Using pseudo labels leads to lower-quality merging as it ignores other minor classes.
% Since we utilize the class distribution as a feature space, 
Since we utilize the predicted class probability as a feature space, 
% in \eqref{eq:feature-space}, 
JSD proves to be more effective than ED.
As $\epsilon$ increases, the correct ratio decreases due to the aggressive merging of superpixels.
We emphasize that $\epsilon$ can determine the trade-off between the quantity and quality of labels.

\subsection{Implementation remarks for practitioners}
\smallskip\noindent\textbf{Fast merging.}
% \subsection{Rationale for line~\ref{line:order} of Algorithm \ref{algorithm2}}
\label{fig:sup-descend}
% \begin{table}[!ht]
% \centering
% \setlength\tabcolsep{6pt}
% \begin{tabular}{l|c}
% \toprule
% Methods & mIoU \\ \midrule
% \textit{SP} \cite{cai2021revisiting} & 63.77 \\ \midrule
% \textit{AMSP+S} (ascending, 10\%) & 64.33 \\ \midrule
% \textit{AMSP+S} (descending, 10\%) & \underline{65.99} \\ \midrule
% \rowcolor{Gray}
% \textit{AMSP+S} (descending, 100\%) & \textbf{66.53} \\ \midrule
% \bottomrule
% \end{tabular}
% \caption{{\em Various merging order.} Experiments are conducted on Cityscapes dataset with an average superpixel size of 256, using 100k costs for two rounds.}
% \label{tab:descending}
% \end{table}
% We utilize a graph to merge superpixels by converting them into nodes and edges.
% However,
% We explain the rationale behind traversing nodes in the descending order of uncertainty in line~\ref{line:order} of Algorithm \ref{algorithm2}.
% We first convert superpixels into a graph for merging, where
The completion of the merging process for an image 
essentially requires
a linear time complexity in the number of the base superpixels.
However, to reduce this,
the complete merging can be 
replaced with a {\it partial merging}
that scans only a subset of base superpixels with high uncertainties
as roots.
This is considerable since 
we will eventually query only a subset of the merged superpixels 
according to 
the acquisition function~\eqref{acquisition_function},
which prioritizes those with high uncertainties.
In Table~\ref{tab:descending}, we
compare the complete and partial mergings in terms of the mIoU at 100k clicks.
As expected, the partial merging on  base superpixels with
top-10\% uncertainty
has only a small gap to the complete merging.
In addition, 
we find that by employing the partial merging, the time complexity is reduced significantly by a factor of 25.98
\footnote{
The per-image runtime of the merging process (CPU-intensive) for Cityscapes is reduced from 12.42s to 0.48s
on a server with two AMD EPYC 7513 32-core processors.
}.
The partial merging is a useful suggestion to save computation resource for practitioners.
% \khy{
A further investigation and discussion on the partial merging are presented in
Appendix~\ref{sec:rationale-merging}.

\begin{table}[!t]
\centering
\setlength\tabcolsep{6pt}
\begin{tabular}{l|ccccc}
\toprule
Epsilon & 0.04   & 0.05 & 0.06 & 0.07 & 0.08 \\ \midrule 
$\text{Merged}_1$ & 0.344 & \textbf{0.346} & 0.344 & 0.340 & 0.336 \\
$\text{Merged}_2$ & 0.347 & 0.346 & \textbf{0.348} & 0.345 & 0.344 \\
$\text{Merged}_3$ & 0.346 & 0.349 & 0.350 & \textbf{0.351} & 0.349 \\
$\text{Merged}_4$ & 0.347 & 0.347 & 0.347 & \textbf{0.348} & 0.346 \\
\bottomrule
\end{tabular}
\caption{{\em Adaptive epsilon.} AF$(G;S)$ for adaptive superpixels generated by varying $\epsilon$ is reported. The subscript indicates the round.} 
\label{tab:adaptive-epsilon}
% \vspace{-5mm}
\end{table}

\begin{table}[!t]
\centering
\setlength\tabcolsep{6pt}
\begin{tabular}{l|c|cc}
\toprule
Method                             & Epsilon & Correct & Incorrect \\ \midrule
Ground Truth                       &    -    & 1.000   & 0.000       \\ \midrule
Pseudo Label                 &    -    & 0.832 & 0.168   \\ \midrule
\multirow{3}{*}{ED}        & 0.05    & 0.915        & 0.085      \\
                                           & 0.10    & 0.901        & 0.099      \\
                                           & 0.15    & 0.891        & 0.109      \\ \midrule
\multirow{3}{*}{JSD} & 0.05    & 0.934        & 0.066      \\
                                           & 0.10    & 0.911        & 0.089      \\
                                           & 0.15    & 0.896        & 0.104      \\
\bottomrule
\end{tabular}
\caption{{\em Various merging criteria.} Using JSD performs more accurate merging than using ED. As $\epsilon$ increases, the rate of incorrect merging increases. }
%\omh{For each method and $\epsilon$, superpixel-meging results are evaluated by the ratio of superpixels that are merged correctly or incorrectly.}
% \caption{{\em Various merging criteria.}
% \hsh{Quality of the superpixel generated from merging algorithm, where a pair of superpixel are merged when their ground-truth label is identical (Ground-truth), when their top-1 model prediction is identical (Pseudo Label), and when the euclidean distance (ED) or jensan-shannon distance (JSD) of their feature is smaller than $\epsilon$.}}
\label{tab:merging-methods}
% \vspace{-5mm}
\end{table}

\begin{table}[!t]
\centering
\setlength\tabcolsep{6pt}
\begin{tabular}{l|c}
\toprule
Methods & mIoU \\ \midrule
\textit{SP}  \cite{cai2021revisiting} & 63.77 \\ \midrule
% \textit{AMSP+S} (bottom 10\%) & 64.33 \\ \midrule
\textit{AMSP+S} (top 10\%) & \underline{65.99} \\ \midrule
\rowcolor{Gray}
\textit{AMSP+S} (complete 100\%) & \textbf{66.53} \\ \midrule
\bottomrule
\end{tabular}
\caption{{\em Various levels of partial merging.} Experiments are conducted under the same setting of Figure~\ref{fig:(a)-effect} with 100k clicks (Cityscapes, superpixel size of 256).}
\label{tab:descending}
\end{table}

\smallskip\noindent\textbf{Compatibility with other base superpixels.}
% \begin{table}[!t]
% \centering
% \setlength\tabcolsep{6pt}
% \begin{tabular}{l|l|c}
% \toprule
% Superpixel algorithm & Method & mIoU \\ \midrule
% \multirow{2}{*}{SEEDS} & \textit{SP}~\cite{cai2021revisiting} & 63.77 \\
%                        & \textit{AMSP+S} (Ours) & \textbf{66.53} \\ \midrule
% \multirow{2}{*}{SLIC}  & \textit{SP}~\cite{cai2021revisiting} & 65.97 \\
%                        & \textit{AMSP+S} (Ours) & \textbf{67.56} \\
% \bottomrule
% \end{tabular}
% \caption{{\em Other base superpixels.} 
% Experiments are carried out under the identical settings, as presented in Table \ref{tab:sieving}.}
% \label{tab:slic}
% \end{table}
For a fair comparison to the previous study~\cite{cai2021revisiting}, we have employed the same superpixel algorithm called SEEDS~\cite{van2012seeds} to generate base superpixels at the beginning.
% However, our algorithm also works with other choices of base superpixels since our merging and sieving processes can be applied on top of any base superpixels. 
However, our merging and sieving processes can be applied on top of any other base superpixels.
Indeed, in Figure~\ref{fig:(c)-effect} (and Appendix~\ref{sec:base-superpixel-sizes}), 
our method consistently shows gains over SP~\cite{cai2021revisiting} when using base superpixels of different sizes.
Furthermore, 
we compare SP and AMSP+S (ours)
when using SLIC instead of SEEDS
in the same setting of Figure~\ref{fig:(a)-effect},
where 
the mIoU's at 100k clicks of
SP and ours have 65.97\% and 67.56\%, respectively,
% our algorithm with different
% under the same setting in Figure~\ref{fig:(a)-effect}, we conduct experiments with SLIC~\cite{achanta2012slic} instead of SEEDS, where the mIoU's at 100k clicks are Ours+SLIC (67.56\%) and SP+SLIC (65.97\%),
\ie., the merging and sieving are also effective with SLIC as they were with SEEDS.
% In addition, without relying on specific superpixel algorithms like SEEDS or SLIC,
We believe that our proposed method can work with any base superpixels
even when they are from 
an unsupervised segmentation method~\cite{ke2022unsupervised} or a foundation model~\cite{kirillov2023segment}.

\section{Conclusion}
In this work, we propose an adaptive active learning framework with adaptive superpixels.
Our merging and sieving methods operate adaptively every round, and the experimental results demonstrate the performance improvement of adaptive merging in various realistic situations.
Furthermore, we suggest novel achievable metrics for evaluating superpixels in advance that are suitable for active learning.

% AIGS
% AI 혁신허브 확인
% 2018R1A5A1060031
% \smallskip\noindent\textbf{Acknowledgements.}
% This work was partly supported by Institute of Information \& communications Technology Planning \&
% Evaluation (IITP) grant funded by the Korea government(MSIT) (No.2019-0-01906, Artificial Intelligence Graduate School Program(POSTECH)) and Institute
% of Information \& communications Technology Planning \& Evaluation (IITP) grant funded by the Korea
% government(MSIT) (No.2021-0-02068, Artificial Intelligence Innovation Hub) and the National Research
% Foundation of Korea(NRF) grant funded by the Korea
% government(MSIT) (No.2018R1A5A1060031).

\vspace{4mm}
%suha--a short version
{
% \small
\noindent \textbf{Acknowledgement.}% \section*{Acknowledgements}
~This work was supported by the IITP grants and
the NRF grant 
funded by Ministry of Science and ICT, Korea
(IITP-2019-0-01906, Artificial Intelligence Graduate School Program (POSTECH); IITP-2021-0-02068, Artificial Intelligence Innovation Hub; NRF-2021M3E5D2A01023887; NRF-2018R1A5A1060031).
}

\newpage
{\small
\bibliographystyle{ieee_fullname}
\bibliography{egbib}
}

\newpage
\appendix

\section*{\fontsize{14pt}{\baselineskip}\selectfont Appendix}
\vspace{3mm}

% \section{Notations}
% The notations used in the paper are defined in Table \ref{tab:notations}, where the subscript $t$ corresponds to round $t$.
% {
% \hypersetup{linkcolor=red}
% \tableofcontents
% \setcounter{tocdepth}{0}
% }

\section{
More 
gain with other base superpixel sizes
}
\label{sec:base-superpixel-sizes}

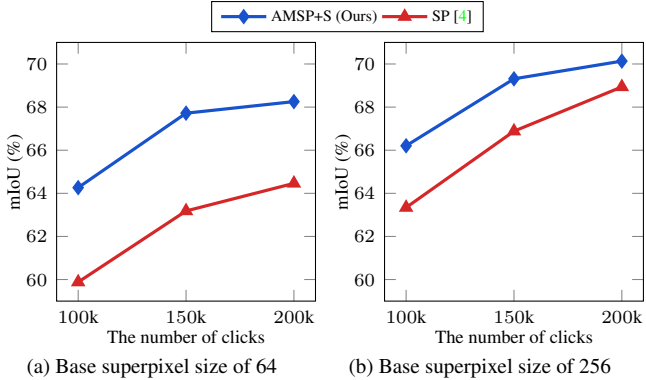
\begin{figure}[!ht]
    \captionsetup[subfigure]{font=footnotesize,labelfont=footnotesize,aboveskip=0.05cm,belowskip=-0.15cm}
    \centering
    \hspace{-5mm}
    \begin{subfigure}{.49\linewidth}
        \centering
        \begin{tikzpicture}
            \begin{axis}[
                legend style={nodes={scale=0.6}, at={(1.65, 1.16)}},
                legend columns=-1,
                xlabel={The number of clicks},
                ylabel={mIoU (\%)},
                width=1.23\linewidth,
                height=1.23\linewidth,
                ymin=59.,
                ymax=71,
                ytick={58, 60, 62, 64, 66, 68, 70, 72, 74},
                xlabel style={yshift=0.15cm},
                ylabel style={yshift=-0.2cm},
                xmin=90,
                xmax=210,
                label style={font=\scriptsize},
                tick label style={font=\scriptsize},
                xticklabel={$\pgfmathprintnumber{\tick}$k}
            ]
            % AM-SP
            \addplot[cdeepBP, very thick, mark=diamond*, mark size=2pt, mark options={solid}] table[col sep=comma, x=x, y=am-sp]{Data/limited_budget_cityscapes_64.csv};
            % SP
            \addplot[cdeepMF, very thick, mark=triangle*, mark size=2pt, mark options={solid}] table[col sep=comma, x=x, y=revisiting]{Data/limited_budget_cityscapes_64.csv};
            
            % SP
            % \addplot[name path=revisiting-l, draw=none, fill=none] table[col sep=comma, x=x, y=revisiting-l]{Data/limited_budget_cityscapes.csv};
            % \addplot[name path=revisiting-u, draw=none, fill=none] table[col sep=comma, x=x, y=revisiting-u]{Data/limited_budget_cityscapes.csv};
            % \addplot[cdeepMF, fill opacity=0.15] fill between[of=revisiting-l and revisiting-u];
            % AM-SP
            % \addplot[name path=am-sp-l, draw=none, fill=none] table[col sep=comma, x=x, y=am-sp-l]{Data/limited_budget_cityscapes.csv};
            % \addplot[name path=am-sp-u, draw=none, fill=none] table[col sep=comma, x=x, y=am-sp-u]{Data/limited_budget_cityscapes.csv};
            % \addplot[cdeepBP, fill opacity=0.15] fill between[of=am-sp-l and am-sp-u]; 
            
            \legend{AMSP+S (Ours), SP~\cite{cai2021revisiting}}
            \end{axis}
        % \node[above] at (3.7, 3.41) {\small Performance for varying budget};
        \end{tikzpicture}
        \caption{Base superpixel size of 64}
        \label{fig:(a)-small-region}
    \end{subfigure}
    \hspace{1mm}
    \begin{subfigure}{.49\linewidth}
        \centering
        \begin{tikzpicture}
            \begin{axis}[
                legend style={nodes={scale=0.6}, at={(1.65, 1.16)}},
                legend columns=-1,
                xlabel={The number of clicks},
                ylabel={mIoU (\%)},
                width=1.23\linewidth,
                height=1.23\linewidth,
                ymin=59,
                ymax=71,
                ytick={60, 62, 64, 66, 68, 70, 72, 74},
                xlabel style={yshift=0.15cm},
                ylabel style={yshift=-0.2cm},
                xmin=90,
                xmax=210,
                label style={font=\scriptsize},
                tick label style={font=\scriptsize},
                xticklabel={$\pgfmathprintnumber{\tick}$k}
            ]
            % Oracle
            % \addplot[cCL, very thick, mark=pentagon*, mark size=2pt, mark options={solid}] table[col sep=comma, x=x, y=oracle-avg]{Data/limited_budget_cityscapes.csv};
            % AM-SP
            \addplot[cdeepBP, very thick, mark=diamond*, mark size=2pt, mark options={solid}] table[col sep=comma, x=x, y=am-sp]{Data/limited_budget_cityscapes_256.csv};
            % M-SP
            % \addplot[cBP, very thick, mark=square*, mark size=2pt, mark options={solid}] table[col sep=comma, x=x, y=m-sp-avg]{Data/limited_budget_cityscapes.csv};
            % SP
            \addplot[cdeepMF, very thick, mark=triangle*, mark size=2pt, mark options={solid}] table[col sep=comma, x=x, y=revisiting]{Data/limited_budget_cityscapes_256.csv};
            \end{axis}
        % \node[above] at (3.7, 3.41) {\small Performance for varying budget};
        \end{tikzpicture}
        \caption{Base superpixel size of 256}
        \label{fig:(b)-small-region}
    \end{subfigure}
    \caption{{\em Effect of base superpixel size on Cityscapes.} The performance difference is greater when the superpixel size is smaller.
    % Each experiment is conducted with three trials and the shaded region indicates range
    }
    \label{fig:supple-region-size-cityscapes}
\end{figure}
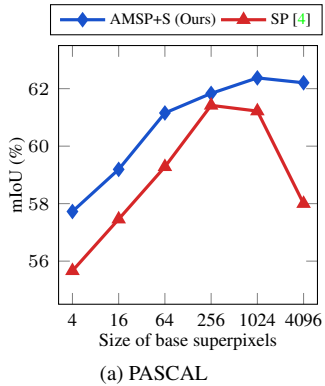
\begin{figure}[!ht]
    \captionsetup[subfigure]{font=footnotesize,labelfont=footnotesize,aboveskip=0.05cm,belowskip=-0.15cm}
    \centering
    \hspace{-5mm}
    \begin{subfigure}{.49\linewidth}
        \centering
        \begin{tikzpicture}
            \begin{axis}[
                legend style={nodes={scale=0.6}, at={(1.02, 1.16)}},
                xlabel={Size of base superpixels},
                ylabel={mIoU (\%)},
                width=1.23\linewidth,
                height=1.23\linewidth,
                ymin=54.5,
                ymax=63.5,
                ytick={54, 56, 58, 60, 62, 64, 66},
                xlabel style={yshift=0.15cm},
                ylabel style={yshift=-0.2cm},
                legend columns=2,
                xmin=0.7,
                xmax=6.3,
                label style={font=\scriptsize},
                tick label style={font=\scriptsize},
                xtick=data,
                xticklabels={4,16,64,256,1024,4096},
            ]
            \addplot[cdeepBP, very thick, mark=diamond*, mark size=2pt, mark options={solid}] table[col sep=comma, x=x, y=sm]{Data/supple_region_size_pascal.csv};
            \addplot[cdeepMF, very thick, mark=triangle*, mark size=2pt, mark options={solid}] table[col sep=comma, x=x, y=revisiting] {Data/supple_region_size_pascal.csv};
            \legend{AMSP+S (Ours), SP~\cite{cai2021revisiting}}
            \end{axis}
        \end{tikzpicture}
        \caption{PASCAL}
    \end{subfigure}
    \caption{{\em Effect of base superpixel size on PASCAL.} Our method exhibits robustness to large superpixels, while the baseline is sensitive. 
    % Each experiment is conducted with three trials and the shaded region indicates range
    }
    \label{fig:supple-region-size-pascal}
\end{figure}

For ease of exposition, Figure~\ref{fig:robustness} presents the gain of our method (compared to \textit{SP} \cite{cai2021revisiting}) for a limited set of base superpixel sizes. In this section, we report an additional investigation
suggesting further gain with different base superpixels.

% \label{fig:sup-small-cityscapes}
\smallskip\noindent\textbf{Further gain on Cityscapes.}
In Figure~~\ref{fig:(a)-small-region},
we additionally provide a comparison between
the proposed method (\textit{AMSP+S})
and \textit{SP} \cite{cai2021revisiting}, 
where the experiment setup with
Cityscapes
is identical to 
that in Figure~\ref{fig:(a)-effect} except that the base superpixel size is 64 (Figure~~\ref{fig:(a)-small-region}) instead of 256 (Figure~~\ref{fig:(b)-small-region}).
Our adaptive merging method (\textit{AMSP+S}) is especially effective when the superpixel size is small in Figure~\ref{fig:(a)-small-region}, thanks to the adaptive merging mechanism.
This observation suggests more significant gain of our method with other choices of base superpixel size than that in Figure~\ref{fig:robustness}.

% The experimental setup used in Figure \ref{fig:(b)-small-region} is identical to that of Figure~\ref{fig:(a)-effect}. 

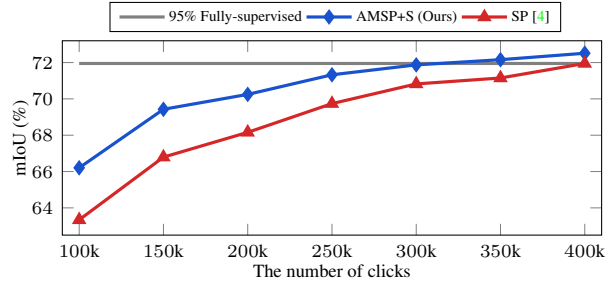
\begin{figure}[!ht]
    \centering
    \begin{tikzpicture}
        \begin{axis}[
            legend style={nodes={scale=0.57}, at={(0.925, 1.2)}},
            legend columns=-1,
            xlabel={The number of clicks},
            ylabel={mIoU (\%)},
            width=1.05\linewidth,
            height=0.5\linewidth,
            ymin=62.5,
            ymax=73.2,
            ytick={64, 66, 68, 70, 72, 74},
            xlabel style={yshift=0.15cm},
            ylabel style={yshift=-0.2cm},
            xmin=90,
            xmax=410,
            label style={font=\scriptsize},
            tick label style={font=\scriptsize},
            xticklabel={$\pgfmathprintnumber{\tick}$k}
        ]
        % 95%
        \addplot[gray, very thick] table[col sep=comma, x=x, y=sup]{Data/Rebuttal/limited_budget_cityscapes_rebuttal.csv};
        % 95%
        \addplot[cdeepBP, very thick, mark=diamond*, mark size=2pt, mark options={solid}] table[col sep=comma, x=x, y=am-sp]{Data/Rebuttal/limited_budget_cityscapes_rebuttal.csv};
        % SP
        \addplot[cdeepMF, very thick, mark=triangle*, mark size=2pt, mark options={solid}] table[col sep=comma, x=x, y=revisiting]{Data/Rebuttal/limited_budget_cityscapes_rebuttal.csv};
        \legend{95\% Fully-supervised, AMSP+S (Ours), SP \cite{cai2021revisiting}}
        \end{axis}
    \end{tikzpicture}
    % \vspace{-2mm}
    \caption{{\em Additional rounds experiments on Cityscapes.} We extend the experiments in Figure~\ref{fig:(a)-effect} up to a budget of 400k. The performance improvement remains consistent across various additional budgets.}
    \label{fig:95}
\end{figure}

\begin{table}[!ht]
\centering
\setlength\tabcolsep{6pt}
\begin{tabular}{l|c}
\toprule
Methods & mIoU \\ \midrule
\textit{SP} \cite{cai2021revisiting} & 63.77 \\ \midrule
\textit{AMSP+S} (bottom 10\%) & 64.33 \\ \midrule
\textit{AMSP+S} (top 10\%) & \underline{65.99} \\ \midrule
\rowcolor{Gray}
\textit{AMSP+S} (complete 100\%) & \textbf{66.53} \\ \midrule
\bottomrule
\end{tabular}
\caption{{\em Various levels of partial merging.} Experiments are conducted under the same setting of Figure~\ref{fig:(a)-effect} with 100k clicks (Cityscapes, superpixel size of 256).}
\label{tab:sup-descending}
\end{table}

\iffalse
\begin{table}[!ht]
\centering
\setlength\tabcolsep{6pt}
\begin{tabular}{l|c}
\toprule
Methods & mIoU \\ \midrule
\textit{SP} \cite{cai2021revisiting} & 63.77 \\ \midrule
\textit{AMSP+S} (ascending, 10\%) & 64.33 \\ \midrule
\textit{AMSP+S} (descending, 10\%) & \underline{65.99} \\ \midrule
\rowcolor{Gray}
\textit{AMSP+S} (descending, 100\%) & \textbf{66.53} \\ \midrule
\bottomrule
\end{tabular}
\caption{{\em Various merging order.} Experiments are conducted on Cityscapes dataset with an average superpixel size of 256, using 100k costs for two rounds.}
\label{tab:descending}
\end{table}
\fi

\smallskip\noindent\textbf{Further gain on PASCAL.}
We also demonstrate a larger gap between 
the proposed method and existing one 
in PASCAL. 
% explore the effectiveness of the proposed framework by altering the base superpixel size on PASCAL.
In Figure~\ref{fig:supple-region-size-pascal}, our adaptive merging method (\textit{AMSP+S}) outperforms the baseline (\textit{SP}) for various superpixel sizes
as we observed in Figure~\ref{fig:robustness}.
We stress that the gain of the proposed method is particularly larger than the one reported in Figure~\ref{fig:robustness} when
the base superpixel size is 4096, which is much larger than 256 used in Figure~\ref{fig:robustness}. This is because 
the sieving procedure to 
suppresses
the noise from dominant labeling 
becomes more crucial when querying large superpixels.
The experimental setup used in Figure \ref{fig:supple-region-size-pascal} is identical to that of Figure \ref{fig:(d)-effect}. 

%\khy{
\smallskip\noindent\textbf{Further rounds on Cityscapes.}
To demonstrate the efficacy of our method across various budgets, we experiment by gradually increasing the budget as illustrated in Figure~\ref{fig:95} on Cityscapes.
The experimental setting in Figure~\ref{fig:95} remains consistent with that of Figure~\ref{fig:(a)-effect}.
The advantage of our method over SP~\cite{cai2021revisiting} is continued in further rounds.
We remark that the proposed method nearly achieves the 95\% mIoU of the fully
supervised model (71.95\%) at 300k clicks, whereas SP does at 400k clicks.

\section{Rationale for line~\ref{line:order} of Algorithm \ref{algorithm2}}
\label{sec:rationale-merging}

\begin{figure*}[!t]
    \captionsetup[subfigure]{font=footnotesize}
    \centering
    \begin{subfigure}{.33\linewidth}
        \centering
        \includegraphics[scale=0.322]{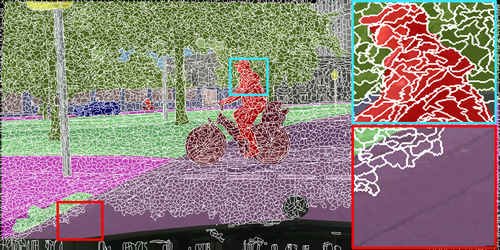}
    \end{subfigure}
    \begin{subfigure}{.33\linewidth}
        \centering
        \includegraphics[scale=0.322]{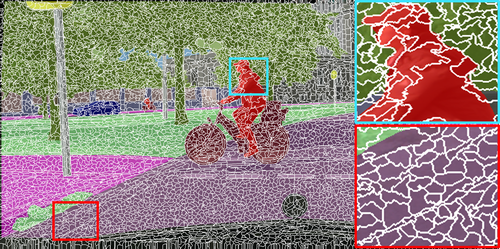}
    \end{subfigure}
    \begin{subfigure}{.33\linewidth}
        \centering
        \includegraphics[scale=0.322]{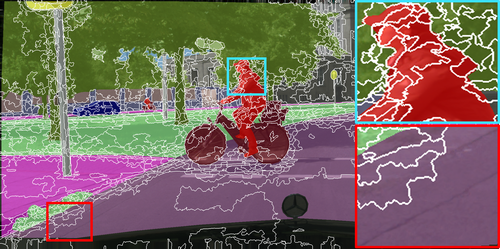}
        % \caption{$ASA(S;G)=1.00, \; AF(G;S)=1.00$}
    \end{subfigure}

    \begin{subfigure}{.33\linewidth}
        \centering
        \includegraphics[scale=0.322]{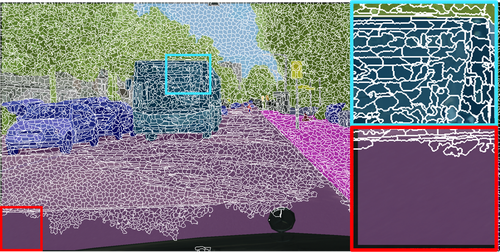}
        \caption{Merging superpixels with low 10\% uncertainty}
        \label{fig:(a)-partial}
    \end{subfigure}
    \begin{subfigure}{.33\linewidth}
        \centering
        \includegraphics[scale=0.322]{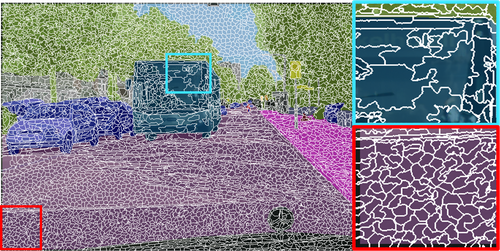}
        \caption{Merging superpixels with high 10\% uncertainty}
        \label{fig:(b)-partial}
    \end{subfigure}
    \begin{subfigure}{.33\linewidth}
        \centering
        \includegraphics[scale=0.322]{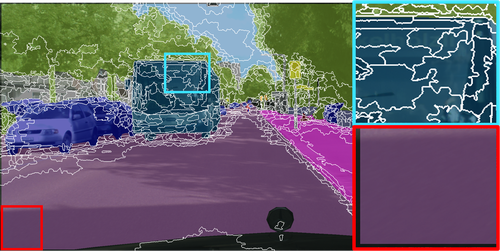}
        % \caption{$ASA(S;G)=1.00, \; AF(G;S)=1.00$}
        \caption{Merging all superpixels}
        \label{fig:(c)-partial}
    \end{subfigure}
    \caption{{\em Qualitative results for partial merging.} 
    % The average uncertainty of pixels in the cyan box is 0.30 for the top images and 0.27 for the bottom images, while for pixels in the red box, it is 0.09 for the top images and 0.00 for the bottom images. 
    % High acquisition values are observed within the cyan boxes, while the red boxes hold low acquisition values.
    The cyan boxes encompass superpixels exhibiting the highest 10\% uncertainty, while the red boxes encompass superpixels with the lowest 10\% uncertainty.
    (b) By merging only a portion of superpixels in the order of high uncertainty, we can reduce time complexity, as it creates similar merged superpixels compared with the cyan box in (c).
    % (\ie in the cyan boxes of (b) and (c)) with high acquisition values. 
    % (a) Only 10 \% of superpixels are merged in the order of low uncertainty. 
    % (b) Only 10 \% of superpixels are merged in the order of high uncertainty. 
    % (c) For merging, our adaptive merging method explores all superpixels in an image.
    }
    \label{fig:descend}
\end{figure*}

\begin{table}[!ht]
\centering
\setlength\tabcolsep{6pt}
\begin{tabular}{l|c}
\toprule
Methods & mIoU \\ \midrule
\textit{SP} \cite{cai2021revisiting} & 63.77 \\ \midrule
\textit{AMSP+S} $(\phi(s;\theta) = 0.0)$ & \underline{65.35} \\ \midrule
\textit{AMSP+S} $(\phi(s;\theta) = 0.2)$ & 61.80 \\ \midrule
\textit{AMSP+S} $(\phi(s;\theta) = 0.4)$ & 57.77 \\ \midrule
\textit{AMSP+S} $(\phi(s;\theta) = 0.6)$ & 45.84 \\ \midrule
\textit{AMSP+S} $(\phi(s;\theta) = 0.8)$ & 38.99 \\ \midrule
% Class-wise & 00.00 \\ \midrule
\rowcolor{Gray}
\textit{AMSP+S} (Kneedle \cite{satopaa2011finding}) & \textbf{66.53} \\ \midrule
\bottomrule
\end{tabular}
\caption{{\em Various sieving methods.} Experiments are conducted on Cityscapes dataset with an average superpixel size of 256, using 100k costs for two rounds.}
\label{tab:sieving}
\end{table}

% We utilize a graph to merge superpixels by converting them into nodes and edges.
% However,
We explain the rationale behind traversing nodes in the descending order of uncertainty in line~\ref{line:order} of Algorithm \ref{algorithm2}.
Our merging process requires a linear time complexity proportional to the size of the base superpixels graph.
However, due to the advantage of merging in descending uncertainty order, we are able to acquire merged superpixels with considerable uncertainty at the beginning of merging.
To reduce merging time complexity, we only merge the top $10\%$ of base superpixels with the highest uncertainty as query candidates.
Table~\ref{tab:sup-descending} shows that it is important to prioritize the merging highly uncertain superpixels,
and
merging along the ascending order of uncertainty degenerates the performance.

In Figure~\ref{fig:descend},
we exemplify the merged superpixels
from the partial merging in the ascending or descending order of uncertainty,
and the full merging, where
the cyan boxes contain
higher values of acquisition function
than the red boxes.
The partial merging with the ascending order of uncertainty regrettably merges
the superpixels that would not be selected in AL, while that with the ascending order
efficiently combines the base superpixels
of which selection is highly like.
This difference indeed results in a huge gap in the final performance as shown in Table~\ref{tab:sup-descending}.

\section{Rationale for the adaptive threshold $\phi(s;\theta)$ in the sieving}
\label{fig:sup-sieving}
\begin{figure}[!t]
    \captionsetup[subfigure]{font=footnotesize,labelfont=footnotesize,aboveskip=0.05cm,belowskip=-0.15cm}
    \centering
    \hspace{-3mm}
    \begin{subfigure}{.47\linewidth}
        \centering
        \begin{tikzpicture}
            \begin{axis}[
                legend style={nodes={scale=0.6}, at={(1.55, 1.16)}},
                legend columns=-1,
                xlabel={$x$},
                ylabel={$f_\theta(\text{road};x)$},
                width=1.23\linewidth,
                height=1.23\linewidth,
                ymin=-0.1,
                ymax=1.1,
                xlabel style={yshift=0.15cm},
                ylabel style={yshift=-0.2cm},
                xmin=-0.5,
                xmax=18.5,
                label style={font=\scriptsize},
                tick label style={font=\scriptsize},
                xtick=data
            ]
            % Road
            \addplot[cdeepBP, very thick, mark=diamond*, mark size=2pt, mark options={solid}] table[col sep=comma, x=x, y=y]{Data/knee_road.csv};
            \draw[orange, very thick, dashed] (2,-1) -- (2,2);
            \addlegendimage{dashed, line width=0.4mm, height=1mm, color=orange}
            \legend{data, knee/elbow}
            % \legend{AMSP+S (Ours), SP~\cite{cai2021revisiting}}
            \end{axis}
        % \node[above] at (3.7, 3.41) {\small Performance for varying budget};
        \end{tikzpicture}
        \caption{Road}
    \end{subfigure}
    \hspace{1mm}    
    \begin{subfigure}{.47\linewidth}
        \centering
        \begin{tikzpicture}
            \begin{axis}[
                legend style={nodes={scale=0.35}, at={(0.03, 0.24)}, anchor=west}, 
                xlabel={$x$},
                ylabel={$f_\theta(\text{pole};x)$},
                width=1.23\linewidth,
                height=1.23\linewidth,
                ymin=-0.1,
                ymax=1.1,
                xlabel style={yshift=0.15cm},
                ylabel style={yshift=-0.2cm},
                xmin=-0.5,
                xmax=18.5,
                label style={font=\scriptsize},
                tick label style={font=\scriptsize},
                xtick=data,
            ]

            % Pole
            \addplot[cdeepBP, very thick, mark=diamond*, mark size=2pt, mark options={solid}] table[col sep=comma, x=x, y=y]{Data/knee_pole.csv};
            \draw[orange, very thick, dashed] (4,-1) -- (4,2);
            \addlegendimage{dashed, line width=0.4mm, height=1mm, color=cdeepBP32}
            % \legend{data, knee/elbow}
            \end{axis}
        \end{tikzpicture}
        \caption{Pole}
    \end{subfigure}
    \caption{{\em Examples of knee points on Cityscapes.} We obtain (a) a high knee value for the common road class and (b) a low knee value for the rare pole class.}
    \label{fig:sup-knee-points}
\end{figure}
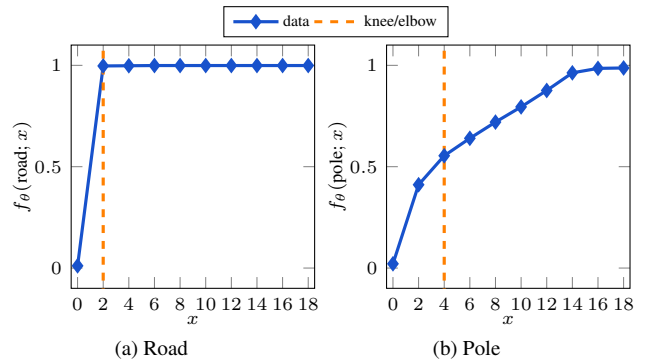

\begin{figure*}[t!]
\captionsetup[subfigure]{font=footnotesize,labelfont=footnotesize,aboveskip=0.05cm,belowskip=-0.15cm}
\centering
\begin{tikzpicture}
    \begin{axis}[
        width  = \textwidth,
        axis y line*=left,
        symbolic x coords={
            \rotatebox{60}{Road},
            \rotatebox{60}{Building},
            \rotatebox{60}{Vegetation},
            \rotatebox{60}{Car},
            \rotatebox{60}{Sidewalk},
            \rotatebox{60}{Sky},
            \rotatebox{60}{Pole},
            \rotatebox{60}{Person},
            \rotatebox{60}{Terrain},
            \rotatebox{60}{Fence},
            \rotatebox{60}{Wall},
            \rotatebox{60}{Sign},
            \rotatebox{60}{Bicycle},
            \rotatebox{60}{Truck},
            \rotatebox{60}{Bus},
            \rotatebox{60}{Train},
            \rotatebox{60}{Light},
            \rotatebox{60}{Rider},
            \rotatebox{60}{Motorcycle},
        },
        axis x line=bottom,
        height = 5.2cm,
        major x tick style = transparent,
        %axis on top,
        ybar=3*\pgflinewidth,
        bar width=4pt,
        ymajorgrids = true,
        ylabel = {IoU},
        xtick = data,
        scaled y ticks = false,
        enlarge x limits=0.3,
        axis line style={-},
        ymin=0.0,ymax=1,
        legend columns=2,
        legend cell align=left,
        legend style={
                nodes={scale=0.6},
                at={(0.5,1.2)},
                anchor=north,
                column sep=1ex
        },
        label style={font=\scriptsize},
        tick label style={font=\scriptsize}
    ]
        \addplot[style={cdeepBP,fill=cdeepBP,mark=none}] coordinates {
            (\rotatebox{60}{Road}, 0.964601)
            (\rotatebox{60}{Building}, 0.883573)
            (\rotatebox{60}{Vegetation}, 0.897434)
            (\rotatebox{60}{Car}, 0.908026)
            (\rotatebox{60}{Sidewalk}, 0.750152)
            (\rotatebox{60}{Sky}, 0.906192)
            (\rotatebox{60}{Pole}, 0.482772)
            (\rotatebox{60}{Person}, 0.729143)
            (\rotatebox{60}{Terrain}, 0.55855)
            (\rotatebox{60}{Fence}, 0.492545)
            (\rotatebox{60}{Wall}, 0.461659)
            (\rotatebox{60}{Sign}, 0.614888)
            (\rotatebox{60}{Bicycle}, 0.67332)
            (\rotatebox{60}{Truck}, 0.559177)
            (\rotatebox{60}{Bus}, 0.733037)
            (\rotatebox{60}{Train}, 0.601527)
            (\rotatebox{60}{Light}, 0.468385)
            (\rotatebox{60}{Rider}, 0.483054)
            (\rotatebox{60}{Motorcycle}, 0.473352)
        };
        \addplot[style={orange,fill=orange,mark=none}] coordinates {
            (\rotatebox{60}{Road}, 0.955517)
            (\rotatebox{60}{Building}, 0.852214)
            (\rotatebox{60}{Vegetation}, 0.880035)
            (\rotatebox{60}{Car}, 0.851916)
            (\rotatebox{60}{Sidewalk}, 0.89579)
            (\rotatebox{60}{Sky}, 0.896663)
            (\rotatebox{60}{Pole}, 0.324208)
            (\rotatebox{60}{Person}, 0.523996)
            (\rotatebox{60}{Terrain}, 0.479043)
            (\rotatebox{60}{Fence}, 0.376733)
            (\rotatebox{60}{Wall}, 0.211172)
            (\rotatebox{60}{Sign}, 0.431976)
            (\rotatebox{60}{Bicycle}, 0.503692)
            (\rotatebox{60}{Truck}, 0.097781)
            (\rotatebox{60}{Bus}, 0.401534)
            (\rotatebox{60}{Train}, 0.128532)
            (\rotatebox{60}{Light}, 0.058156)
            (\rotatebox{60}{Rider}, 0.04937)
            (\rotatebox{60}{Motorcycle}, 0.01)
        };
        % \addplot[style={cdeepMF,fill=cdeepMF,mark=none}]
             % coordinates {(Road, 19.22) (Building, 21.29) (Vegetation, 21.58)};
        \legend{AMSP+S (Kneedle \cite{satopaa2011finding}), AMSP+S $( \phi(s) = 0.6 )$}
    \end{axis}
\end{tikzpicture}
\caption{{\em Class-wise IoU according to $\phi(s;\theta)$.} Applying the same $\phi(s)$ of 0.6 to all pixels results in excessive sieving for relatively rare classes, leading to decreased performance for these classes (\eg Light, Rider, and Motorcycle). Based on the ground-truth, class labels are organized in order of the total pixel count for each class.}
\label{fig:sup-class-dist}
\end{figure*}
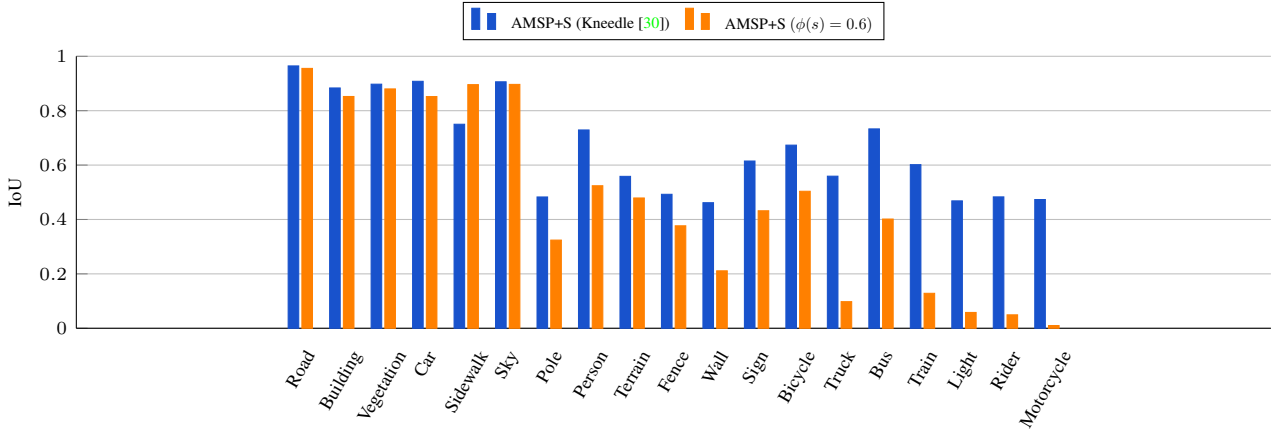

\begin{figure*}[t!]
    \captionsetup[subfigure]{font=footnotesize}
    \centering
    \begin{subfigure}{.33\linewidth}
        \centering
        \includegraphics[scale=0.322]{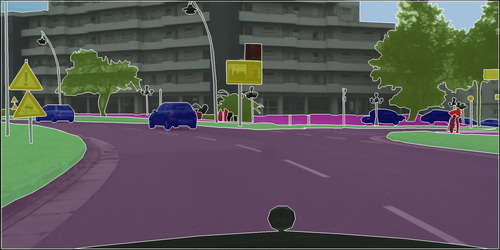}
        \caption{Semantic segmentation}
        \label{subfig:sementic-seg}
    \end{subfigure}
    \begin{subfigure}{.33\linewidth}
        \centering
        \includegraphics[scale=0.322]{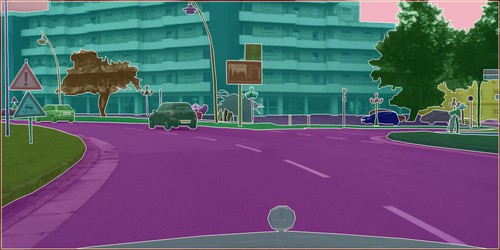}
        \caption{Panoptic segmentation}
    \end{subfigure}
    \begin{subfigure}{.33\linewidth}
        \centering
        \includegraphics[scale=0.322]{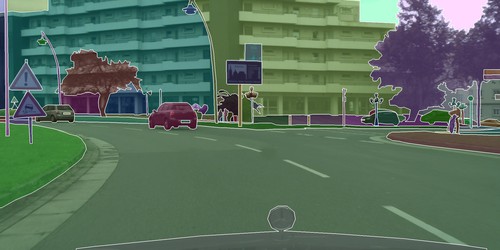}
        \caption{Oracle superpixels (ours)}
        \label{subfig:oralce-seg}
    \end{subfigure} 
    \caption{{\em Difference between conventional segmentations and oracle superpixels.} (a) When sharing the same class label, they are depicted as identical superpixels (\ie green color on separate trees).  (b) Although a building is divided by a pole, it is represented as a single superpixel (\ie cyan color). (c) We consider a building as two distinct superpixels (\ie cyan and light yellow colors).}
    \label{fig:sup-oracle-superpixels}
\end{figure*}

We provide the reason for introducing 
the threshold function $\phi(s)$
personalized for each superpixel $s$, described in Section \ref{sec:sieving-technique}.
We obtain the dominant label $\text{D}(s)$ for a queried superpixel $s$, however, we only propagate the label to pixels $x \in s$ that are predicted to have a positive impact on the training of model $\theta$ as:
\begin{equation}
h(s;\theta) := \{ x \in s: f_\theta \big( \text{D}(s); x \big) \geq \phi(s; \theta) \} \;,
\end{equation}
where $f_\theta \left( \text{D}(s);x \right)$ implies the confidence of pixel $x$ to dominant label $\text{D}(s)$ given $\theta$ and $\phi(s;\theta)$ determines the degree of sieving.
In Table \ref{tab:sieving}, we study the effect of various $\phi(s;\theta)$.
When the same $\phi(s;\theta)$ is applied to all pixels, it causes class imbalance by leaving relatively easy classes as described in Figure \ref{fig:sup-class-dist}.
To avoid this issue, we utilize the Kneedle algorithm \cite{satopaa2011finding} to obtain different $\phi(s;\theta)$ for each superpixel $s$.
Specifically, $\phi(s; \theta)$ is a knee point of the cumulative distribution function of values of $f_\theta \big( \text{D}(s); x \big)$ in superpixel $x \in s$.
However, for the Kneedle algorithm to work accurately, the curve of cumulative distribution must be either convex or concave. 
In addition, the algorithm may provide inaccurate knee points on very smooth curves.
To address this issue, we use a subset of uniformly sampled values based on $f_\theta(\text{D}(s);x)$, instead of using the distribution for all pixels.
We sample 20 and 5 pixels for Cityscapes and PASCAL datasets, respectively.
In Figure \ref{fig:sup-knee-points}, different knee points are detected according to the dominant class of superpixels.

% \khy{
\smallskip\noindent\textbf{Effect of sieving.} 
Our sieving method exhibits a significant effect on larger superpixels, as illustrated in Figure \ref{fig:(c)-effect} and Figure \ref{fig:supple-region-size-pascal}.
Especially, in Figure \ref{fig:supple-region-size-pascal} with a large base superpixel size of 4096, the first sieving excises 45.87\% of the mislabeled pixels that disagree with their dominant labels.
Furthermore, we observe that the sieving
is progressively refined round by round.
For instance, in Figure \ref{fig:(a)-effect}, the portion of the mislabeled labels removed by the sieving increases over four rounds as
follows: 3.58\%, 8.54\%, 10.46\%, and 12.43\%.
Our sieving technique enhances label quality by retaining only high-confidence labels and continuously improves through multiple rounds.
% }

\begin{figure*}[t!]
    \captionsetup[subfigure]{font=footnotesize,labelfont=footnotesize,aboveskip=0.05cm,belowskip=-0.15cm}
    \centering
    \hspace{-1mm}
    \begin{subfigure}{.235\linewidth}
        \centering
        \begin{tikzpicture}
            \begin{axis}[
                legend style={nodes={scale=0.35}, at={(0.03, 0.24)}, anchor=west}, 
                xlabel={AF$(G;S)$},
                ylabel={mIoU (\%)},
                width=1.23\linewidth,
                height=1.23\linewidth,
                ymin=50.8,
                ymax=63.2,
                xlabel style={yshift=0.15cm},
                ylabel style={yshift=-0.2cm},
                ytick={51, 53, 55, 57, 59, 61, 63},
                legend columns=2,
                xmin=0.06,
                xmax=0.42,
                label style={font=\scriptsize},
                tick label style={font=\scriptsize},
                x tick label style={
                    /pgf/number format/.cd,
                        fixed,
                }
            ]
            \addplot[cdeepBP, only marks] table[col sep=comma, x=AFGS, y=mIoU]{Data/correlation_afgs_semantic.csv};
            \addplot[very thick, orange] table[col sep=comma, x=AFGS, y={create col/linear regression = {y=mIoU}}
            ]{Data/correlation_afgs_semantic.csv};
            \draw (0.5\linewidth, 0.35\linewidth) node {\scriptsize $\text{Corr}=0.57$};
            \end{axis}
        \end{tikzpicture}
        \caption{Semantic segmentation}
    \end{subfigure}
    \hspace{1mm}
    \begin{subfigure}{.235\linewidth}
        \centering
        \begin{tikzpicture}
            \begin{axis}[
                legend style={nodes={scale=0.35}, at={(0.03, 0.24)}, anchor=west}, 
                xlabel={AF$(G;S)$},
                ylabel={mIoU (\%)},
                width=1.23\linewidth,
                height=1.23\linewidth,
                ymin=50.8,
                ymax=63.2,
                xlabel style={yshift=0.15cm},
                ylabel style={yshift=-0.2cm},
                ytick={51, 53, 55, 57, 59, 61, 63},
                legend columns=2,
                xmin=0.21,
                xmax=0.39,
                label style={font=\scriptsize},
                tick label style={font=\scriptsize},
                x tick label style={
                    /pgf/number format/.cd,
                        fixed,
                }
            ]
            \addplot[cdeepBP, only marks] table[col sep=comma, x=AFGS, y=mIoU]{Data/correlation_afgs_panoptic.csv};
            \addplot[very thick, orange] table[col sep=comma, x=AFGS, y={create col/linear regression = {y=mIoU}}
            ]{Data/correlation_afgs_panoptic.csv};
            \draw (0.5\linewidth, 0.35\linewidth) node {\scriptsize$\textbf{Corr}=\textbf{0.95}$};
            \end{axis}
        \end{tikzpicture}
        \caption{Panoptic segmentation}
    \end{subfigure}
    \hspace{1mm}
    \begin{subfigure}{.235\linewidth}
        \centering
        \begin{tikzpicture}
            \begin{axis}[
                legend style={nodes={scale=0.35}, at={(0.03, 0.24)}, anchor=west}, 
                xlabel={AF$(G;S)$},
                ylabel={mIoU (\%)},
                width=1.23\linewidth,
                height=1.23\linewidth,
                ymin=50.8,
                ymax=63.2,
                xlabel style={yshift=0.15cm},
                ylabel style={yshift=-0.2cm},
                ytick={51, 53, 55, 57, 59, 61, 63},
                legend columns=2,
                xmin=0.173,
                xmax=0.371,
                label style={font=\scriptsize},
                tick label style={font=\scriptsize},
                x tick label style={
                    /pgf/number format/.cd,
                        fixed,
                }
            ]
            \addplot[cdeepBP, only marks] table[col sep=comma, x=AFGS, y=mIoU]{Data/correlation_afgs.csv};
            \addplot[very thick, orange] table[col sep=comma, x=AFGS, y={create col/linear regression = {y=mIoU}}
            ]{Data/correlation_afgs.csv};
            \draw (0.5\linewidth, 0.35\linewidth) node {\scriptsize$\textbf{Corr}=\textbf{0.95}$};
            \end{axis}
        \end{tikzpicture}
        \caption{Oracle superpixels}
    \end{subfigure}
    \caption{{\em Relationship between AF$(G;S)$ and mIoU varying $G$.} AF$(G;S)$ and mIoU exhibit a high correlation when ground-truth $G$ is represented by the panoptic segmentation and oracle superpixels in Figure \ref{fig:sup-oracle-superpixels}. For the correlation calculation, \textit{Oracle} in Table \ref{tab:quantitative} is excluded.}
    \label{fig:sup-afgs-g}
\end{figure*}
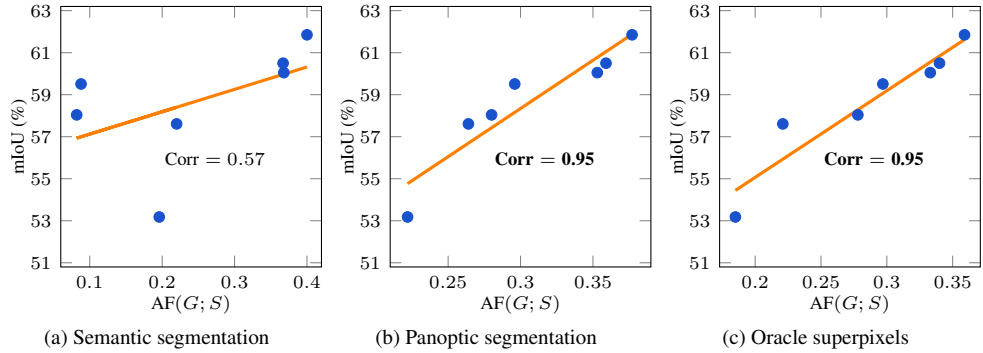

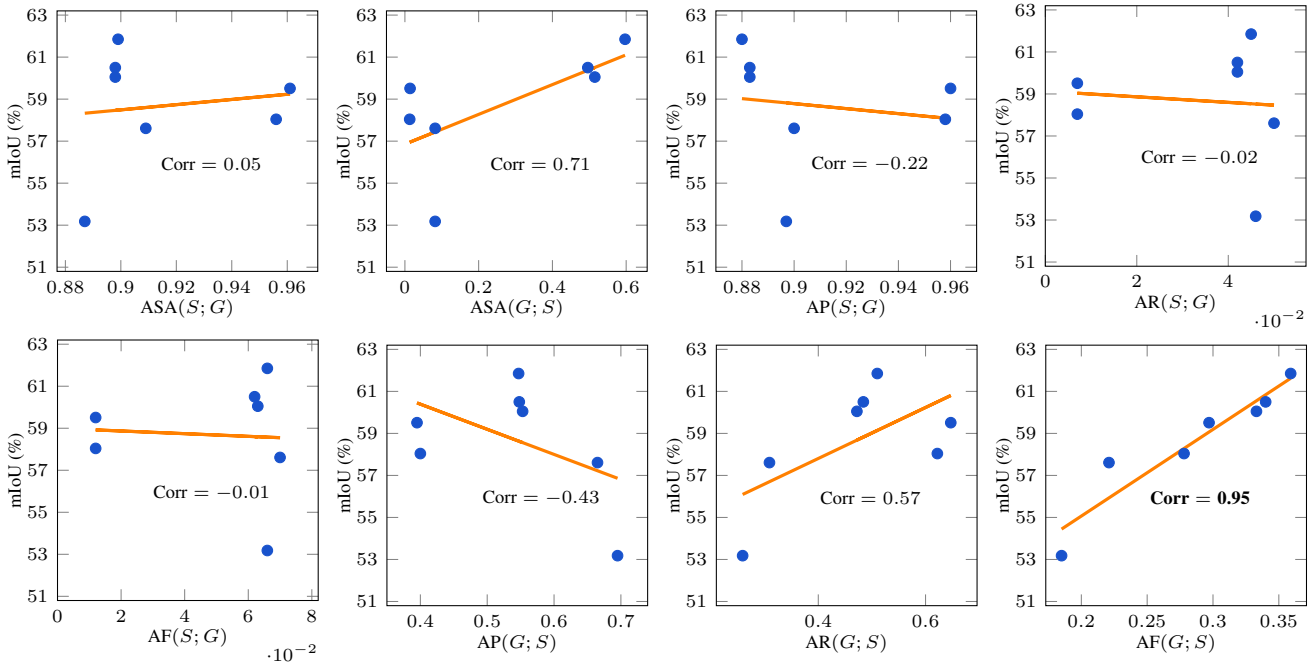
\begin{figure*}[!ht]
    \captionsetup[subfigure]{font=scriptsize,labelfont=scriptsize,aboveskip=0.05cm,belowskip=-0.15cm}
    \centering
    \hspace{-1mm}
    \begin{subfigure}{.235\linewidth}
        \centering
        \begin{tikzpicture}
            \begin{axis}[
                legend style={nodes={scale=0.35}, at={(0.03, 0.24)}, anchor=west}, 
                xlabel={ASA$(S;G)$},
                ylabel={mIoU (\%)},
                width=1.23\linewidth,
                height=1.23\linewidth,
                ymin=50.8,
                ymax=63.2,
                ytick={51, 53, 55, 57, 59, 61, 63},
                xlabel style={yshift=0.15cm},
                ylabel style={yshift=-0.2cm},
                legend columns=2,
                xmin=0.877,
                xmax=0.971,
                label style={font=\scriptsize},
                tick label style={font=\scriptsize},
                x tick label style={
                    /pgf/number format/.cd,
                        fixed,
                }
            ]
            \addplot[cdeepBP, only marks] table[col sep=comma, x=ASASG, y=mIoU]{Data/correlation_asasg.csv};
            \addplot[very thick, orange] table[col sep=comma, x=ASASG, y={create col/linear regression = {y=mIoU}}
            ]{Data/correlation_asasg.csv};
            \draw (0.5\linewidth, 0.35\linewidth) node {\scriptsize$\text{Corr} = 0.05$};
            \end{axis}
        \end{tikzpicture}
        % \caption{ASA(S;G) vs mIoU}
    \end{subfigure}
    \hspace{1mm}
    \begin{subfigure}{.235\linewidth}
        \centering
        \begin{tikzpicture}
            \begin{axis}[
                legend style={nodes={scale=0.35}, at={(0.03, 0.24)}, anchor=west}, 
                xlabel={ASA$(G;S)$},
                ylabel={mIoU (\%)},
                width=1.23\linewidth,
                height=1.23\linewidth,
                ymin=50.8,
                ymax=63.2,
                xlabel style={yshift=0.15cm},
                ylabel style={yshift=-0.2cm},
                ytick={51, 53, 55, 57, 59, 61, 63},
                legend columns=2,
                xmin=-0.05,
                xmax=0.657,
                label style={font=\scriptsize},
                tick label style={font=\scriptsize},
                x tick label style={
                    /pgf/number format/.cd,
                        fixed,
                }
            ]
            \addplot[cdeepBP, only marks] table[col sep=comma, x=ASAGS, y=mIoU]{Data/correlation_asags.csv};
            \addplot[very thick, orange] table[col sep=comma, x=ASAGS, y={create col/linear regression = {y=mIoU}}
            ]{Data/correlation_asags.csv};
            \draw (0.5\linewidth, 0.35\linewidth) node {\scriptsize$\text{Corr}=0.71$};
            \end{axis}
        \end{tikzpicture}
    \end{subfigure}
    \hspace{1mm}
    \begin{subfigure}{.235\linewidth}
        \centering
        \begin{tikzpicture}
            \begin{axis}[
                legend style={nodes={scale=0.35}, at={(0.03, 0.24)}, anchor=west}, 
                xlabel={AP$(S;G)$},
                ylabel={mIoU (\%)},
                width=1.23\linewidth,
                height=1.23\linewidth,
                ymin=50.8,
                ymax=63.2,
                ytick={51, 53, 55, 57, 59, 61, 63},
                xlabel style={yshift=0.15cm},
                ylabel style={yshift=-0.2cm},
                legend columns=2,
                xmin=0.87,
                xmax=0.97,
                label style={font=\scriptsize},
                tick label style={font=\scriptsize},
                x tick label style={
                    /pgf/number format/.cd,
                        fixed,
                }
            ]
            \addplot[cdeepBP, only marks] table[col sep=comma, x=APSG, y=mIoU]{Data/correlation_apsg.csv};
            \addplot[very thick, orange] table[col sep=comma, x=APSG, y={create col/linear regression = {y=mIoU}}
            ]{Data/correlation_apsg.csv};
            \draw (0.5\linewidth, 0.35\linewidth) node {\scriptsize$\text{Corr} = -0.22$};
            \end{axis}
        \end{tikzpicture}
        % \caption{ASA(S;G) vs mIoU}
    \end{subfigure}
    \hspace{1mm}
    \begin{subfigure}{.235\linewidth}
        \centering
        \begin{tikzpicture}
            \begin{axis}[
                legend style={nodes={scale=0.35}, at={(0.03, 0.24)}, anchor=west}, 
                xlabel={AR$(S;G)$},
                ylabel={mIoU (\%)},
                width=1.23\linewidth,
                height=1.23\linewidth,
                ymin=50.8,
                ymax=63.2,
                xlabel style={yshift=0.15cm},
                ylabel style={yshift=-0.2cm},
                ytick={51, 53, 55, 57, 59, 61, 63},
                legend columns=2,
                xmin=0,
                xmax=0.057,
                label style={font=\scriptsize},
                tick label style={font=\scriptsize},
                x tick label style={
                    /pgf/number format/.cd,
                        fixed,
                    /tikz/.cd,
                }
            ]
            \addplot[cdeepBP, only marks] table[col sep=comma, x=ARSG, y=mIoU]{Data/correlation_arsg.csv};
            \addplot[very thick, orange] table[col sep=comma, x=ARSG, y={create col/linear regression = {y=mIoU}}
            ]{Data/correlation_arsg.csv};
            \draw (0.5\linewidth, 0.35\linewidth) node {\scriptsize$\text{Corr}=-0.02$};
            \end{axis}
        \end{tikzpicture}
    \end{subfigure}
    \hspace{-5mm}
    \begin{subfigure}{.235\linewidth}
        \centering
        \begin{tikzpicture}
            \begin{axis}[
                legend style={nodes={scale=0.35}, at={(0.03, 0.24)}, anchor=west}, 
                xlabel={AF$(S;G)$},
                ylabel={mIoU (\%)},
                width=1.23\linewidth,
                height=1.23\linewidth,
                ymin=50.8,
                ymax=63.2,
                ytick={51, 53, 55, 57, 59, 61, 63},
                xlabel style={yshift=0.15cm},
                ylabel style={yshift=-0.2cm},
                legend columns=2,
                xmin=0,
                xmax=0.082,
                label style={font=\scriptsize},
                tick label style={font=\scriptsize},
                x tick label style={
                    /pgf/number format/.cd,
                        fixed,
                    /tikz/.cd,
                }
            ]
            \addplot[cdeepBP, only marks] table[col sep=comma, x=AFSG, y=mIoU]{Data/correlation_afsg.csv};
            \addplot[very thick, orange] table[col sep=comma, x=AFSG, y={create col/linear regression = {y=mIoU}}
            ]{Data/correlation_afsg.csv};
            \draw (0.5\linewidth, 0.35\linewidth) node {\scriptsize$\text{Corr} = -0.01$};
            \end{axis}
        \end{tikzpicture}
        % \caption{ASA(S;G) vs mIoU}
    \end{subfigure}
    \hspace{1mm}
    \begin{subfigure}{.235\linewidth}
        \centering
        \begin{tikzpicture}
            \begin{axis}[
                legend style={nodes={scale=0.35}, at={(0.03, 0.24)}, anchor=west}, 
                xlabel={AP$(G;S)$},
                ylabel={mIoU (\%)},
                width=1.23\linewidth,
                height=1.23\linewidth,
                ymin=50.8,
                ymax=63.2,
                xlabel style={yshift=0.15cm},
                ylabel style={yshift=-0.2cm},
                ytick={51, 53, 55, 57, 59, 61, 63},
                legend columns=2,
                xmin=0.35,
                xmax=0.74,
                label style={font=\scriptsize},
                tick label style={font=\scriptsize},
                x tick label style={
                    /pgf/number format/.cd,
                        fixed,
                }
            ]
            \addplot[cdeepBP, only marks] table[col sep=comma, x=APGS, y=mIoU]{Data/correlation_apgs.csv};
            \addplot[very thick, orange] table[col sep=comma, x=APGS, y={create col/linear regression = {y=mIoU}}
            ]{Data/correlation_apgs.csv};
            \draw (0.5\linewidth, 0.35\linewidth) node {\scriptsize$\text{Corr}=-0.43$};
            \end{axis}
        \end{tikzpicture}
    \end{subfigure}
    \hspace{1mm}
    \begin{subfigure}{.235\linewidth}
        \centering
        \begin{tikzpicture}
            \begin{axis}[
                legend style={nodes={scale=0.35}, at={(0.03, 0.24)}, anchor=west}, 
                xlabel={AR$(G;S)$},
                ylabel={mIoU (\%)},
                width=1.23\linewidth,
                height=1.23\linewidth,
                ymin=50.8,
                ymax=63.2,
                ytick={51, 53, 55, 57, 59, 61, 63},
                xlabel style={yshift=0.15cm},
                ylabel style={yshift=-0.2cm},
                legend columns=2,
                xmin=0.21,
                xmax=0.696,
                label style={font=\scriptsize},
                tick label style={font=\scriptsize},
                x tick label style={
                    /pgf/number format/.cd,
                        fixed,
                }
            ]
            \addplot[cdeepBP, only marks] table[col sep=comma, x=ARGS, y=mIoU]{Data/correlation_args.csv};
            \addplot[very thick, orange] table[col sep=comma, x=ARGS, y={create col/linear regression = {y=mIoU}}
            ]{Data/correlation_args.csv};
            \draw (0.5\linewidth, 0.35\linewidth) node {\scriptsize$\text{Corr} = 0.57$};
            \end{axis}
        \end{tikzpicture}
        % \caption{ASA(S;G) vs mIoU}
    \end{subfigure}
    \hspace{1mm}
    \begin{subfigure}{.235\linewidth}
        \centering
        \begin{tikzpicture}
            \begin{axis}[
                legend style={nodes={scale=0.35}, at={(0.03, 0.24)}, anchor=west}, 
                xlabel={AF$(G;S)$},
                ylabel={mIoU (\%)},
                width=1.23\linewidth,
                height=1.23\linewidth,
                ymin=50.8,
                ymax=63.2,
                xlabel style={yshift=0.15cm},
                ylabel style={yshift=-0.2cm},
                ytick={51, 53, 55, 57, 59, 61, 63},
                legend columns=2,
                xmin=0.173,
                xmax=0.371,
                label style={font=\scriptsize},
                tick label style={font=\scriptsize},
                x tick label style={
                    /pgf/number format/.cd,
                        fixed,
                }
            ]
            \addplot[cdeepBP, only marks] table[col sep=comma, x=AFGS, y=mIoU]{Data/correlation_afgs.csv};
            \addplot[very thick, orange] table[col sep=comma, x=AFGS, y={create col/linear regression = {y=mIoU}}
            ]{Data/correlation_afgs.csv};
            \draw (0.5\linewidth, 0.35\linewidth) node {\scriptsize$\textbf{Corr}=\textbf{0.95}$};
            \end{axis}
        \end{tikzpicture}
    \end{subfigure}
    \caption{{\em Relationship between metrics and mIoU.} The correlation between AF$(G;S)$ and mIoU is especially high. For the correlation calculation, \textit{Oracle} in Table \ref{tab:quantitative} is excluded.}
    \label{fig:sup-correlation}
\end{figure*}

% To address this issue, we sample the values uniformly by confidence order for all pixels inside the superpixel, without using the distribution for all pixels, which is described in Figure \ref{fig:sup-knee-points}.
% We sample 20 and 5 pixels for Cityscapes and PASCAL datasets, respectively.
% hyperparameter for sieving, easy class, difficult class, different confidence
% When we set the same threshold for all pixels, only the common classes remain, and the IoU of rare classes decreases in Figure \ref{fig:sup-class-dist}.

\begin{figure*}[t!]
    \captionsetup[subfigure]{font=footnotesize}
    \centering
    \begin{subfigure}[!ht]{.245\linewidth}
        \centering
        \includegraphics[scale=0.238]{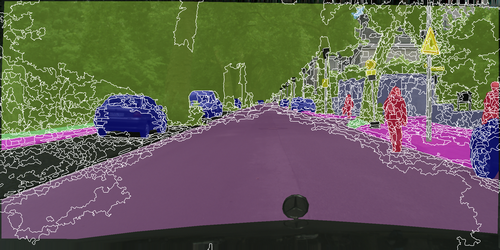}
    \end{subfigure}
    \begin{subfigure}[!ht]{.245\linewidth}
        \centering
        \includegraphics[scale=0.238]{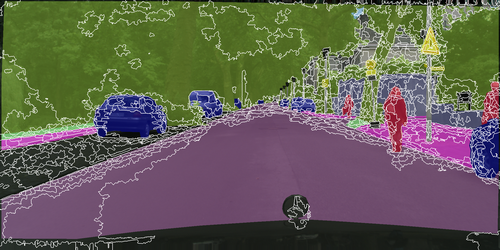}
    \end{subfigure}
    \begin{subfigure}[!ht]{.245\linewidth}
        \centering
        \includegraphics[scale=0.238]{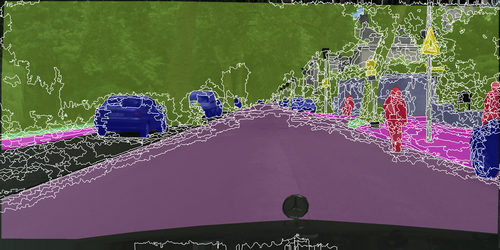}
    \end{subfigure}
    \begin{subfigure}[!ht]{.245\linewidth}
        \centering
        \includegraphics[scale=0.238]{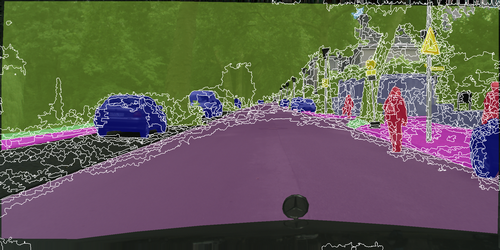}
    \end{subfigure}

    \begin{subfigure}[!ht]{.245\linewidth}
        \centering
        \includegraphics[scale=0.238]{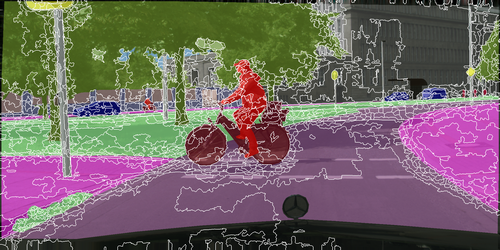}
    \end{subfigure}
    \begin{subfigure}[!ht]{.245\linewidth}
        \centering
        \includegraphics[scale=0.238]{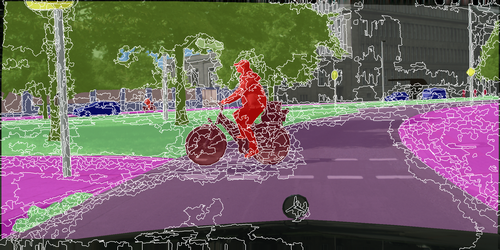}
    \end{subfigure}
    \begin{subfigure}[!ht]{.245\linewidth}
        \centering
        \includegraphics[scale=0.238]{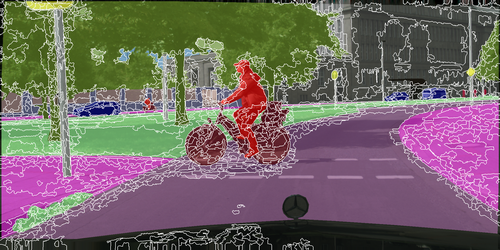}
    \end{subfigure}
    \begin{subfigure}[!ht]{.245\linewidth}
        \centering
        \includegraphics[scale=0.238]{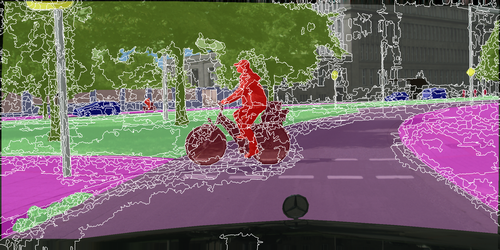}
    \end{subfigure}

    \begin{subfigure}[!ht]{.245\linewidth}
        \centering
        \includegraphics[scale=0.238]{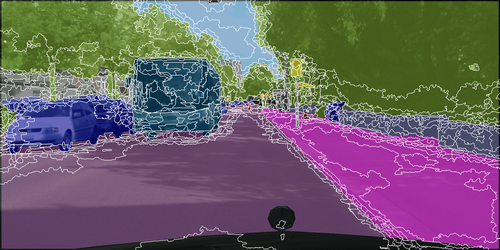}
        \caption{Adaptive merged $(t=1)$}
    \end{subfigure}
    \begin{subfigure}[!ht]{.245\linewidth}
        \centering
        \includegraphics[scale=0.238]{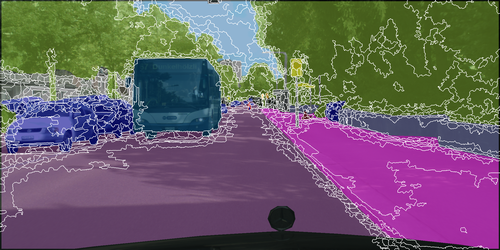}
        \caption{Adaptive merged $(t=2)$}
    \end{subfigure}
    \begin{subfigure}[!ht]{.245\linewidth}
        \centering
        \includegraphics[scale=0.238]{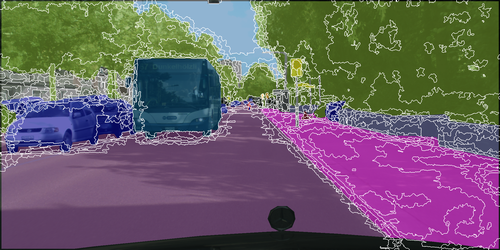}
        \caption{Adaptive merged $(t=3)$}
    \end{subfigure}
    \begin{subfigure}[!ht]{.245\linewidth}
        \centering
        \includegraphics[scale=0.238]{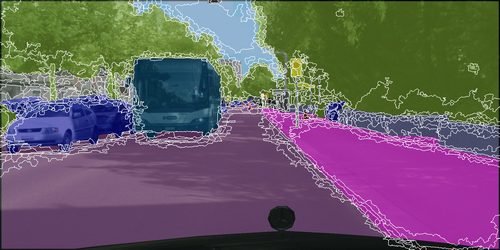}
        \caption{Adaptive merged $(t=4)$}
    \end{subfigure}
    % \caption{{\em Round} (a) We begin active learning with over-segmented superpixels. (b, c) In each round $t$, we merge superpixels in an adaptive manner using the model from the previous round. % $t-1$. 
    % (d) As the round progresses, adaptive superpixels look similar to oracle ones.}
    \caption{{\em Qualitative results with varying round.}
    (a-d) Superpixels generated with proposed adaptive merging at rounds 1 to 4.
    Thanks to the improved model, we observe that the merging becomes more accurate as the round increases. We use the model reported in Figure~\ref{fig:(a)-effect}.}
    \label{fig:sup-round}
\end{figure*}

\begin{figure*}[t!]
    \captionsetup[subfigure]{font=footnotesize}
    \centering
    \begin{subfigure}[!ht]{.245\linewidth}
        \centering
        \includegraphics[scale=0.238]{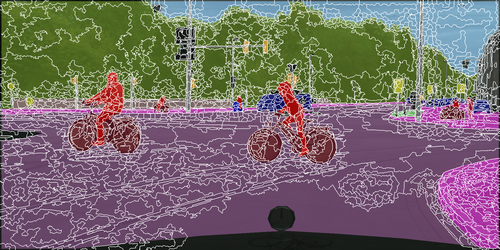}
    \end{subfigure}
    \begin{subfigure}[!ht]{.245\linewidth}
        \centering
        \includegraphics[scale=0.238]{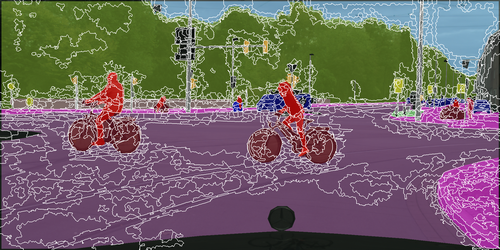}
    \end{subfigure}
    \begin{subfigure}[!ht]{.245\linewidth}
        \centering
        \includegraphics[scale=0.238]{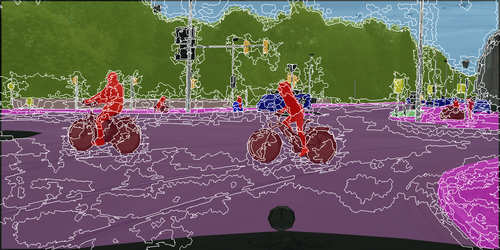}
    \end{subfigure}
    \begin{subfigure}[!ht]{.245\linewidth}
        \centering
        \includegraphics[scale=0.238]{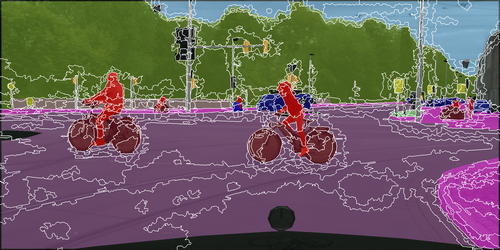}
    \end{subfigure}

    \begin{subfigure}[!ht]{.245\linewidth}
        \centering
        \includegraphics[scale=0.238]{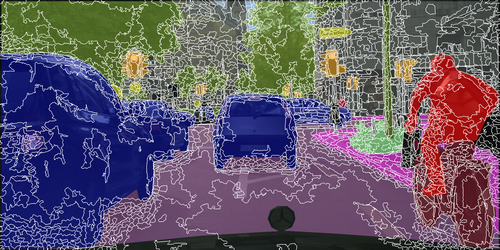}
    \end{subfigure}
    \begin{subfigure}[!ht]{.245\linewidth}
        \centering
        \includegraphics[scale=0.238]{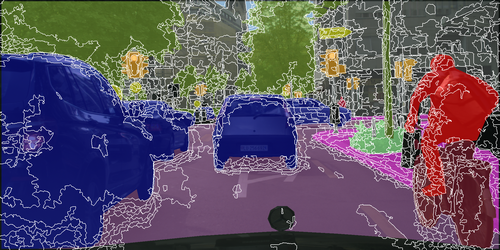}
    \end{subfigure}
    \begin{subfigure}[!ht]{.245\linewidth}
        \centering
        \includegraphics[scale=0.238]{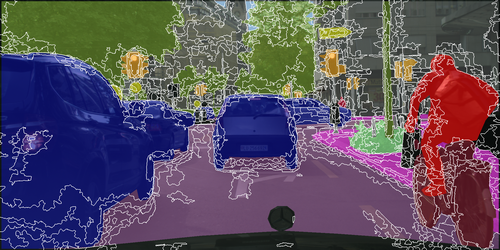}
    \end{subfigure}
    \begin{subfigure}[!ht]{.245\linewidth}
        \centering
        \includegraphics[scale=0.238]{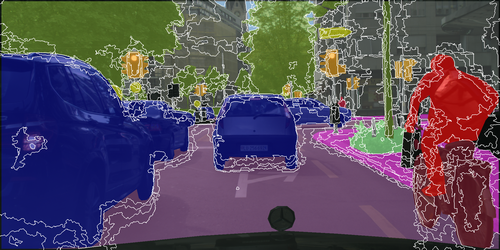}
    \end{subfigure}

    \begin{subfigure}[!ht]{.245\linewidth}
        \centering
        \includegraphics[scale=0.238]{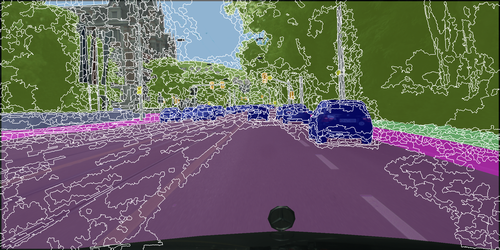}
        \caption{Adaptive merged $(\epsilon=0.05)$}
    \end{subfigure}
    \begin{subfigure}[!ht]{.245\linewidth}
        \centering
        \includegraphics[scale=0.238]{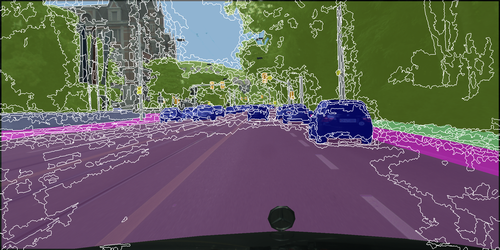}
        \caption{Adaptive merged $(\epsilon=0.1)$}
    \end{subfigure}
    \begin{subfigure}[!ht]{.245\linewidth}
        \centering
        \includegraphics[scale=0.238]{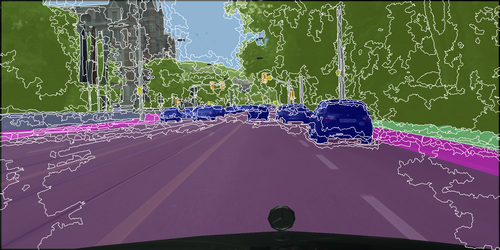}
        \caption{Adaptive merged $(\epsilon=0.15)$}
    \end{subfigure}
    \begin{subfigure}[!ht]{.245\linewidth}
        \centering
        \includegraphics[scale=0.238]{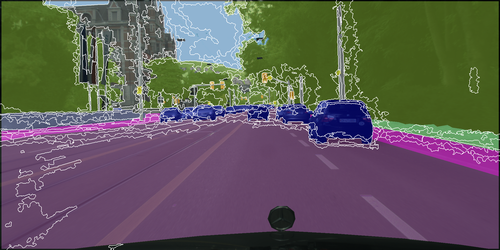}
        \caption{Adaptive merged $(\epsilon=0.2)$}
    \end{subfigure}
    % \caption{{\em Epsilon.} (a) We begin active learning with over-segmented superpixels. (b, c) In each round $t$, we merge superpixels in an adaptive manner using the model from the previous round. % $t-1$. 
    % (d) As the round progresses, adaptive superpixels look similar to oracle ones.}
    \caption{{\em Qualitative results with varying $\epsilon$.}
    (a-d) Superpixels are generated with proposed adaptive merging with $\epsilon$: 0.05, 0.1, 0.15, 0.2. % $t-1$. 
    We observe that an increase in $\epsilon$ gives more aggressive merging. Merging is conducted on Cityscapes with a base superpixel size of 256. }
    \label{fig:sup-epsilon}
\end{figure*}

\begin{figure*}[t!]
    \captionsetup[subfigure]{font=footnotesize}
    \centering
    \begin{subfigure}{.33\linewidth}
        \centering
        \includegraphics[scale=0.322]{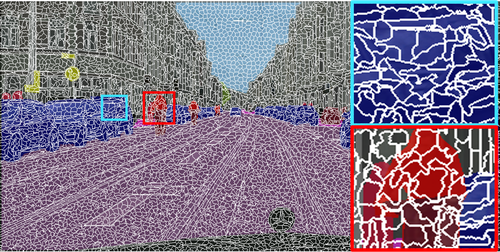}
        % \caption{$\text{ASA}(S;G)=0.021, \; \text{AF}(G;S)=0.355$}
        % \vspace{2mm}
        % \label{(a)-qualitative}
    \end{subfigure}
    \begin{subfigure}{.33\linewidth}
        \centering
        \includegraphics[scale=0.322]{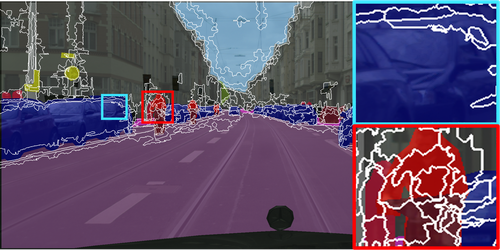}
        % \includegraphics[scale=0.5]{Figures/fig4_b_1.png}
        % \caption{$\text{ASA}(S;G)=0.89, \; \text{AF}(G;S)=0.283$}
        % \vspace{2mm}
        % \label{(b)-qualitative}
    \end{subfigure}
    \begin{subfigure}{.33\linewidth}
        \centering
        \includegraphics[scale=0.322]{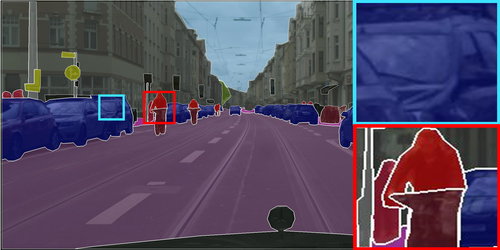}
        % \includegraphics[scale=0.5]{Figures/fig4_c_1.png}
        % \caption{$ASA(S;G)=1.00, \; AF(G;S)=1.00$}
        % \vspace{2mm}
        % \label{(c)-qualitative}
    \end{subfigure}

    \begin{subfigure}{.33\linewidth}
        \centering
        \includegraphics[scale=0.322]{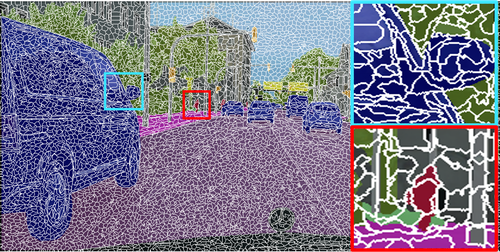}
    \end{subfigure}
    \begin{subfigure}{.33\linewidth}
        \centering
        \includegraphics[scale=0.322]{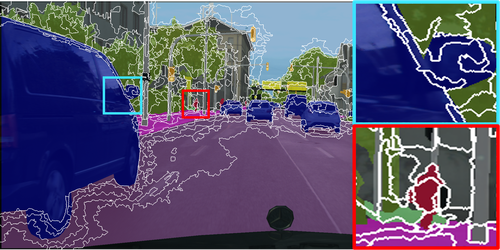}
    \end{subfigure}
    \begin{subfigure}{.33\linewidth}
        \centering
        \includegraphics[scale=0.322]{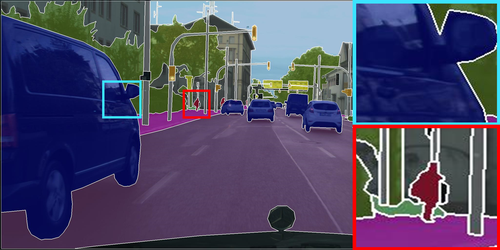}
        % \includegraphics[scale=0.5]{Figures/fig4_c_2.png}
        % \caption{$ASA(S;G)=1.00, \; AF(G;S)=1.00$}
    \end{subfigure}

    \begin{subfigure}{.33\linewidth}
        \centering
        \includegraphics[scale=0.322]{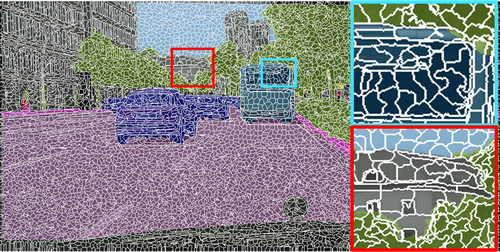}
        \caption{Base superpixels~\cite{van2012seeds}}
    \end{subfigure}
    \begin{subfigure}{.33\linewidth}
        \centering
        \includegraphics[scale=0.322]{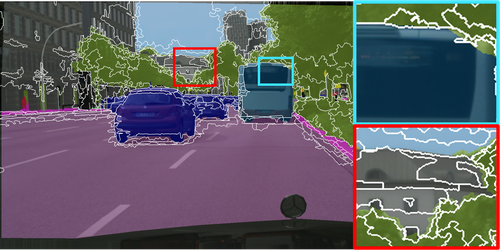}
        \caption{Merged superpixels (Ours)}
    \end{subfigure}
    \begin{subfigure}{.33\linewidth}
        \centering
        \includegraphics[scale=0.322]{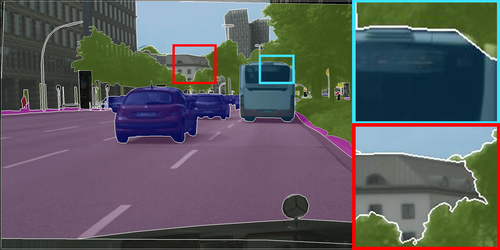}
        % \includegraphics[scale=0.5]{Figures/fig4_c_2.png}
        % \caption{$ASA(S;G)=1.00, \; AF(G;S)=1.00$}
        \caption{Oracle superpixels}
    \end{subfigure}
    %\caption{{\em Qualitative results of adaptive superpixels.} As the round progresses, (a) over-segmented superpixels becomes (b) adaptively merged ones, and they resemble (c) oracle superpixels, especially for the classes that the model is confident about.}
    \caption{{\em Qualitative results of adaptive superpixels.} (a) Base superpixel generated by SEEDS~\cite{van2012seeds} with size 256. (b) Superpixels generated with proposed adaptive merging at round 4. (c) Oracle superpixels generated from the ground truth.}
    \label{fig:sup-same-paper}
\end{figure*}

\section{Further discussion on the oracle superpixels}
\label{fig:sup-oracle}

In Section \ref{para:oracle-superpixels},
we introduce the oracle superpixels,
which we believe is an achievable optimal set of superpixels for active learning.
For clarification,  
we provide the detailed process of generating the proposed oracle superpixels.
In addition, we provide further insights
into the achievable notion of optimal superpixels.
%In order to demonstrate effectiveness of our merging process, we require an upper bound of AL performance we aim to achieve. To this end, we generate ideal superpixels, each without any noise. Such superpixels are called oracle superpixels.

The Cityscapes dataset is equipped with the ground-truth annotations for semantic segmentation, represented by dense pixel-wise labels: \ie., each pixel in an annotated image is assigned an ID that represents a ground-truth semantic category~(Figure~\ref{subfig:sementic-seg}). In such annotation, each group of pixels that share the same ID aligns perfectly with the boundary of semantic objects. However, each such group is not guaranteed to be a single-connected component of pixels.
% and hence is not a proper superpixel.
For example, different cars in Figure~\ref{subfig:sementic-seg} are assigned the same blue color despite being physically separated, and a car divided into two parts due to an obstructing pole is still colored blue. 
% In Figure~\ref{subfig:sementic-seg}, for example, different cars are away from each other, but are assigned the same blue color. 
% In addition, a car in Figure~\ref{subfig:sementic-seg} is divided into two parts due to an obstructing pole, but is also colored blue. 
This is opposed to what we hope to achieve by merging two adjacent superpixels repeatedly.
To address this issue, we subdivide each superpixel as necessary to ensure that every pixel within a superpixel is adjacent to each other.
% so that every pixel within a superpixel is adjacent to each other. 
We utilize OpenCV~\cite{opencv_library} and Shapely~\cite{shapely2007} to identify the maximal connected component of pixels sharing the same semantic. 
We apply the same procedure to annotated images in the PASCAL dataset
% to identify maximal connected components.
% Similarly, PASCAL dataset provides semantic annotations in dense pixel-wise assignment of classes. To identify boundary-preserving, maximal connnected components, we apply the aforementioned procedure to the annotated images. 
Figure~\ref{fig:sup-oracle-superpixels} illustrates the distinction between conventional semantic and panoptic segmentation and our oracle superpixels.

The Cityscapes and PASCAL datasets are divided into 327k and 16k oracle superpixels, respectively.
% , respectively to 408K and 16k oracle superpixels.
It is worth noting that the PASCAL has a lower number of oracle superpixels due to the smaller number of classes per image.
In other words, only a few objects are of interest in each image, and the rest are simply treated as the background.

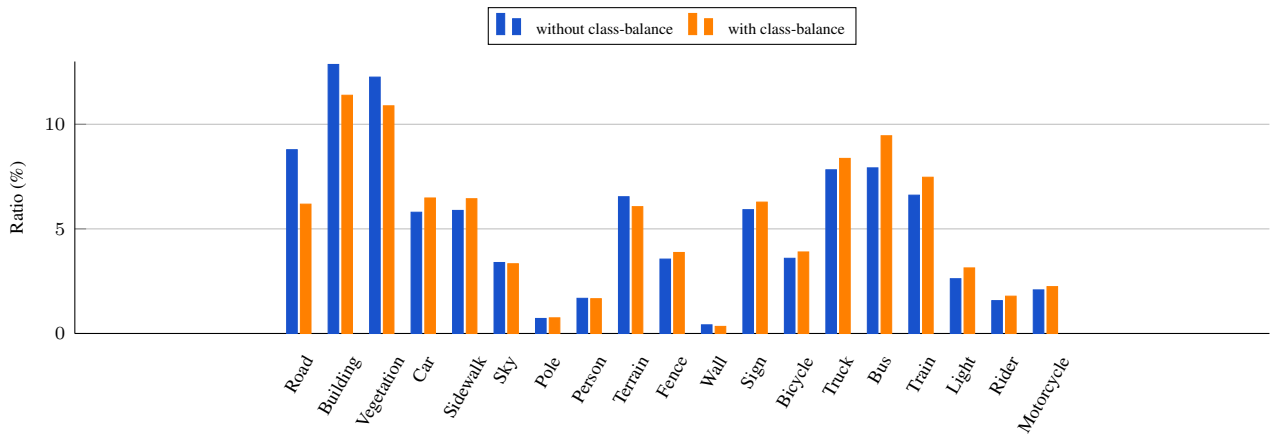
\begin{figure*}[t!]
\captionsetup[subfigure]{font=footnotesize,labelfont=footnotesize,aboveskip=0.05cm,belowskip=-0.15cm}
\centering
\begin{tikzpicture}
    \begin{axis}[
        width  = \textwidth,
        axis y line*=left,
        symbolic x coords={
            \rotatebox{60}{Road},
            \rotatebox{60}{Building},
            \rotatebox{60}{Vegetation},
            \rotatebox{60}{Car},
            \rotatebox{60}{Sidewalk},
            \rotatebox{60}{Sky},
            \rotatebox{60}{Pole},
            \rotatebox{60}{Person},
            \rotatebox{60}{Terrain},
            \rotatebox{60}{Fence},
            \rotatebox{60}{Wall},
            \rotatebox{60}{Sign},
            \rotatebox{60}{Bicycle},
            \rotatebox{60}{Truck},
            \rotatebox{60}{Bus},
            \rotatebox{60}{Train},
            \rotatebox{60}{Light},
            \rotatebox{60}{Rider},
            \rotatebox{60}{Motorcycle},
        },
        axis x line=bottom,
        height = 5.2cm,
        major x tick style = transparent,
        %axis on top,
        ybar=3*\pgflinewidth,
        bar width=4pt,
        ymajorgrids = true,
        ylabel = {Ratio (\%)},
        xtick = data,
        scaled y ticks = false,
        enlarge x limits=0.3,
        axis line style={-},
        ymin=0.0,ymax=13,
        legend columns=2,
        legend cell align=left,
        legend style={
                nodes={scale=0.6},
                at={(0.5,1.2)},
                anchor=north,
                column sep=1ex
        },
        label style={font=\scriptsize},
        tick label style={font=\scriptsize}
    ]
        \addplot[style={cdeepBP,fill=cdeepBP,mark=none}] coordinates {
            (\rotatebox{60}{Road}, 8.784105735)
            (\rotatebox{60}{Building}, 12.85866043)
            (\rotatebox{60}{Vegetation}, 12.25386072)
            (\rotatebox{60}{Car}, 5.79546044)
            (\rotatebox{60}{Sidewalk}, 5.881853763)
            (\rotatebox{60}{Sky}, 3.392110938)
            (\rotatebox{60}{Pole}, 0.717479588)
            (\rotatebox{60}{Person}, 1.680341609)
            (\rotatebox{60}{Terrain}, 6.53641597)
            (\rotatebox{60}{Fence}, 3.555288862)
            (\rotatebox{60}{Wall}, 0.41288854)
            (\rotatebox{60}{Sign}, 5.917555467)
            (\rotatebox{60}{Bicycle}, 3.593630835)
            (\rotatebox{60}{Truck}, 7.823520075)
            (\rotatebox{60}{Bus}, 7.91594498)
            (\rotatebox{60}{Train}, 6.611388774)
            (\rotatebox{60}{Light}, 2.617009992)
            (\rotatebox{60}{Rider}, 1.569643873)
            (\rotatebox{60}{Motorcycle}, 2.082839412)
        };
        \addplot[style={orange,fill=orange,mark=none}] coordinates {
            (\rotatebox{60}{Road}, 6.181659949)
            (\rotatebox{60}{Building}, 11.38706875)
            (\rotatebox{60}{Vegetation}, 10.88211877)
            (\rotatebox{60}{Car}, 6.476668129)
            (\rotatebox{60}{Sidewalk}, 6.440680182)
            (\rotatebox{60}{Sky}, 3.333062128)
            (\rotatebox{60}{Pole}, 0.747530823)
            (\rotatebox{60}{Person}, 1.661918156)
            (\rotatebox{60}{Terrain}, 6.061194039)
            (\rotatebox{60}{Fence}, 3.871472033)
            (\rotatebox{60}{Wall}, 0.337991765)
            (\rotatebox{60}{Sign}, 6.279293782)
            (\rotatebox{60}{Bicycle}, 3.896345415)
            (\rotatebox{60}{Truck}, 8.368686854)
            (\rotatebox{60}{Bus}, 9.450410149)
            (\rotatebox{60}{Train}, 7.468290798)
            (\rotatebox{60}{Light}, 3.133602892)
            (\rotatebox{60}{Rider}, 1.782747288)
            (\rotatebox{60}{Motorcycle}, 2.239258099)
        };
        % \addplot[style={cdeepMF,fill=cdeepMF,mark=none}]
             % coordinates {(Road, 19.22) (Building, 21.29) (Vegetation, 21.58)};
        \legend{without class-balance, with class-balance}
    \end{axis}
\end{tikzpicture}
\caption{{\em Effect of class-balanced acquisition function.} According to the ground-truth, class labels are arranged based on the total pixel count for each class, \ie classes become rarer in images as you move from left to right along the x-axis. We observe that classes on the left are selected less with the class-balanced term, while classes on the right are selected more.}
\label{fig:class-balanced}
\end{figure*}

\section{Further analysis on the achievable metrics}
\label{fig:sup-metrics}
In Table \ref{tab:quantitative}, we evaluate various superpixels using eight metrics with oracle superpixels as ground-truth $G$.
% We utilize oracle superpixels as ground-truth G and evaluate various superpixels with eight metrics in Table \ref{tab:quantitative}.
Figure \ref{fig:sup-correlation} shows the correlation between each metric and mIoU. 
We observe that our AF$(G;S)$ can be utilized to look-ahead a model's performance in active learning without training. 
In addition, we examine how different ground-truth $G$ impacts AF$(G;S)$.
In the field of semantic segmentation, two conventional segmentations, semantic and panoptic segmentations in Figure \ref{fig:sup-oracle-superpixels}, are widely used as ground-truth.
Figure \ref{fig:sup-afgs-g} indicates that using panoptic segmentation and oracle superpixels for $G$ results in higher correlation between AF$(G;S)$ and mIoU than semantic segmentation.
However, obtaining panoptic segmentation requires more costs than semantic segmentation since it utilizes additional instance information.
It is worth noting that our oracle superpixels (Figure \ref{subfig:oralce-seg}) can be easily generated even in cost-limited practical situations as they are produced from semantic segmentation (Figure \ref{subfig:sementic-seg}).

% \section{Effect of pixel-level class popularity}
% uncertainty

% class-balanced

% size-aware class-balanced

\begin{table*}[t!]
\centering
\setlength\tabcolsep{6pt}
\begin{tabular}{c|l}
\toprule
Notations & Description \\ \midrule
$\mathcal{I}$ & the set of unlabeled images \\ \midrule
$\mathcal{C}$ & the set of class labels \\ \midrule
$t$ & a round \\ \midrule
$x$ & a pixel \\ \midrule
$s$ & a superpixel \\ \midrule
$S_t(i)$ & the set of superpixels in an image $i$ in round $t$ \\ \midrule
$\mathcal{S}_t$ & the set of superpixels in all images in round $t$, $\mathcal{S}_t := \bigcup_{i \in \mathcal{I}} S_t(i)$ \\ \midrule
$B$ & the query budget per round \\ \midrule
$\mathcal{B}_t$ & the set of $B$ selected superpixels in round $t$, $B_t \subset \mathcal{S}_t, |B_t| = B$\\ \midrule
$\theta_t$ & the model at the end of round $t$ \\ \midrule
$y_\theta(x)$ & the estimated dominant label of pixel $x$ given $\theta$ \\ \midrule
$\text{D}(s)$ & the true dominant label of superpixel $s$ \\ \midrule
$\text{D}_\theta(s)$ & the estimated dominant label of superpixel $s$ given $\theta$ \\ \midrule
\multirow{2}{*}{$\mathcal{G}(S) := (S, \mathcal{E}(S))$} & the graph consisting of the superpixels in $S$ as nodes and
the edge set $\mathcal{E}(S)$ \\
& such that $(s, n) \in \mathcal{E}(S)$ for each pair of adjacent superpixels $s, n \in S$.  \\ \midrule
% $\mathcal{E}(S)$      & the set of edges in graph $\mathcal{G}(S)$ \\ \midrule
$\epsilon$      & the hyperparameter for merging in \eqref{eq:jsd} \\
\bottomrule
\end{tabular}
\caption{{\em Notations.} The notations used in the paper are defined.}
\label{tab:notations}
\end{table*}

\section{Additional qualitative adaptive superpixels}
% \section{Additional qualitative analyses on the adaptive merging}
% \section{Qualitative analyses}
\label{fig:sup-qual}
To facilitate comprehension of the merged superpixels, we display superpixels generated across diverse settings.
The appearance of merged superpixels is mainly determined by the model's performance and $\epsilon$. 
Figure \ref{fig:sup-round} highlights that as the round progresses, the model's performance improves, leading to more accurate merging. 
With the model fixed at round 4, Figure \ref{fig:sup-epsilon} shows the impact of adjusting $\epsilon$.
As $\epsilon$ grows, the merging process intensifies, ultimately decreasing the overall number of superpixels.
In addition, Figure \ref{fig:sup-same-paper} shows further examples of our merged superpixels.

\section{Class-balanced sampling}
% \khy{
% Our acquisition function in \eqref{acquisition_function} gives priority to uncertain superpixels of rare classes. 
To observe the impact of the class-balanced acquisition function in~\eqref{acquisition_function}, we analyze the class distribution of selected superpixels both with and without the class-balanced term. 
In Figure~\ref{fig:class-balanced}, where class labels are sorted such that the left (road) and right (motorcycle) ends represent the most and least popular classes, it is evident that the class-balanced term results in a higher selection of rarer classes, as intended.

\end{document}